\documentclass[twoside,11pt]{article}

\usepackage[abbrvbib, preprint]{jmlr2e}

\usepackage{blindtext}
\usepackage{subcaption}
\usepackage{amsmath}
\usepackage{graphicx}
\usepackage{placeins}

\usepackage{lastpage}
\jmlrheading{23}{2025}{1-\pageref{LastPage}}{1/21; Revised 5/25}{9/25}{21-0000}{Hesham Abdelmotaleb et al}

\ShortHeadings{Word Embedding Techniques for Classification of star ratings}{Word Embedding Techniques for Classification of Star Ratings}
\firstpageno{1}

\begin{document}

\title{Word Embedding Techniques for Classification of Star Ratings}

\author{
\name Hesham Abdelmotaleb \email hesham.abdelmotaleb@plymouth.ac.uk \\
       \addr Centre for Mathematical Sciences\\
       University of Plymouth\\
  \AND
\name Craig McNeile \email craig.mcneile@plymouth.ac.uk \\
       \addr Centre for Mathematical Sciences\\
       University of Plymouth\\
        \AND 
\name Ma{\l}gorzata Wojty\'s \email malgorzata.wojtys@plymouth.ac.uk \\
       \addr Centre for Mathematical Sciences\\
       University of Plymouth\\
}

\editor{My editor}

\maketitle
    \begin{abstract}
        Telecom services are at the core of today's societies' everyday needs. The availability of numerous online forums and discussion platforms enables telecom providers to improve their services by exploring the views of their customers to learn about common issues that the customers face. Natural Language Processing (NLP) tools can be used to process the free text collected.

One way of working with such data is to represent text as numerical vectors using one of many word embedding models based on neural networks. This research uses a novel dataset of telecom customers' reviews to perform an extensive study showing how different word embedding algorithms can affect the text classification process. Several state-of-the-art word embedding techniques are considered, including BERT, Word2Vec and Doc2Vec, coupled with several classification algorithms. The important issue of feature engineering and  dimensionality reduction is addressed and several PCA-based approaches are explored. Moreover, the energy consumption used by the different word embeddings is investigated. The findings show that some word embedding models can lead to consistently better text classifiers in terms of precision, recall and F1-Score. In particular, for the more challenging classification tasks, BERT combined with PCA stood out with the highest performance metrics. Moreover, our proposed PCA approach of combining word vectors using the first principal component shows clear advantages in performance over the traditional approach of taking the average.

\end{abstract}


\section{Introduction}

There is potentially a lot of useful insight into customer views of
a company and its products from free text in online discussion forums and social media
posts. For example, persistent negative online views of a product may result in
future reduced revenue from customers no longer buying products or stopping paying for subscription service (customer churn). Hence, it is
important to be able to process free text using Natural
Language Processing (NLP) tools. This could include summarizing a large chunk of free text or incorporating features from free text into classifiers.

The goal of this research is to investigate techniques to build classifiers of customer star ratings from the customer's free text. 
For the dataset, we use short free texts about telecom 
companies written by customers on online platforms. Broadband and mobile connectivity are crucial to the running
of the economy. In principle, it is easy for customers to switch broadband
providers, so monitoring customer feedback and building classifiers to
predict customer churn is important for the profitability of a telecom company. We use open data to carefully study the performance of critical aspects of classification algorithms.

In this era of big data, it is strategic to study the performance of machine learning algorithms as the sizes of datasets increase \citep{Upadhyay2020Gradient}. Also, due to concerns about the climate crisis, an increasingly important aspect of data science computations is minimizing the energy consumed in training and using machine learning models \citep{lottick2019, mehlin2023towards}. Hence, we measure the energy consumption and CO2 emissions in our benchmarks.

In the empirical study, we consider the classification of star ratings based on telecom customers' opinions written online. 
Specifically, we explore the use of word embeddings and feature extraction and perform a comparative study of a wide range of algorithms in the context of a classification task in the telecom data field, to find the most effective approach that is time-efficient and accurate. We do not explicitly make use of the currently popular large language models (LLM) to build classifiers. However, we note that word embeddings are a key part of LLMs \citep{Mahajan2025Revisiting}.

Several previous studies explored building classifiers to predict star ratings, highlighting the importance of this task in many applications, including sentiment analysis, customer churn prediction and text classification. \cite{Binder2019} explained the idea of stars in the field of online customer reviews, while applying regression to model the rating value. After using a two-step approach to assess the results, an underlying model for preferences on a continuous scale and a nonlinear transformation of the underlying preferences onto the rating scale were employed. \cite{Monett2016} focused on predicting the review's star rating using annotated mobile app reviews, collected from Google Play Store, by using sentiment analysis as a feature extraction  model and then regression algorithms to predict each review's rating. After exploring two approaches: with and without neutral phrases in the regression model, \cite{Monett2016} found that using "linear regression with polarity average will produce better results",  and a lower MSE was observed when excluding neutral phrases.
In a different study, \cite{Saumya2018} researched how ranking online consumer reviews can be measured, expected and calculated. The authors extracted 17 features from three types of datasets, combining linguistic, numerical and NLP features. 
The results showed that Random Forest generally performed better than Naive Bayes and SVM, as in the 'baseline conceptual model' it achieved F1-Score values $0.88$ compared to $0.70$ for Naive Bayes and $0.80$ for SVM.

The focus of this research is to compare a broad spectrum of popular word embedding methods in the context of building a classification model for star ratings based on short texts written by customers. Many studies that compare selected word embedding algorithms for various tasks have been conducted. For example, \cite{Wang2018Comparison} evaluated word embeddings that trained from different corpora (clinical notes, biomedical literature, Wikipedia and news) for biomedical NLP applications. The authors found that embeddings trained on clinical notes (EHR) and biomedical literature (MedLit) capture medical semantics better than general-domain embeddings trained from GloVe and Google News. They also found that adding word embeddings as features generally improved classification results but did not always guarantee a better performance. 
\cite{Sabbeh2023Comparative} compared BERT, AraBERT, FastText, Word2Vec, AraVec and GloVe for Arabic sentiment classification, using five datasets with different dialects and domains. The authors found that AraBERT showed better performance amongst other word embeddings and that BiLSTM classifier achieved higher accuracy than CNN for three datasets, but CNN performed better on a smaller dataset. Moreover, the custom-trained embeddings generally outperformed pre-trained ones, emphasizing the importance of domain-specific training.
\cite{Wang2020Comparative} studied the classic and contextualized word embeddings for text classification comparing Word2Vec, GloVe and FastText to ELMo and BERT, using CNN and BiLSTM as classifiers.
The results indicated that CNN outperformed BiLSTM for document classification, whereas BiLSTM was more effective for sentence-level sentiment analysis. BERT generally outperformed ELMo, particularly for longer documents and the study suggests using CNN with classic embeddings for document classification and BiLSTM with contextualized embeddings for sentence-level classification.

Feature engineering and dimension reduction are  an important and challenging aspects of model building when using word embeddings because of the very high dimensionality of the feature space. 
\cite{Lebret2013Word} explored the use of Principal Component Analysis (PCA) with word embeddings, specifically applying Hellinger PCA to a word co-occurrence matrix. The authors found that PCA could produce embeddings comparable to those derived from deep learning models while being much faster and computationally efficient. 
The resulting embeddings performed well on NLP tasks such as Named Entity Recognition and sentiment analysis. Furthermore, fine-tuning these embeddings with neural networks enhanced performance, particularly for tasks requiring semantic understanding. The study suggests that while deep learning is useful for task-specific adaptations, simple spectral methods like PCA can be an effective alternative for generating high-quality word embeddings.
 In a different study, \cite{Gupta2019Improving} explored how to improve word embeddings using Kernel Principal Component Analysis (KPCA). The authors demonstrated that applying KPCA to word similarity matrices improved its performance, providing a better initialization for word embedding models such as Word2Vec and FastText, especially for small datasets and infrequent words. Additionally, KPCA embeddings lead to faster convergence reducing training time while maintaining high-quality representations. Moreover, \cite{Taloba2018Comparative} investigated the use of PCA for feature extraction with TF-IDF in the context of text classification. Random Forest, SVM, Decision Tree and KNN classifiers were considered. The authors found that applying PCA to TF-IDF text representation improved classifier performance. Among the classifiers tested, Random Forest showed the most significant performance gain with PCA. Therefore, overall, PCA is a promising method for enhancing text classification by optimizing feature representations.

The aim of this research is to study and compare a broad range of state-of-the-art word embeddings employed for a task of short texts classification to distinguish between customers' satisfaction levels. We investigate the following word embedding methods: Word2Vec, FastText, BERT and Doc2Vec and explore two ways of combining the word vectors: using the average and the first principal component. PCA-based dimension reduction for BERT and Doc2Vec is also studied. Moreover, TF-IDF text representations are included for comparison.  We apply supervised classification algorithms: Logistic Regression, Random Forest, Gradient Boosting classifier, Stochastic Gradient Descent, Decision Tree, Support Vector Machine and K-Nearest Neighbours (KNN). Their performance is measured using precision, recall and F1-Score \citep{Powers2020Evaluation}. The methods are applied to data collected from online comments of telecommunications customers and several scenarios are considered including binary classification and multiclass classification of the number of stars. All the datasets are original for this research and were obtained by scraping the data from online forums. By investigating several popular word embeddings, various feature extraction methods and a wide range of classifiers in one study, a comprehensive comparison is achieved, making the findings highly relevant for practical applications in the studied domain.

The paper is organised as follows. Section \ref{SECTION4.1} discusses the word embeddings and feature engineering methods to be applied to the datasets and describes the study design. Section \ref{ResultsSection} discusses the results, including the performance, time and energy consumption. In section \ref{ConclusionSection} we report the conclusions from the study.

\section{Methodology and Study Design}
\label{SECTION4.1}

\subsection{Word Embedding Models}

Word Embeddings (WEs) are a type of NLP techniques that represent words as numerical vectors. 
They have been shown to be useful and effective for many NLP tasks, including text classification, machine translation and sentiment analysis \citep{Deho2018}.
Once the WEs have been trained, they can be used as input to Machine Learning models to classify customer service scripts into different categories. Then, their outputs can help to develop and increase customer service efficiency and to find common customer issues. From the many types of WEs that have been developed, this study examines the following:

\begin{enumerate}

\item Word2Vec \citep{Mikolov2013Efficient}, which is based on employing a neural network and involves predicting the context of a word in a sentence \citep{Kottur2016}. Word2Vec generates vectors for each word using a neural network model which was trained on a large corpus of text.

\item Doc2Vec, which is an extension of Word2Vec that represents entire documents as numerical vectors. Doc2Vec also trains WEs by predicting the context of a word in a document \citep{Shao2018}.

\item FastText is another extension of Word2Vec that can handle out-of-vocabulary words and was developed by Facebook AI Research. FastText algorithm trains WEs by representing each word as a bag of character n-grams \citep{Lakmal2020}, producing 300-dimensional vector for each word \citep{Yao2020text}. FastText supports both Skip-gram and CBOW structures.

\item BERT (Bidirectional Encoder Representations from Transformers) \citep{Chang2019} is a deep learning algorithm that is typically trained on a very large corpus of text. BERT can be used not only to create WEs but also other NLP tasks, such as machine translation and question answering \citep{Wang2019}.  BERT considers the context of each word in two directions, as it uses a bidirectional transformer algorithm that considers the long-range dependencies between all words in any sentence.
This makes BERT able to understand the word embedding values better  compared with traditional word embedding techniques like Word2Vec and GloVe \citep{Pennington2014}. 
BERT was pre-trained on 3.3 billion tokens \citep{Clark2019does}.
 
\end{enumerate}

In additon, the traditional TF-IDF \citep{tfidf} text representations are produced and included in the study for comparison. 
 
\subsection{Feature Engineering}
\label{SECTION4.3}

All WE algorithms are trained to extract the embedded meaning of the text and represent it as a numerical vector for either the word or sentence or the whole document. Some methods, including Word2Vec, GloVe, FastText and BERT extract the embedded vector for each word, while others such as Doc2Vec can extract one embedded vector for the entire document.
Using a WE algorithm for the entire text document makes the extracted vector potentially ready for machine learning models, as each sample row has one numerical vector of attributes. However, using WE algorithms that extract word-by-word vectors creates a challenge as to a suitable mathematical method to combine the vectors and create a lower-dimensional set of features for the whole document.

A frequently applied approach is to obtain the average vector over all terms in a document. So, if $v_1,\ldots,v_{n_d}\in R^m$ are term vectors for document $d$, then $x_d = (v_1+\ldots+v_{n_d})/n_d$ represents its feature vector, where $n_d$ is the number of terms in document $d$ and $m$ is the dimension of  the word embedding vectors. Here, each term contributes equally into the resulting vector $x_d$.

Another possible approach is to employ Principal Component Analysis (PCA) \citep{multstats} as a dimension reduction and feature extraction method. We consider several PCA-based ways to represent features  for various word embeddings, as detailed below.

\subsubsection{PCA for Word2Vec and FastText}
\label{pca1-wf}

Word2Vec and FastText produce separate embedding vectors for each term in a document. In this case, the PCA method will be used as a way of modifying term-related weights in the average vector $x_d$. Namely, the feature vector for document $d$ will be in the form $w_1v_1+\ldots+w_{n_d}v_{n_d}$, where $w_1,\ldots,w_{n_d}$ are weights related to terms in the document. The weights are obtained from the PCA method by taking the first principal component. Therefore, this PCA-based feature representation is obtained in two steps. First, for each document $d$, term vectors $v_1,\ldots,v_{n_d}\in R^m$ are obtained and placed in a matrix $A_d = [v_1,\ldots,v_{n_d}]\in R^{m\times n_d}$ with each column corresponding to one term. The second step is to apply the PCA to this matrix, then select the first principal component, which, mathematically, is a weighted average of the column vectors. Note that here the PCA method is performed separately for each document $d$.

To determine a better approach to feature engineering for Word2Vec and FastText, we compare taking the arithmetic mean of vectors, which is the more popular practice, to applying PCA which attempts to maximise the amount of information extracted from the terms' vectors. 
This will produce four text representations, which we denote in this study as W2V-Average, W2V-PCA, FT-Average, FT-PCA, for Word2Vec and FastText word embeddings, respectively. For Word2Vec, the dimension $m$ of word embedding vectors and therefore the dimension of feature vectors $x_d$, is equal to $100$ and for FastText $m=300$.

\subsubsection{PCA for BERT}

For BERT, we consider another PCA-based feature extraction approach consisting of the following steps: (1) apply the mean pooling to the individual term vectors to obtain one vector representing each document; (2) use PCA to reduce the dimension of the  document vectors resulting from the above step.

Here, the PCA technique is applied to reduce the dimension of the averaged vectors for all documents together, rather than to combine the vectors of individual terms separately for each document. As a result, the dimension of BERT embeddings is reduced from 384 to 50.

Two text representations are compared: BERT-Average and BERT-PCA, where the BERT-Average dataset is produced using the function \texttt{mean\_pooling} from \texttt{transformers} library, which computes the weighted averages of 384-dimensional word embedding vectors for each document and BERT-PCA is produced by applying PCA dimensionality reduction to the BERT-Average vectors as described above. 

\subsubsection{PCA for Doc2Vec}

We also invetigate an application of PCA to Doc2Vec document vectors, reducing the features' dimension from 300 to 100, to examine its usefulness in comparison to taking the entire document vector. The feature representations obtained from the above procedures are denoted in this study  as Doc2Vec and Doc2Vec-PCA, respectively.

\subsection{Study Design}
\label{SECTION4.4}

\begin{figure}[!htbp] 
\begin{center}
\includegraphics[height = 13cm]{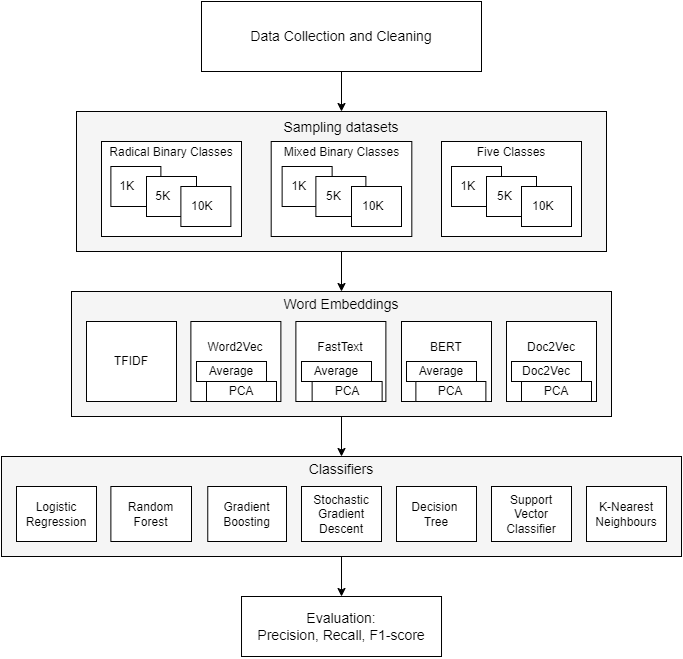}
\end{center}
\caption{The pipeline of the empirical study.}
\label{pipeline}
\end{figure}

The study in organised in the following general steps: (1) creating different datasets, (2) applying feature extraction techniques with word embeddings, (3) applying different classifiers to the obtained features and (4) comparing time consumed and performance metrics for the resulting features and classifiers. Figure \ref{pipeline} shows the pipeline of the study and the details are described in the following paragraphs. 

\subsubsection{Step 1: Creating the different datasets}
\label{ch4-data}

The starting dataset consisted of online customer reviews for 29 telecommunication brands in the UK, scraped from the Trustpilot platform and contained 367,000 reviews.  
Each review was combined with a star rating, ranging between 1 and 5 stars, given by the reviewer to produce a labeled dataset suitable to be used to build a text classifier. 
Text cleaning tools were applied to to the dataset to remove all unwanted characters, make all letters small and remove non-English letters and words.

Next, three types of datasets were created using random sampling:
\begin{itemize}
    \item {\bf Radical-Binary Data} consisting of reviews with 1-star, labeled as "Bad" and 5-star, labeled as "Good", in equal proportions. Three sets of sample size 1000, 5000 and 10,000, respectively, were randomly selected.
    \item {\bf Mixed-Binary Data} is a dataset with two classes "Bad" and "Good", where the class "Bad" consists of a mixture of 1-star and 2-star reviews. The class "Good" consists of reviews that have a ranking value of 4 or 5 stars. The aim of studying this dataset is to introduce less obvious patterns in reviews for a more challenging classification task. Three sets of size 1000, 5000 and 10,000, respectively, were randomly selected.
    \item {\bf Multi-Class Data} with five classes was created by selecting an equal number of reviews for each star ranking, from 1 to 5. This set of data is used to apply multiclass classification and to explore the performance of word embeddings and classifiers in a more challenging scenario. Three sets of size 1000, 5000 and 10,000, respectively, were randomly selected.
\end{itemize}

\noindent

The created datasets consisted of two columns: Text and Type, where the Text column contained the cleaned customer's review and the Type column contained either a binary class label ("Good" or "Bad") in the case of the binary class datasets or the qualitative label for the number of stars ranging from 1-star to 5-stars in the case of the multiclass datasets.

\subsubsection{Step 2: Feature extraction techniques}
\label{features}

 Feature extraction algorithms convert the raw text into represented numerical values, that aim to reflect its meaning. In this study, nine word-embedding-based techniques were explored, as  described in section \ref{SECTION4.3}:
W2V-Average, W2V-PCA, FT-Average, FT-PCA, BERT-Average, BERT-PCA, Doc2Vec, Doc2Vec-PCA and TF-IDF.

These word embedding techniques were applied to the nine datasets producing different text representations of each one of them, that will be used with classifiers.

\subsubsection{Step 3: Classifiers}
\label{classifiers1}

We employed the following seven classification algorithms for each dataset to predict the star rating based on the customer's written text: Logistic Regression, Random Forest, Gradient Boosting, Stochastic Gradient Descent, Decision Tree, Support Vector Machine and K-Nearest Neighbours (KNN). These classifiers were imported in Python language from the Sklearn library and trained using the parameters' settings detailed in Appendix \ref{appendix4}.

\subsubsection{Step 4: Evaluation}

The comparison of word embeddings and classifiers focused on analysis of time consumption and classifiers' performance and included various processing set-ups.

To assess one of the practical aspects of the methods considered, the time taken to extract the features using WEs and time to train the classifier was measured. 
To measure the time taken in parts of the training, the time function from Python was used. Each time measurement was repeated 5 times and the average reported to decrease any possible bias.

Moreover, to evaluate the classifiers' performance, precision, recall and F1-Score \citep{Powers2020Evaluation} were computed. The evaluation measures used in this experiment are macro averages calculated as the unweighted mean of precision, recall and F1-Score for all classes in the outcome variable given by the formulae below:
\[
\text{Macro-Average-Precision} = \frac{1}{K} \sum_{i=1}^K \text{Precision}_i = \frac{1}{K} \sum_{i=1}^K \frac{\text{TP}_i}{\text{TP}_i + \text{FP}_i},
\]
\[
\text{Macro-Average-Recall} = \frac{1}{K} \sum_{i=1}^K \text{Recall}_i = \frac{1}{K} \sum_{i=1}^K \frac{\text{TP}_i}{\text{TP}_i + \text{FN}_i},
\]
\[
\text{Macro-Average-F1-Score} = \frac{1}{K} \sum_{i=1}^K \text{F1-Score}_i = \frac{1}{K} \sum_{i=1}^K \frac{2 \cdot \text{Precision}_i \cdot \text{Recall}_i}{\text{Precision}_i + \text{Recall}_i},
\]
\noindent
where \( K \) is the number of classes, \( \text{TP}_i \) is the number of True Positives for class \( i \), \( \text{FP}_i \) is the number of False Positives for class \( i \), \( \text{FN}_i \) is the number of False Negatives for class \( i \). 
The above evaluation measures were calculated using five-fold cross-validation and the {\it classification\textunderscore report} method from the Sklearn library in Python was employed.

In this study, we used the Google Colab environment\footnote{https://colab.research.google.com/} to run the Python code for all processes, 
which is a cloud platform widely used in machine learning and deep learning applications. See \cite{carneiro2018performance} for an early discussion of it. The use of this environment allowed us to study and compare the performance of the algorithms on the CPU and GPUs. In particular, for the BERT word embedding algorithm, in most cases the code would not run with the standard memory options for the CPUs and GPUs and so the HighRAM option needed to be employed.
CPU-HighRAM and T4-GPU-HighRAM processing sets were tested, by measuring the feature extraction time for each word embedding and the training time for each classifier. Furthermore, when studying the energy consumption and CO2 emissions of word embeddings in subsection \ref{energy}, seven processing sets were considered: T4-GPU-HighRAM, CPU-HighRAM, CPU, T4-GPU, A100-GPU-HighRAM, V100-GPU, V100-GPU-HighRAM.


\section{Results}
\label{ResultsSection}

In this section, the results of experiments are presented and analysed. Subsection \ref{featuretimes} describes the results for the comparison between word embeddings with respect to the time consumed by the feature extraction algorithm. Two processing sets are considered CPU-HighRAM and T4-GPU-HighRAM. Subsection \ref{classifierstime} is focused on classifiers' training times and comparisons between classifiers and between word embeddings are made. In subsection \ref{classifiersperf}, the results on classifiers' performance metrics are analyzed. This includes comparisons between word embeddings (subsection \ref {Comparisonbetweenwordembeddings}), comparisons between the Average and PCA approaches to feature extraction (subsection \ref{ComparisonbetweenaveragesandPCAforfeatureselection}) and the overall comparison between classifiers (subsection \ref{Comparisonbetweentheclassifiers}). In subsection \ref{Time-Accuracy} the patterns for relationships between the consumed time and the classifiers' performance are discussed.
Finally, consumed energy and produced CO2 emissions are discussed in subsection \ref{energy}.

\subsection{Feature Extraction Times}
\label{featuretimes}

\begin{figure}[!htbp] 
\centering

\begin{subfigure}[b]{1\textwidth}
\centering
\includegraphics[width = 0.48\textwidth]{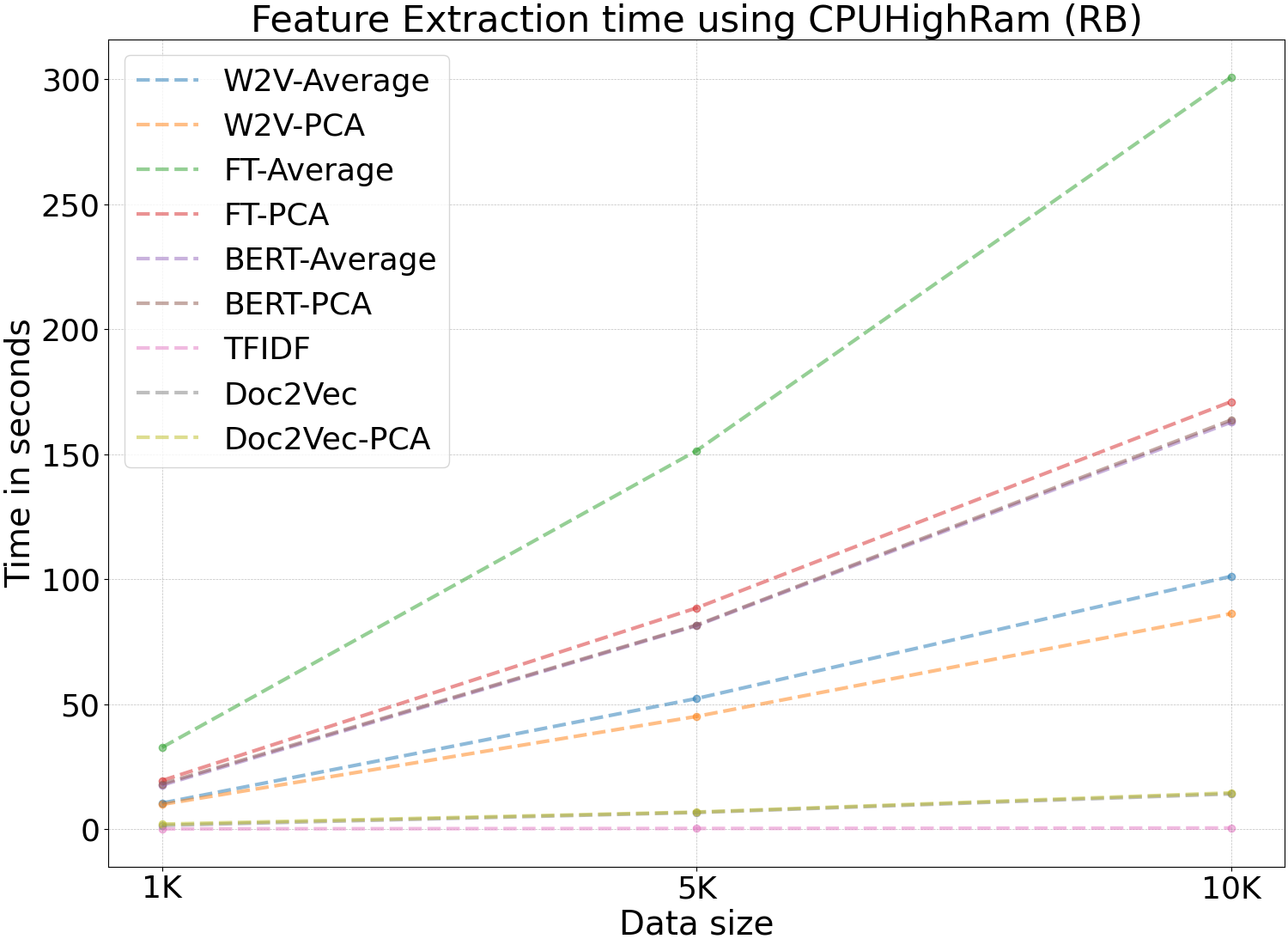}
\includegraphics[width = 0.48\textwidth]{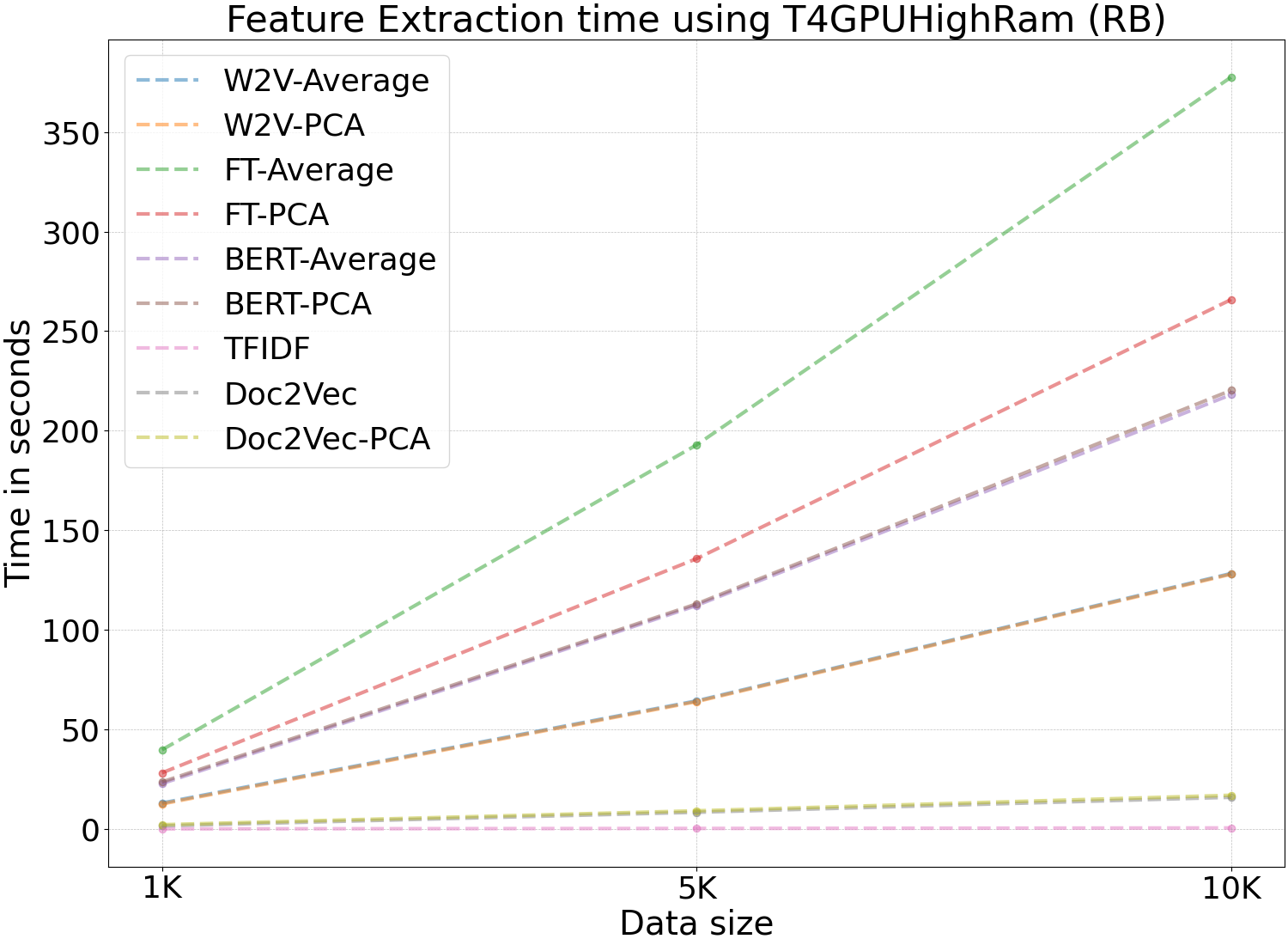}
\caption{Radical Binary data}
\end{subfigure}
\begin{subfigure}[b]{1\textwidth}
\centering
\includegraphics[width = 0.48\textwidth]{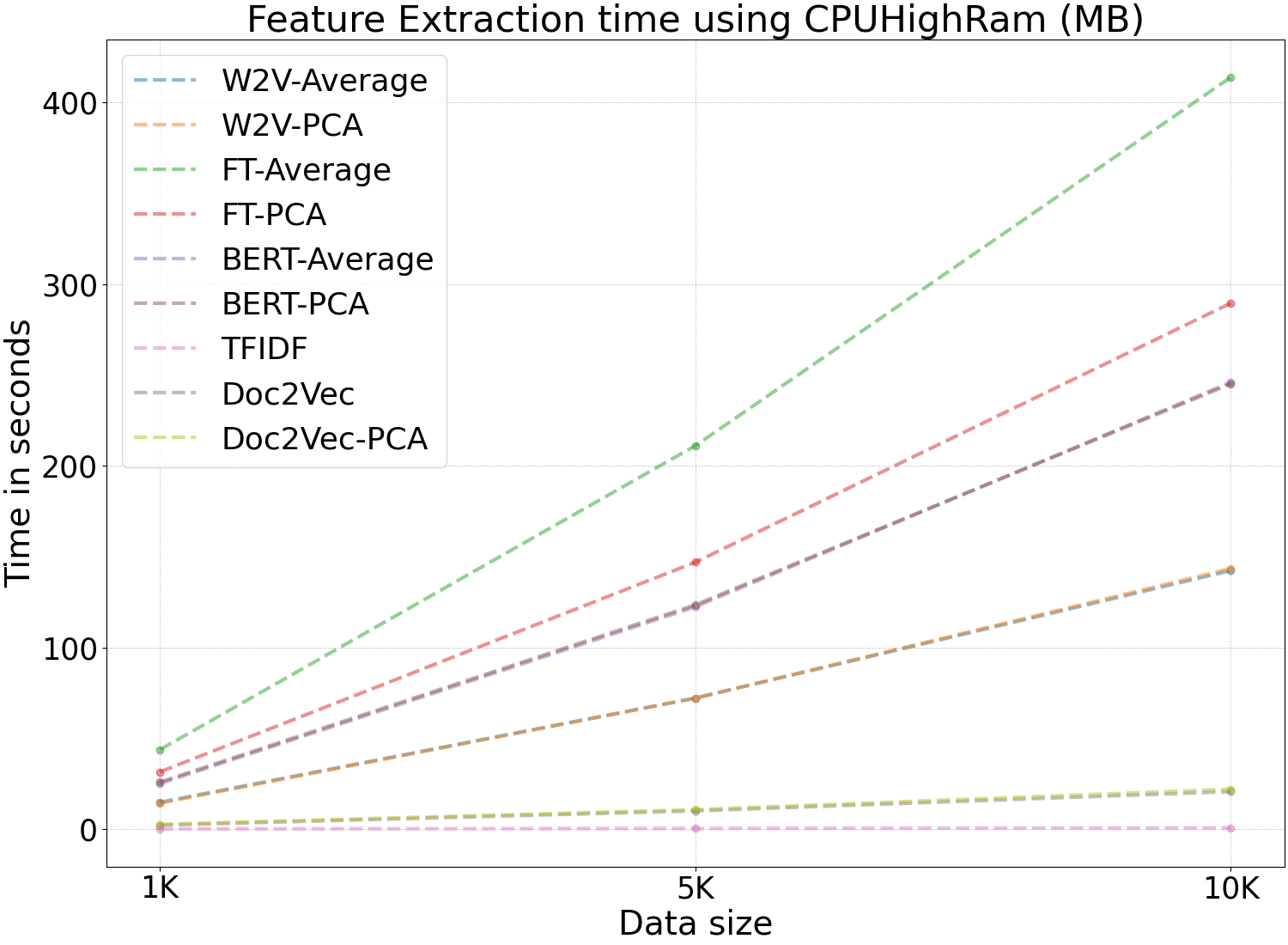}
\includegraphics[width = 0.48\textwidth]{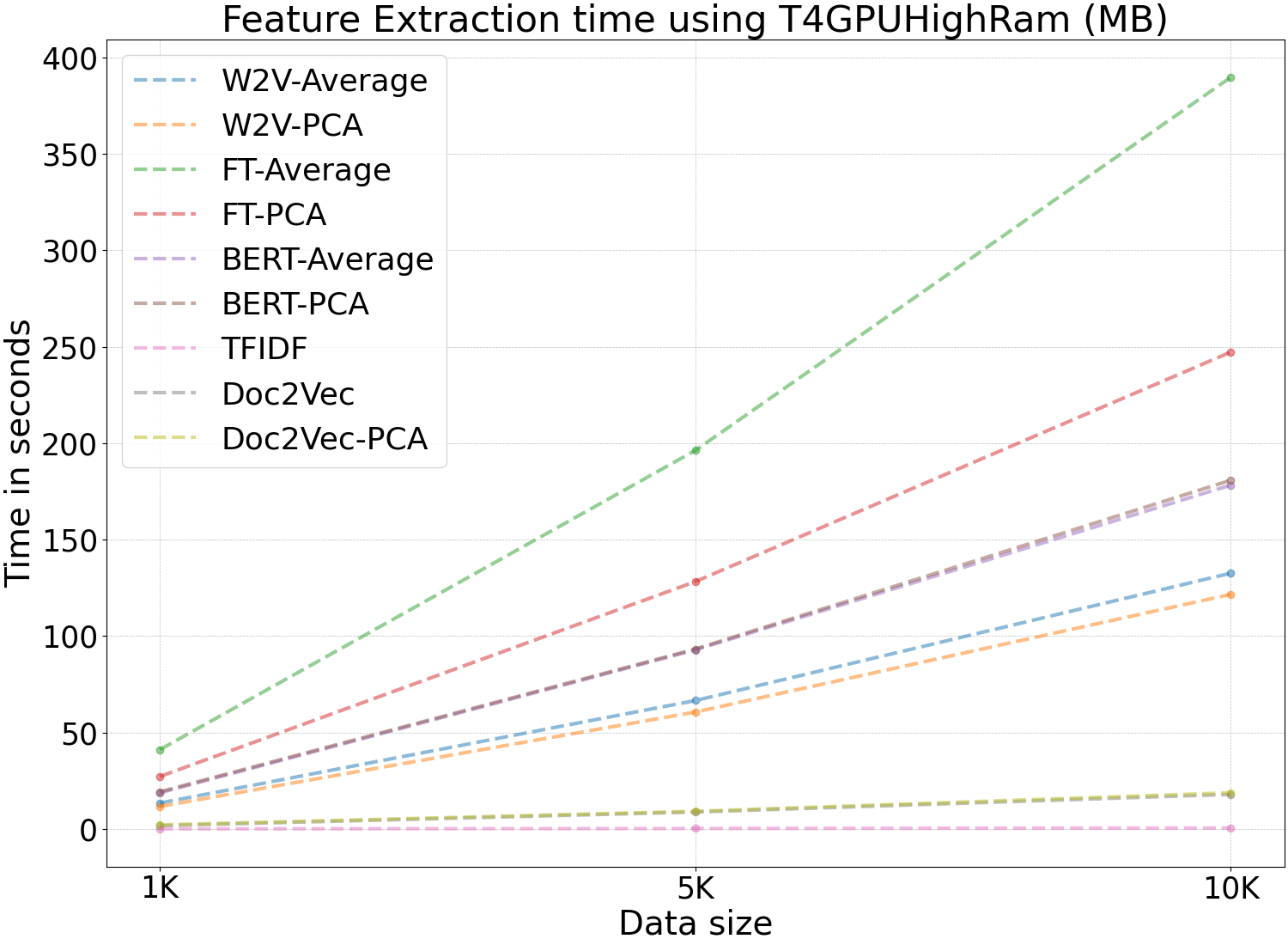}
\caption{Mixed Binary data}
\end{subfigure}
\begin{subfigure}[b]{1\textwidth}
\centering
\includegraphics[width = 0.48\textwidth]{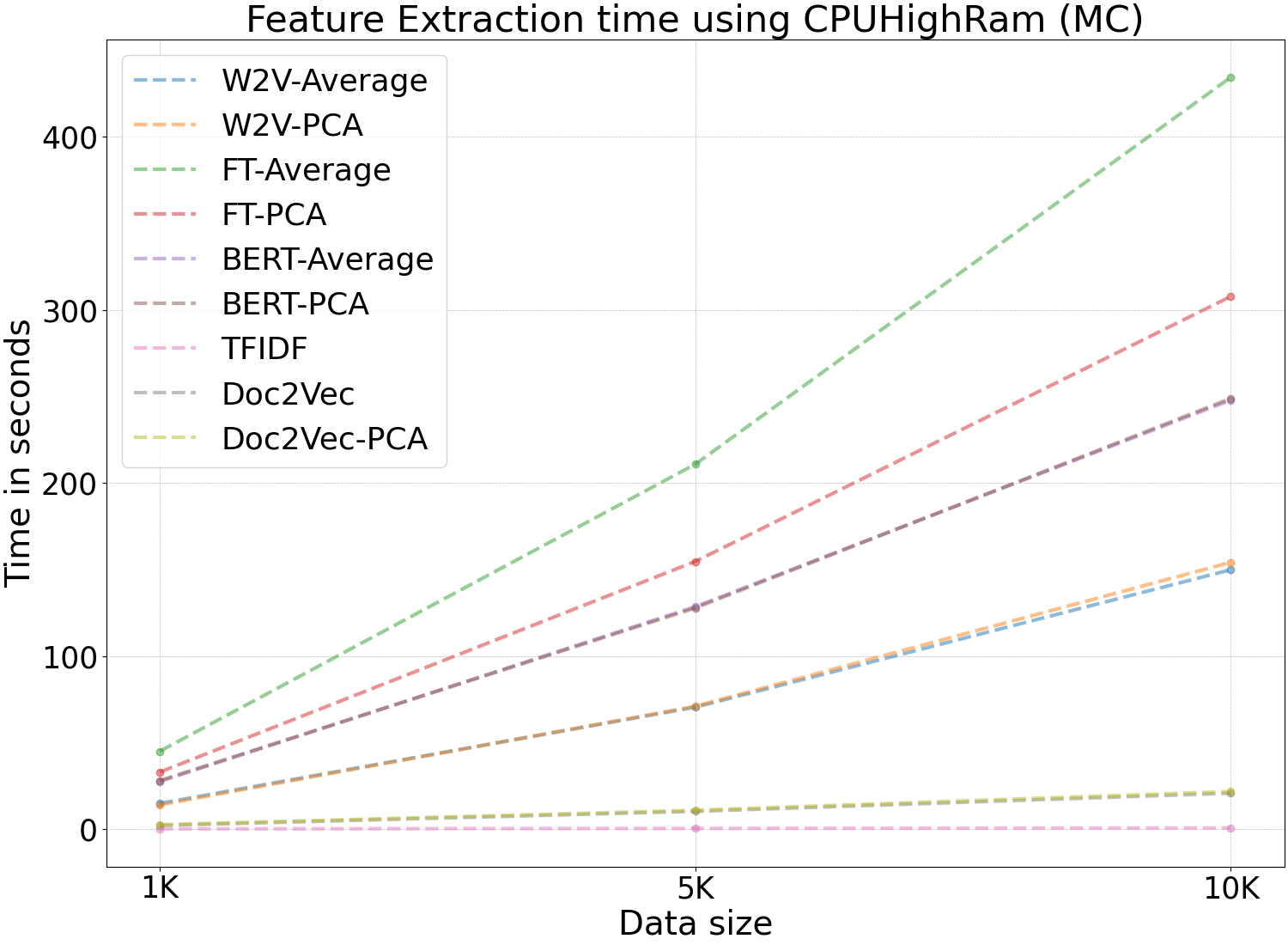}
\includegraphics[width = 0.48\textwidth]{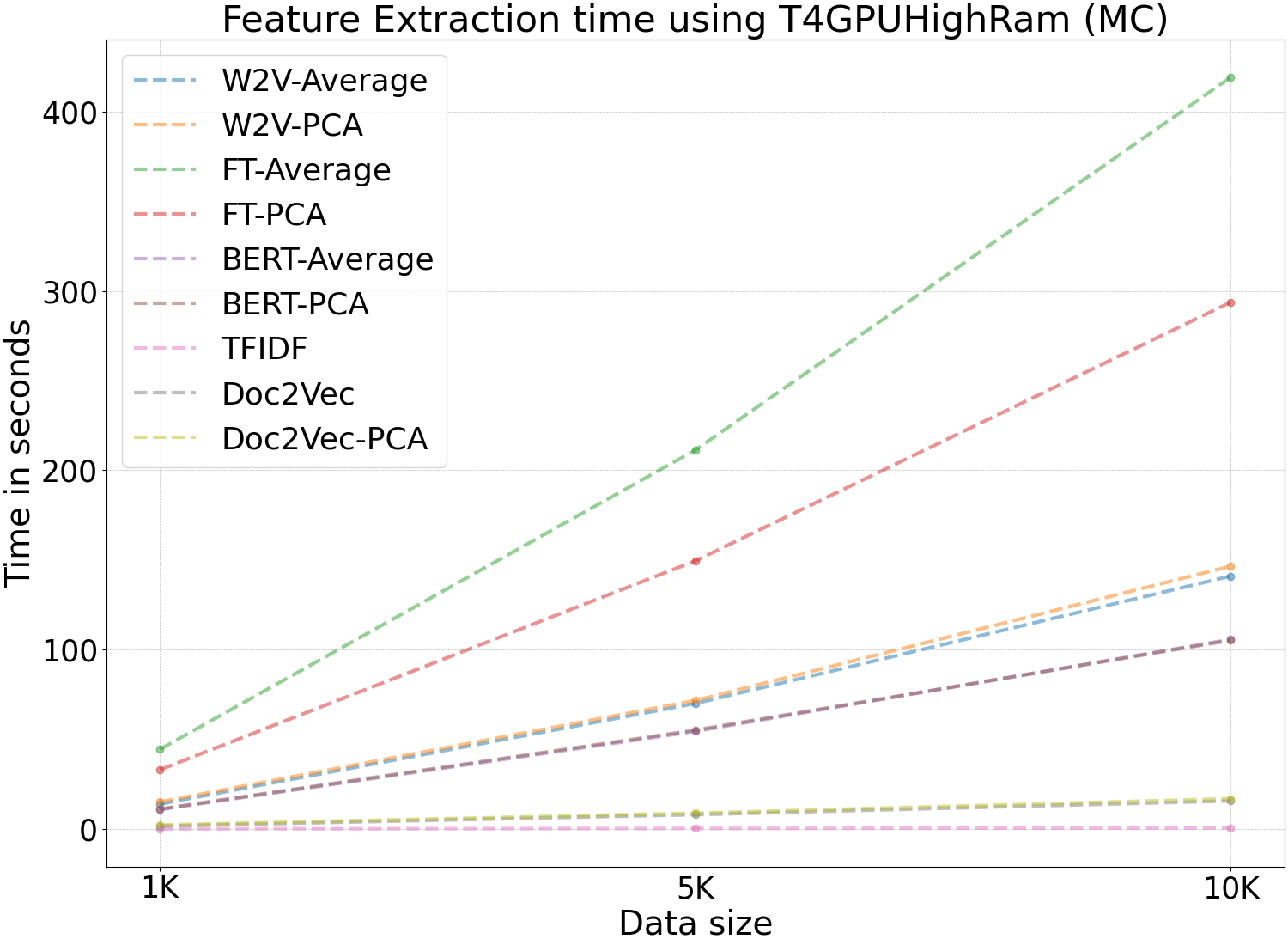}
\caption{Multi-Class data}
\end{subfigure}
\caption{Feature extraction time versus sample size for 1K, 5K and 10K sample sizes,  processed by CPU-HighRAM (left panel) and T4-GPU-HighRam (right panel) for (a) Radical Binary data, (b) Mixed Binary data and (c) Multi-Class data.  }
\label{Featureextractiontime}
\end{figure}

The results of feature extraction times for the nine scenarios of text representations and two processing settings are visualised in Figure \ref{Featureextractiontime}. 

Unsurprisingly, the TF-IDF method needed the least amount of time. Doc2Vec was consistently the second fastest word embedding. Both, Word2Vec and BERT required considerably longer time than Doc2Vec and their relative speeds varied depending on the processor and dataset, with Word2Vec being faster than BERT in all cases except two. FastText word embeddings needed the longest time in every scenario and usually required roughly 400 times more time than TF-IDF and 20 times more than Doc2Vec. Similar patterns held for all three datasets, processor and memory settings and sample sizes: 1000, 5000 and or 10,000.

Studying the effect of increasing the sample size on the feature extraction time, as visualised in Figure \ref{Featureextractiontime}, TF-IDF and Doc2Vec had a relatively constant feature extraction time for the three dataset sizes, which means that increasing the data size five- and ten-fold had a negligible effect on the needed time. On the other hand, FastText's time increased with the sample size at the steepest rate among all other word embedding types, followed by BERT and Word2Vec's times which increased at a slower linear rate. The increase rates were approximately linear in each case.

When comparing the results for the two processing powers, in most cases CPU-HighRAM needed slightly more time than T4-GPU-HighRAM and the increase typically ranged from 1\% to 10\%. However, for the Radical Binary dataset, the opposite effect occurred and CPU-HigRAM was slightly faster than T4-GPU-HighRAM. Generally, the processing power appeared to play an important role for BERT, where the biggest differences can be observed as in some cases CPU-HighRAM needed almost double the time than T4-GPU-HighRAM for BERT embeddings.

\subsection{Classifiers' Training Times}
\label{classifierstime}

\begin{figure}[!htbp] 
\centering

\begin{subfigure}[b]{1\textwidth}
\centering
\includegraphics[width = 0.49\textwidth]{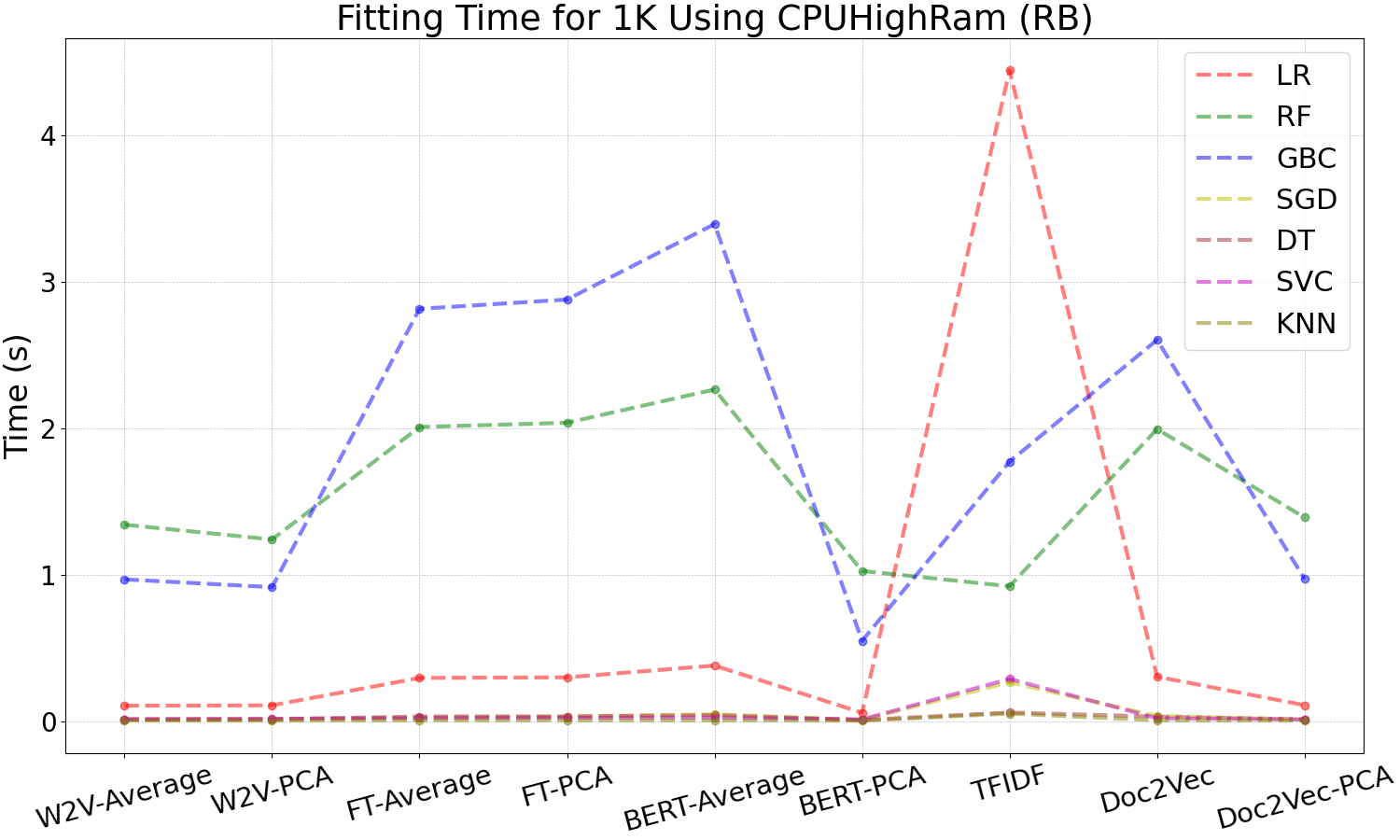}
\includegraphics[width = 0.49\textwidth]{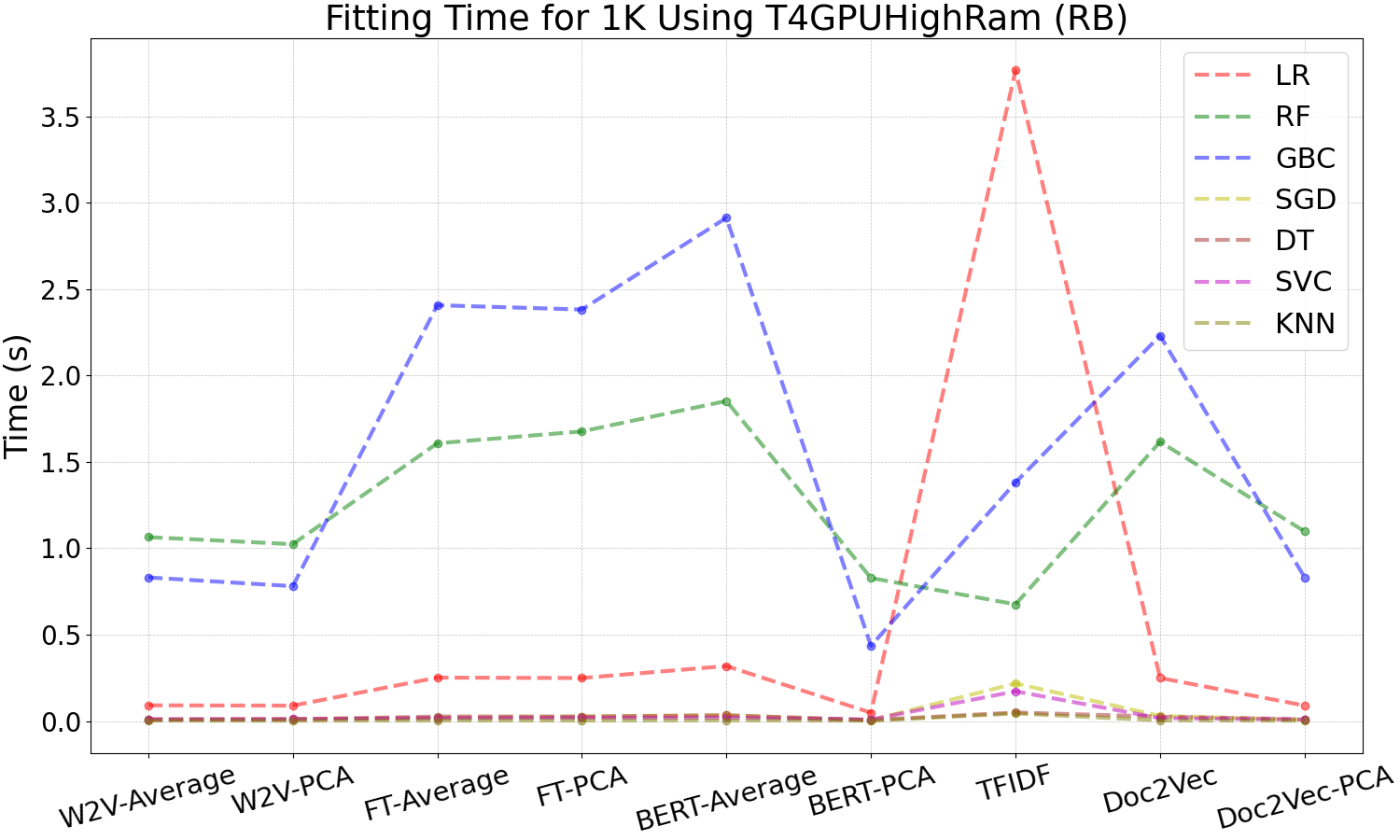}
\caption{Sample size $n=1000$}
\end{subfigure}

\begin{subfigure}[b]{1\textwidth}
\centering
\includegraphics[width = 0.49\textwidth]{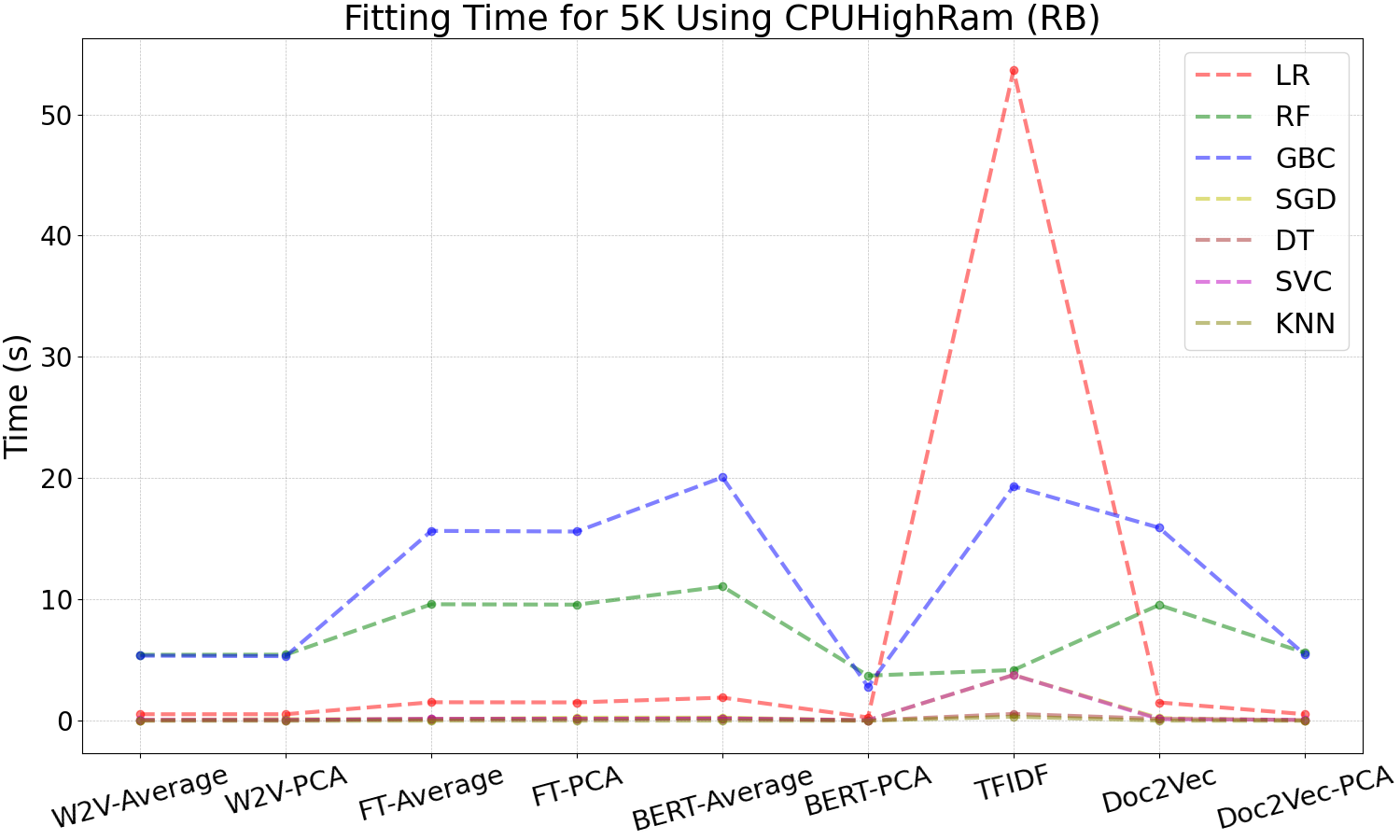}
\includegraphics[width = 0.49\textwidth]{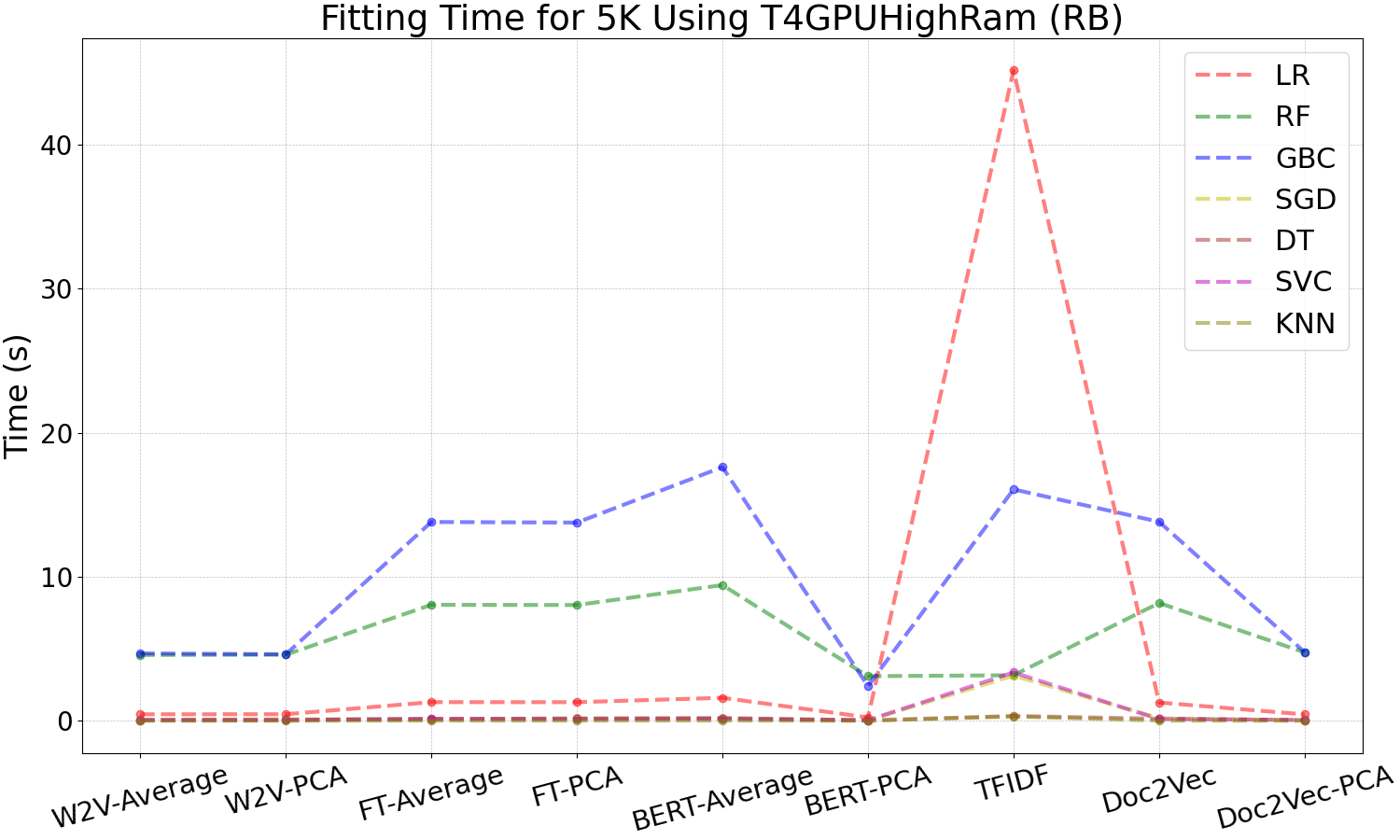}
\caption{Sample size $n=5000$}
\end{subfigure}

\begin{subfigure}[b]{1\textwidth}
\centering
\includegraphics[width = 0.49\textwidth]{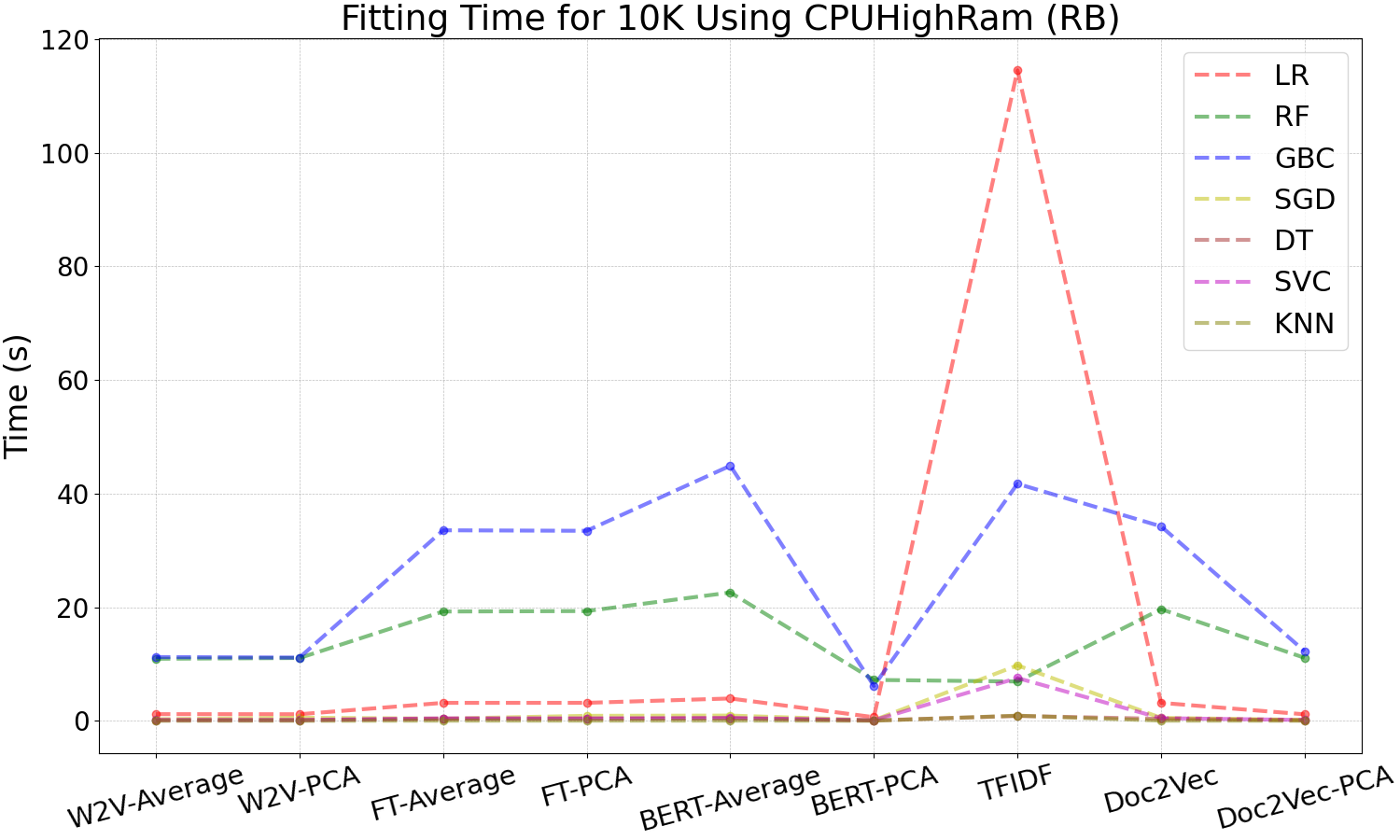}
\includegraphics[width = 0.49\textwidth]{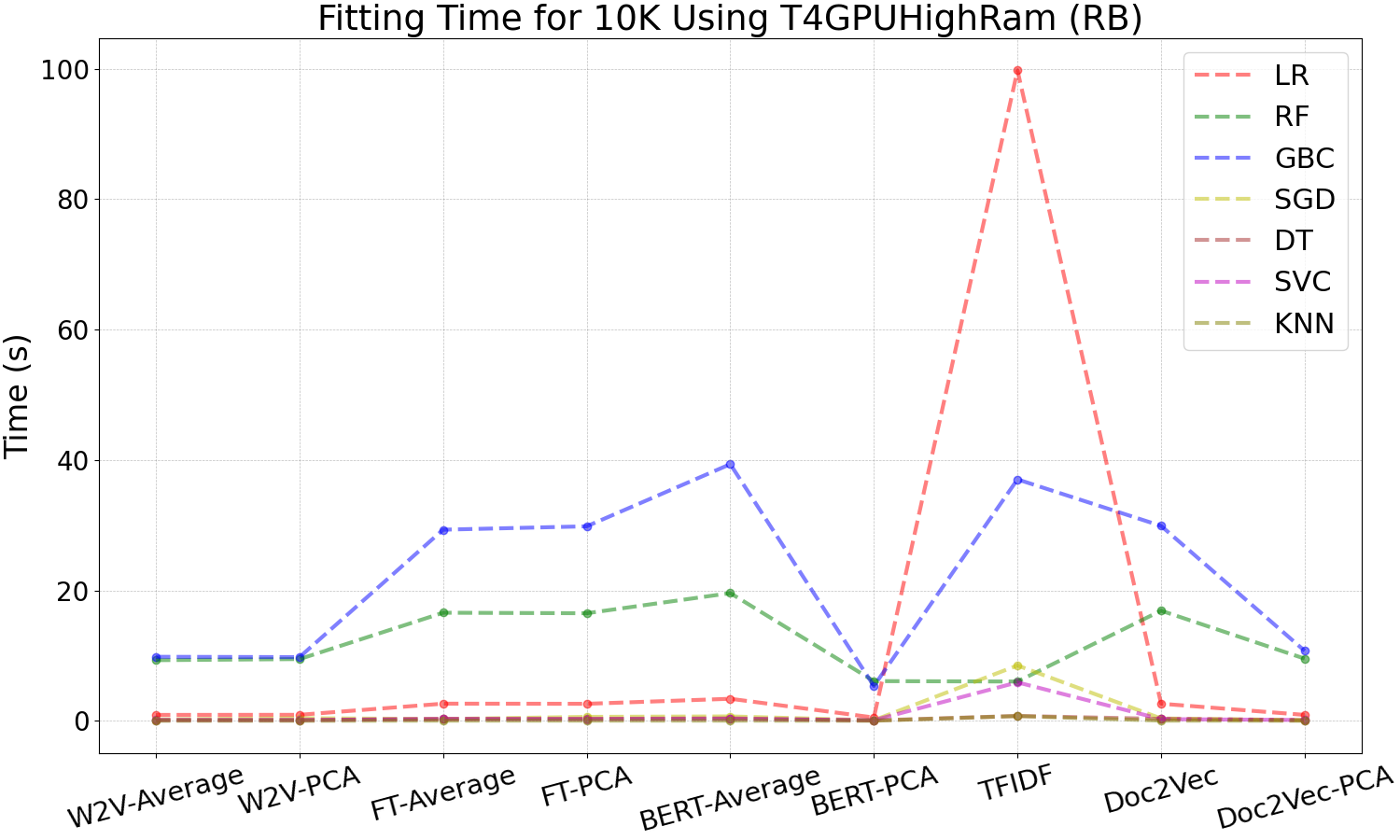}
\caption{Sample size $n=10,000$}
\end{subfigure}

\caption{Fitting time for classifiers for 1K, 5K and 10K sample sizes using CPU-HighRAM (left-hand side panel)  and T4-GPU-HighRAM (right-hand side panel) for nine types of word embedding as indicated on the horizontal axes, for Radical Binary data.}

\label{Fitting1}
\end{figure}

\begin{figure}[!htbp] 
\centering

\begin{subfigure}[b]{1\textwidth}
\centering
\includegraphics[width = 0.49\textwidth]{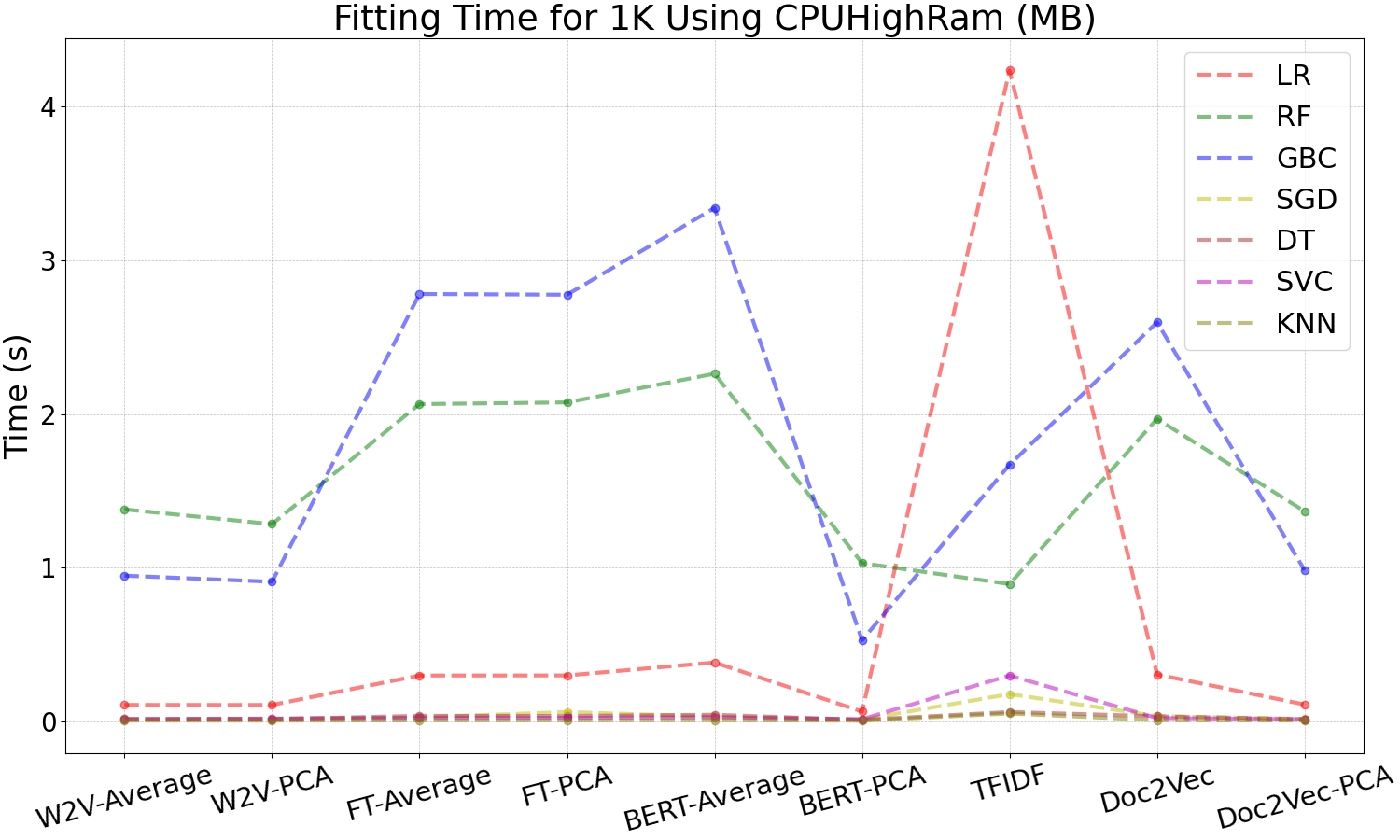}
\includegraphics[width = 0.49\textwidth]{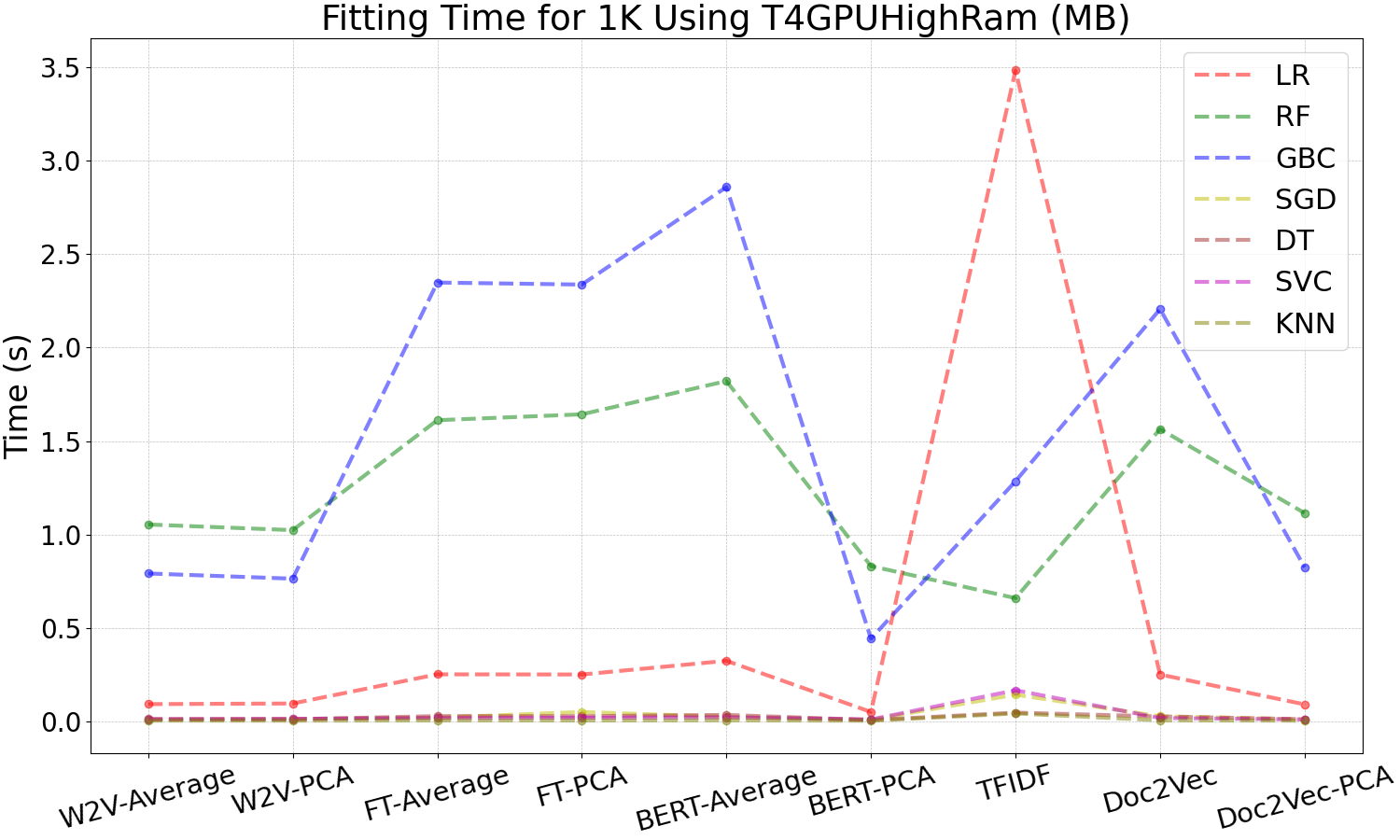}
\caption{Sample size $n=1000$}
\end{subfigure}
\begin{subfigure}[b]{1\textwidth}
\centering
\includegraphics[width = 0.49\textwidth]{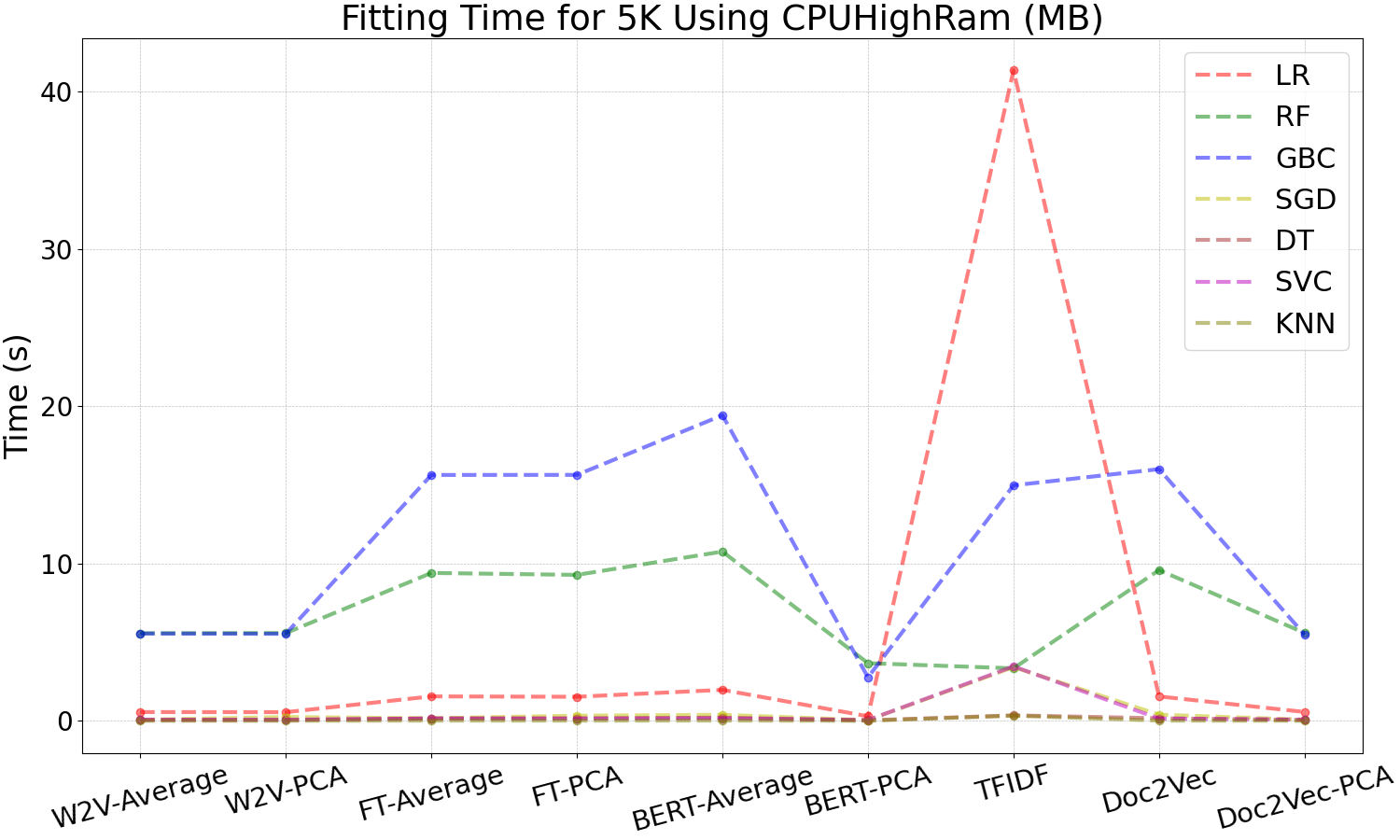}
\includegraphics[width = 0.49\textwidth]{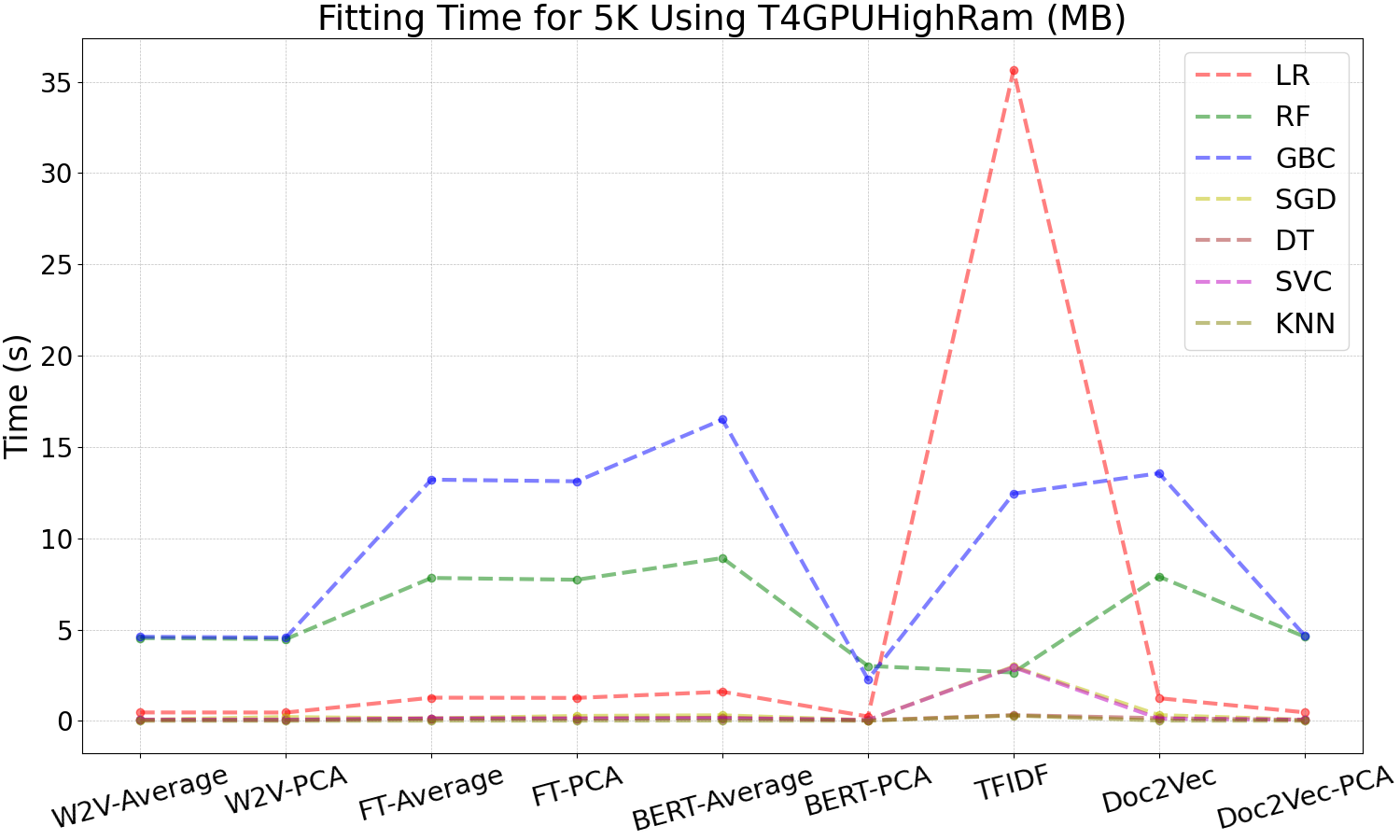}
\caption{Sample size $n=5000$}
\end{subfigure}
\begin{subfigure}[b]{1\textwidth}
\centering
\includegraphics[width = 0.49\textwidth]{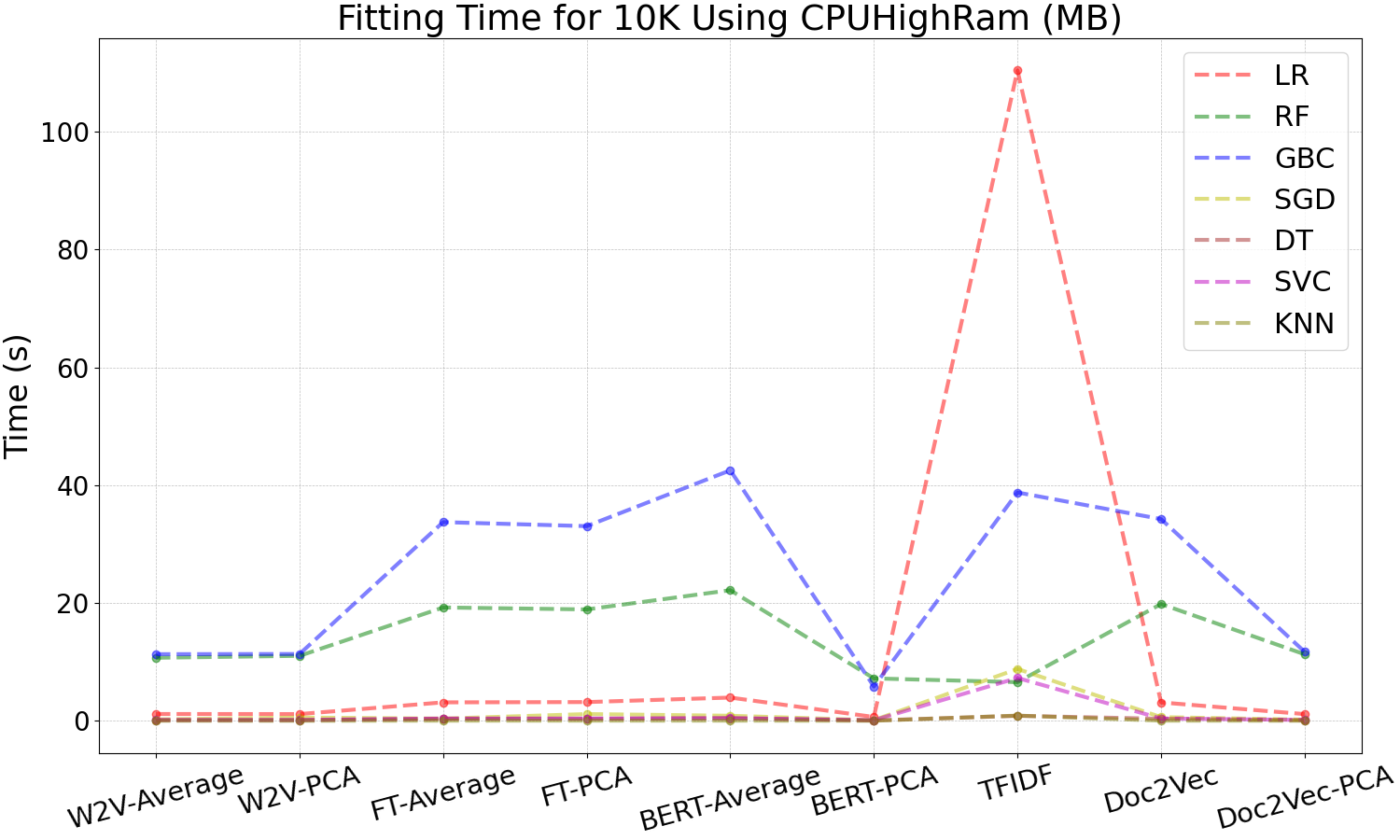}
\includegraphics[width = 0.49\textwidth]{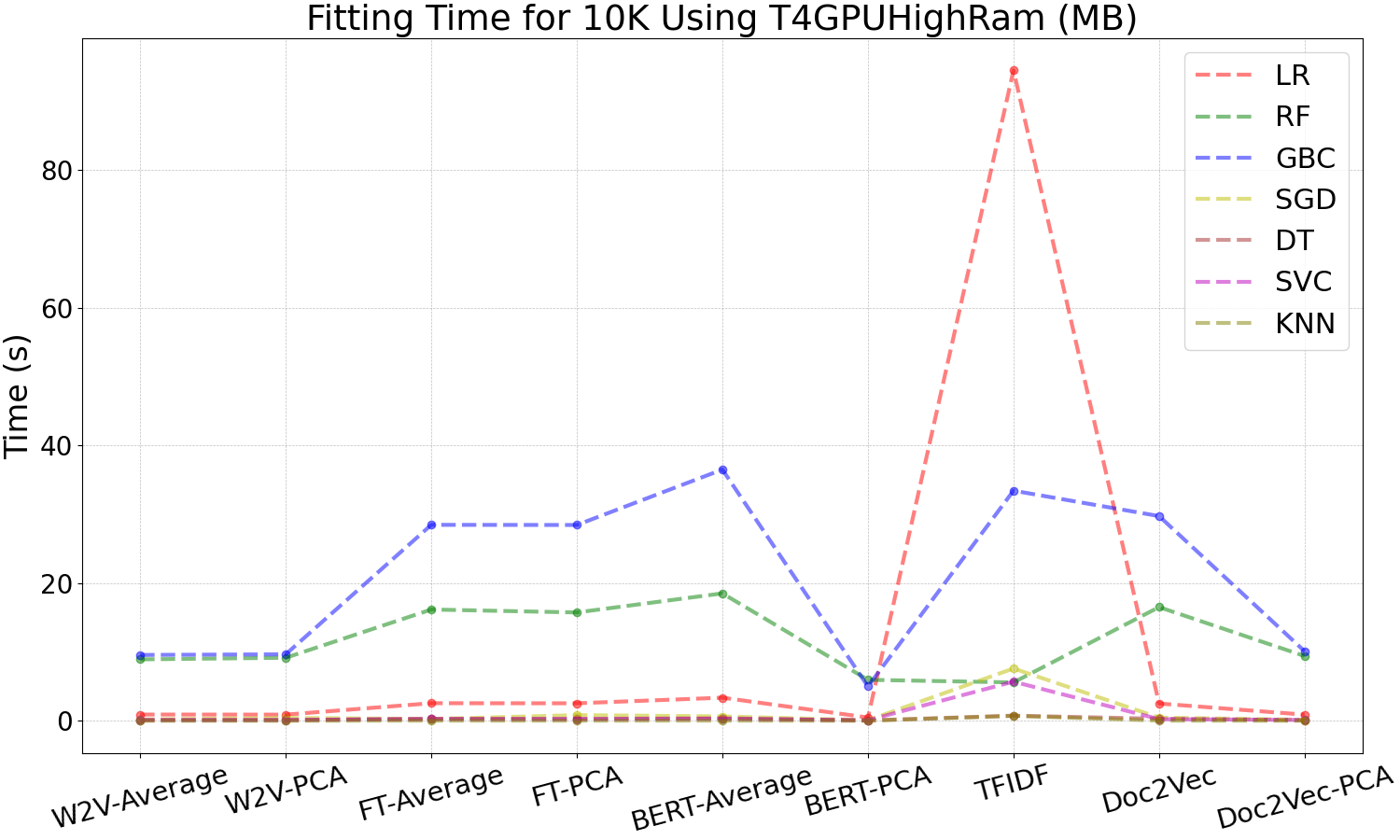}
\caption{Sample size $n=10,000$}
\end{subfigure}

\caption{Fitting time for classifiers for 1K, 5K and 10K sample sizes using CPU-HighRAM (left-hand side panel)  and T4-GPU-HighRAM (right-hand side panel) for nine types of word embedding as indicated on the horizontal axes, for Mixed Binary data.}

\label{Fitting2}
\end{figure}

\begin{figure}[!htbp] 
\centering

\begin{subfigure}[b]{1\textwidth}
\centering
\includegraphics[width = 0.49\textwidth]{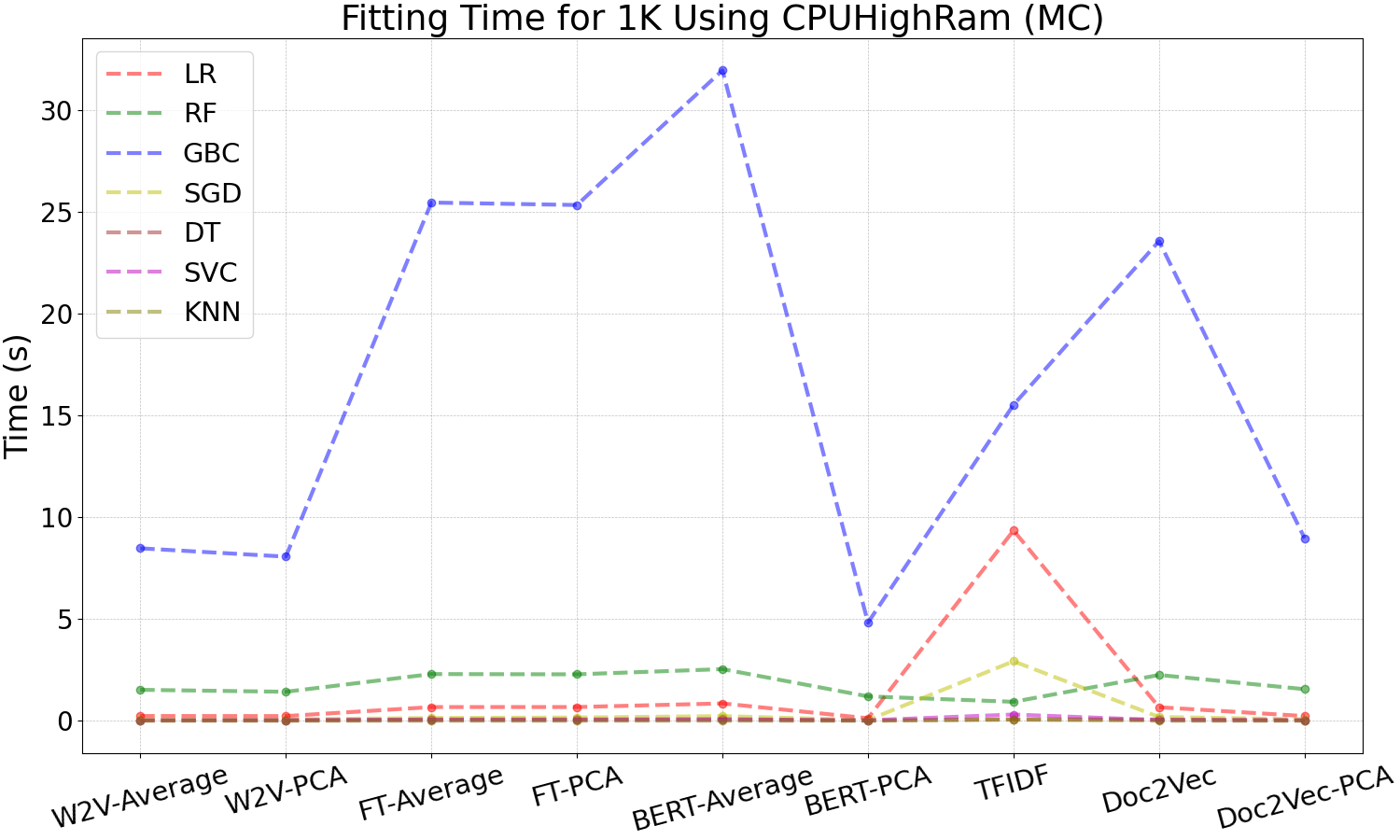}
\includegraphics[width = 0.49\textwidth]{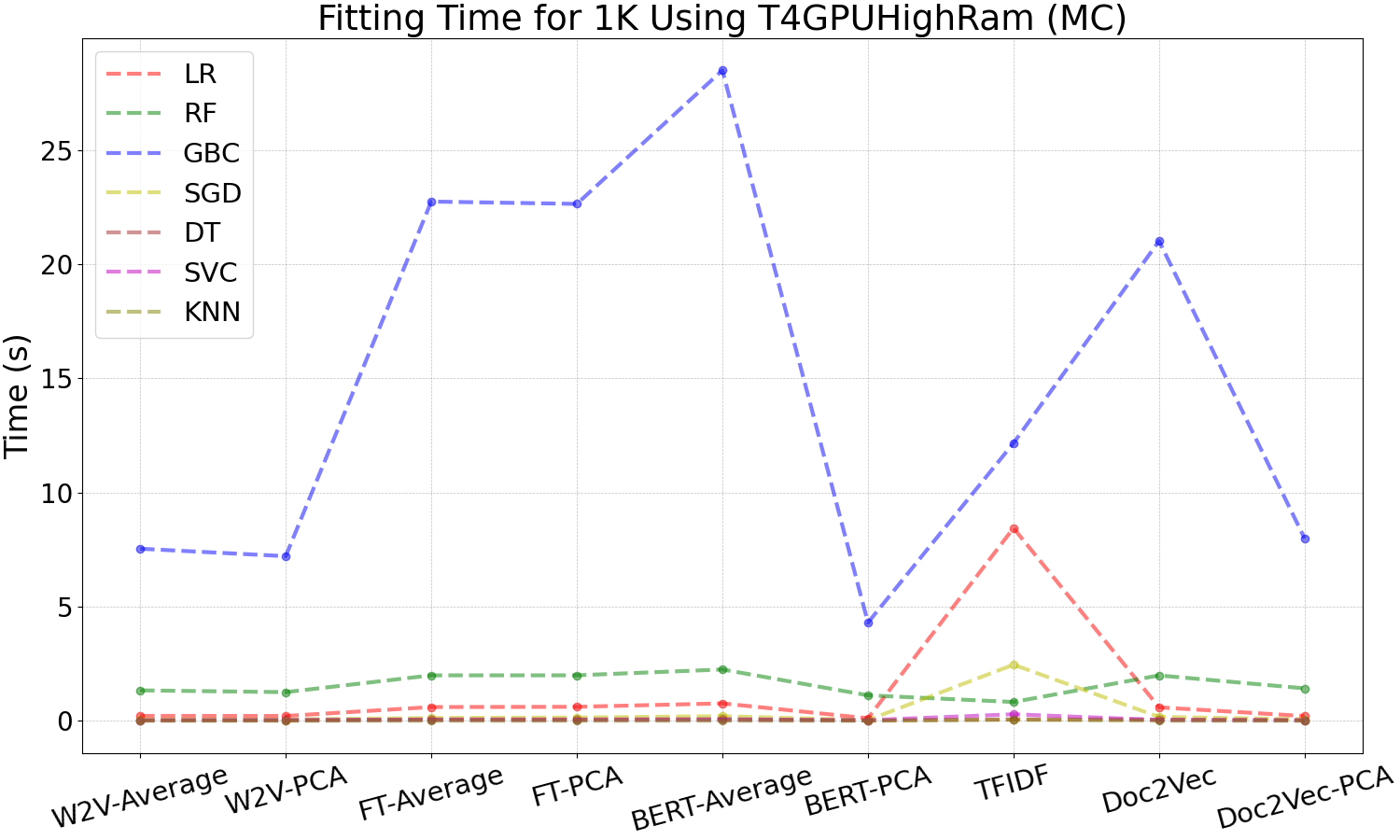}
\caption{Sample size $n=1000$}
\end{subfigure}
\begin{subfigure}[b]{1\textwidth}
\centering
\includegraphics[width = 0.49\textwidth]{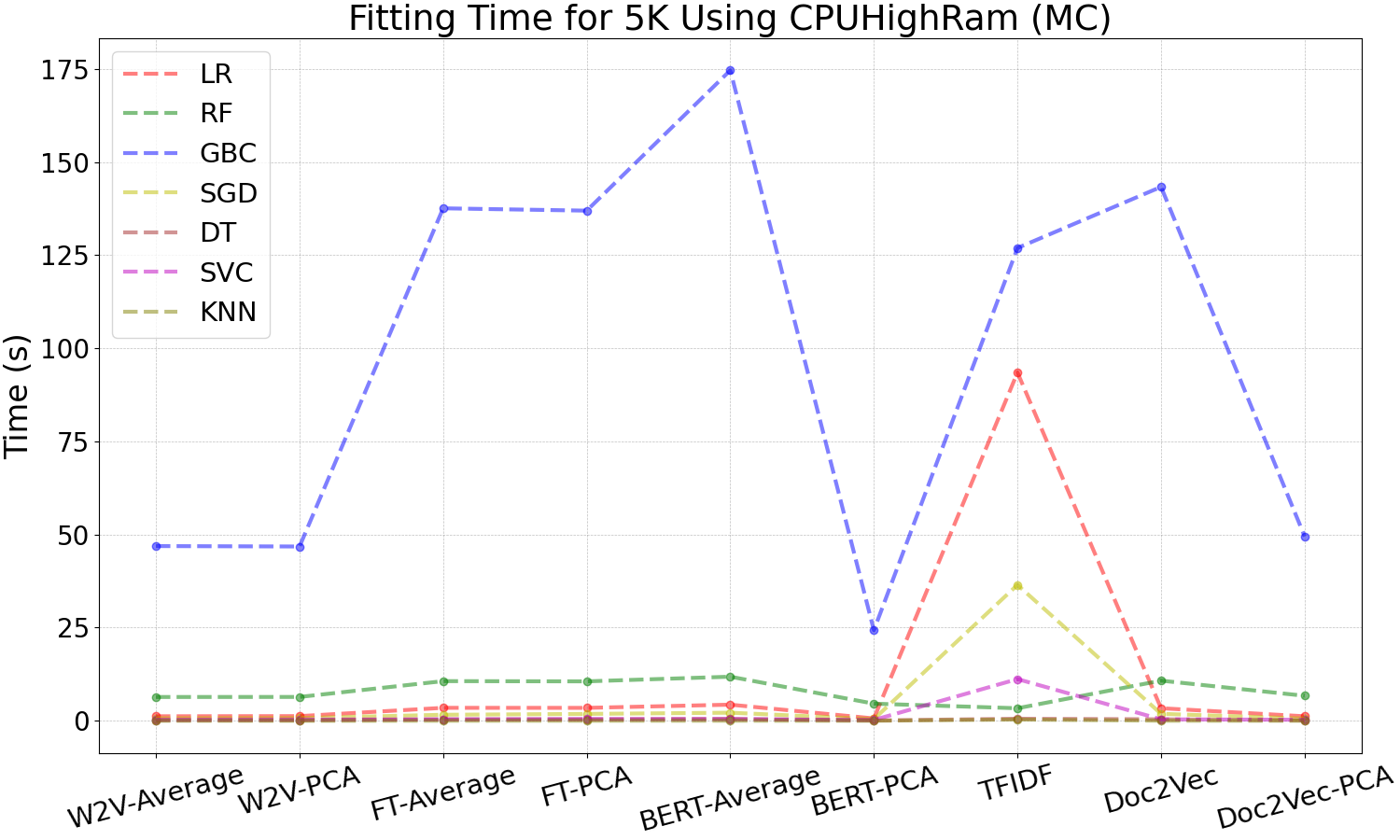}
\includegraphics[width = 0.49\textwidth]{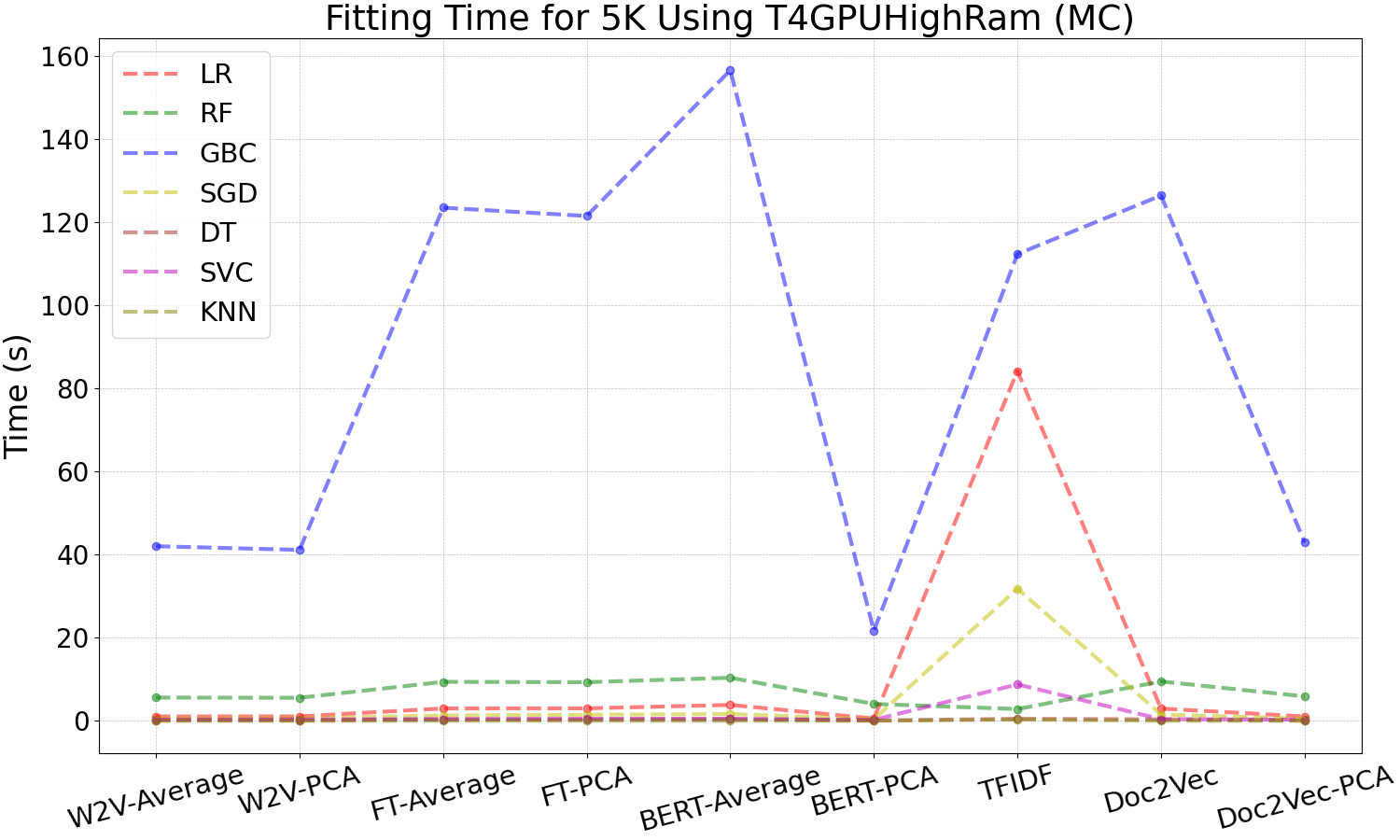}
\caption{Sample size $n=5000$}
\end{subfigure}
\begin{subfigure}[b]{1\textwidth}
\centering
\includegraphics[width = 0.49\textwidth]{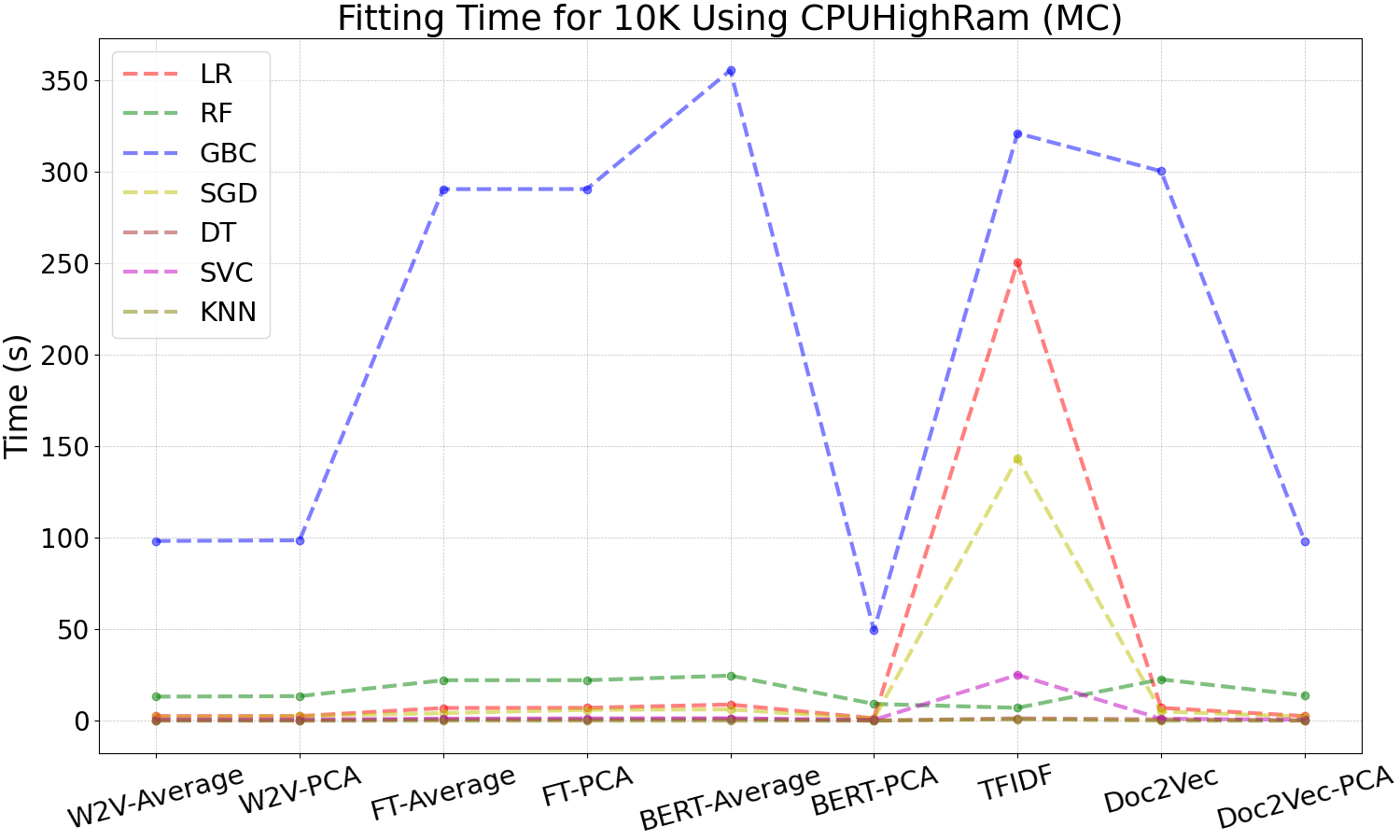}
\includegraphics[width = 0.49\textwidth]{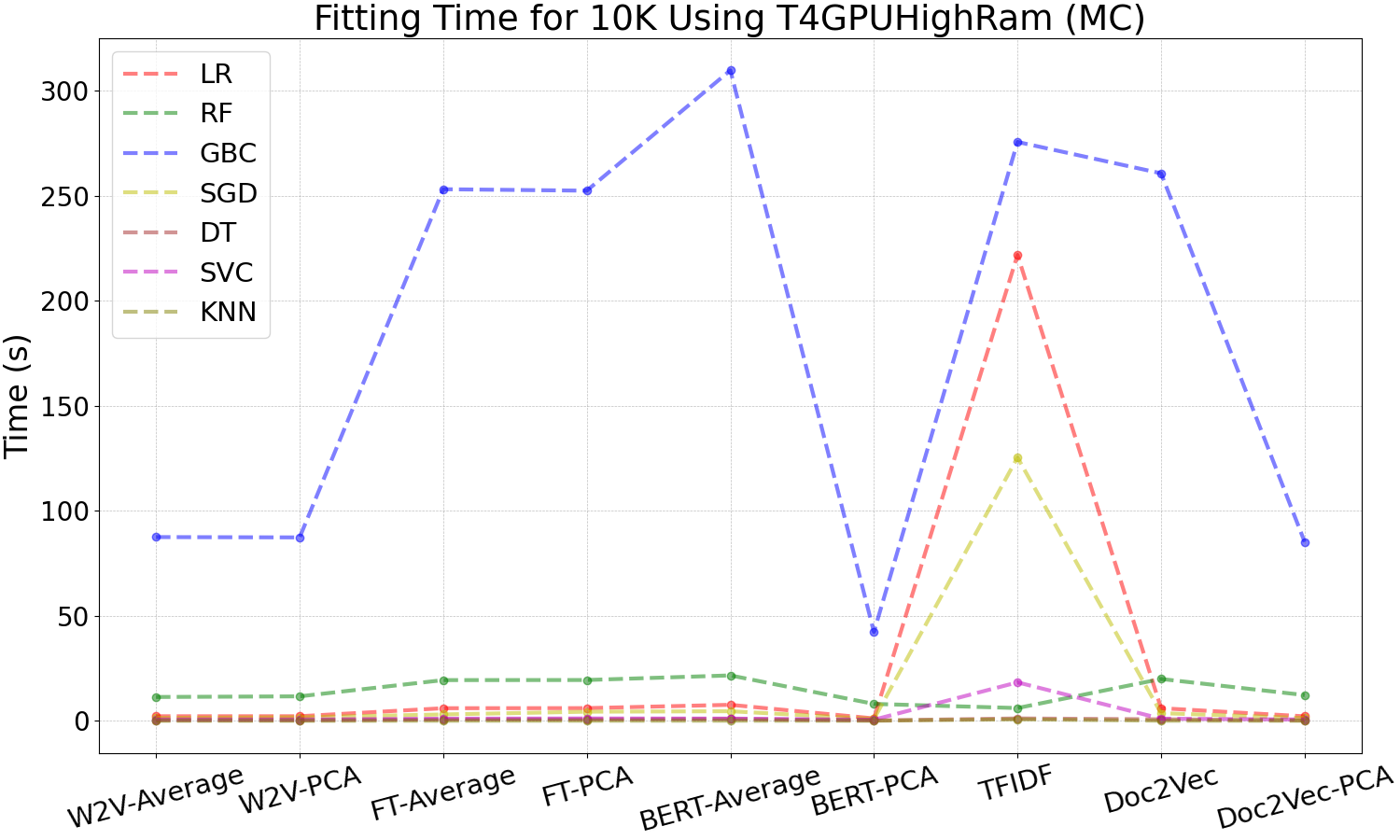}
\caption{Sample size $n=10,000$}
\end{subfigure}

\caption{Fitting time for classifiers for 1K, 5K and 10K sampel sizes using CPU-HighRAM (left-hand side panel)  and T4-GPU-HighRAM (right-hand side panel) for nine types of word embedding as indicated on the horizontal axes, for Multi-Class data.}

\label{Fitting3}
\end{figure}

\begin{figure}[!htbp] 
\centering

\begin{subfigure}[b]{0.49\textwidth}
\centering
\includegraphics[width = 0.94\textwidth]{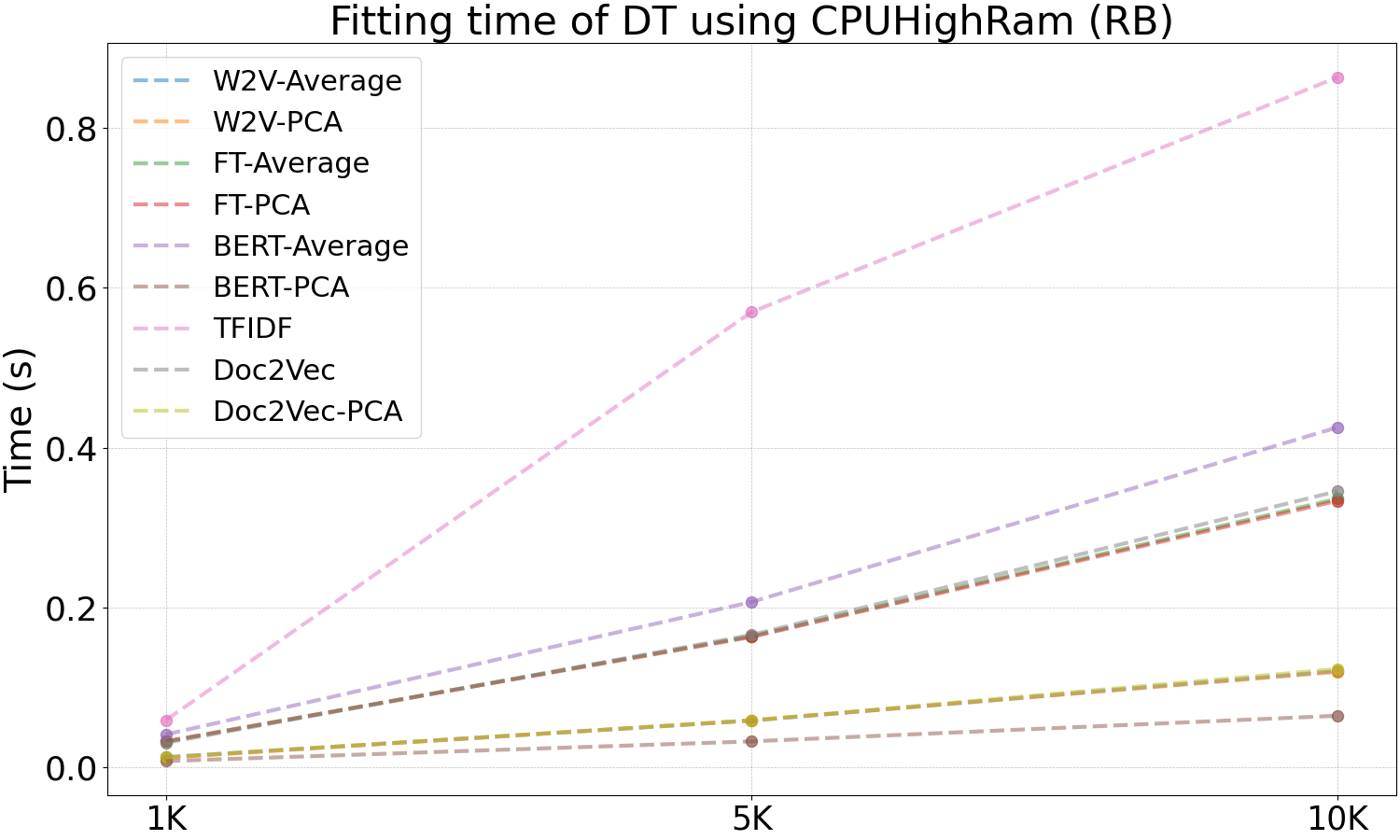}
\caption{Decision tree}
\end{subfigure}
\begin{subfigure}[b]{0.49\textwidth}
\centering
\includegraphics[width = 0.94\textwidth]{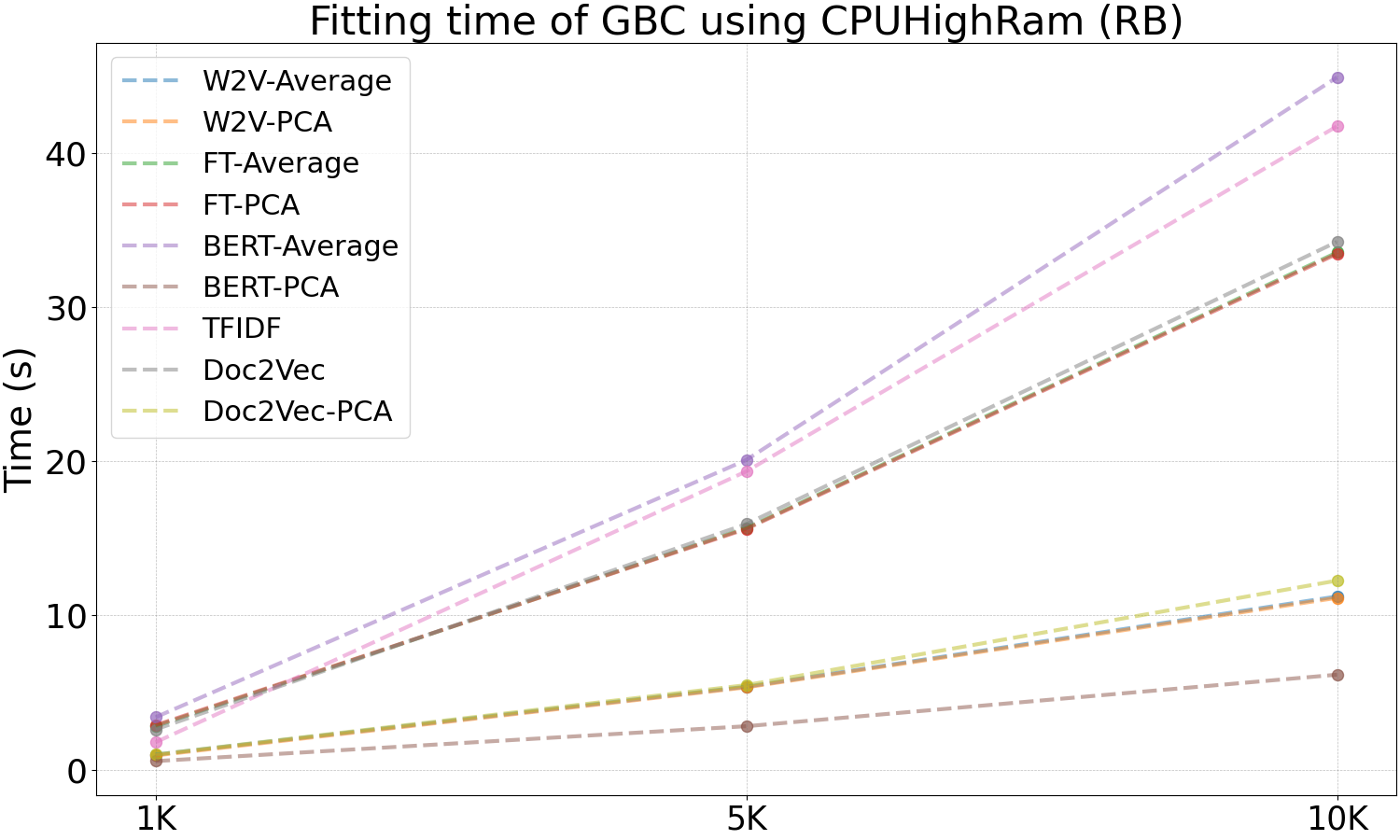}
\caption{Gradient boosting}
\end{subfigure}
\begin{subfigure}[b]{0.49\textwidth}
\centering
\includegraphics[width = 0.94\textwidth]{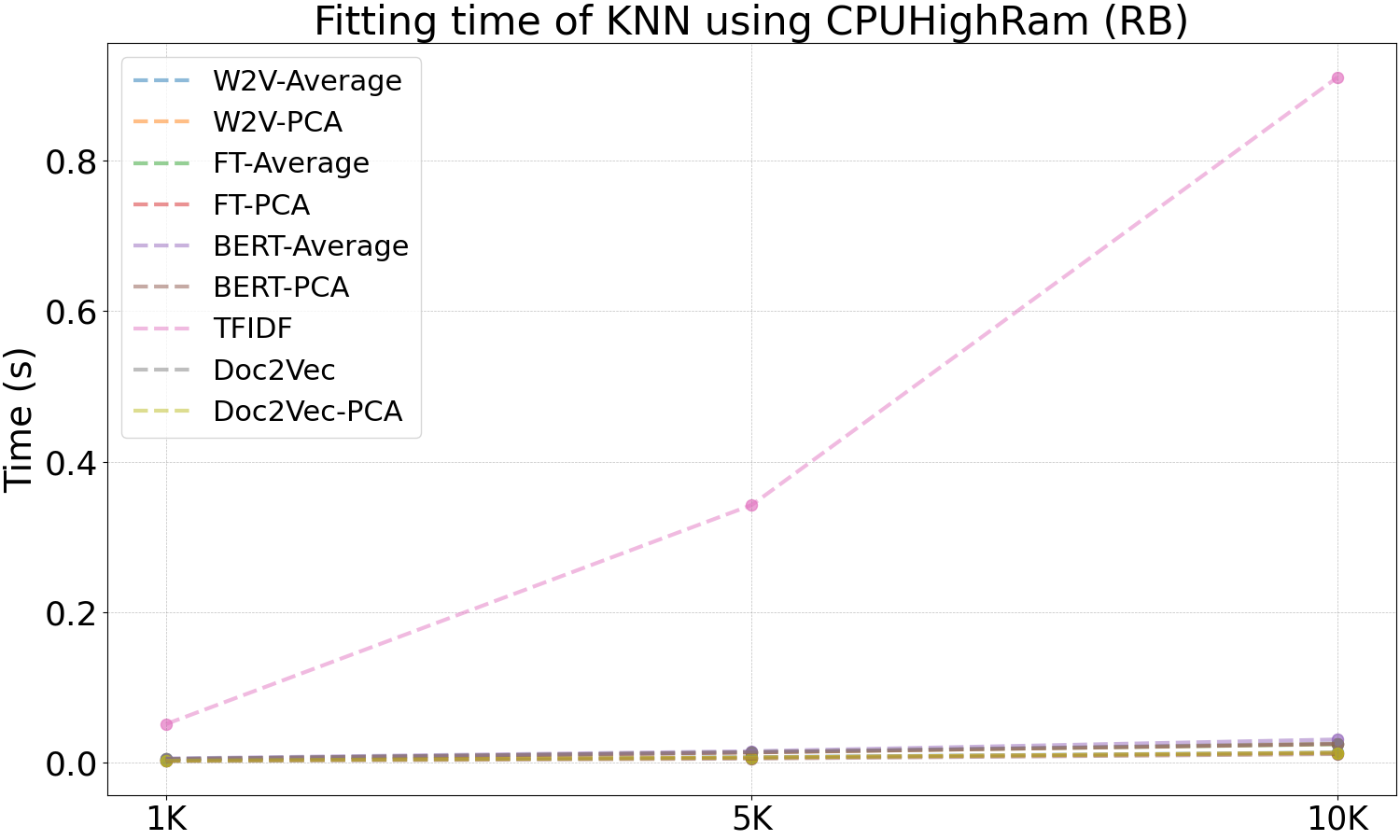}
\caption{K-nearest neighbours}
\end{subfigure}
\begin{subfigure}[b]{0.49\textwidth}
\centering
\includegraphics[width = 0.94\textwidth]{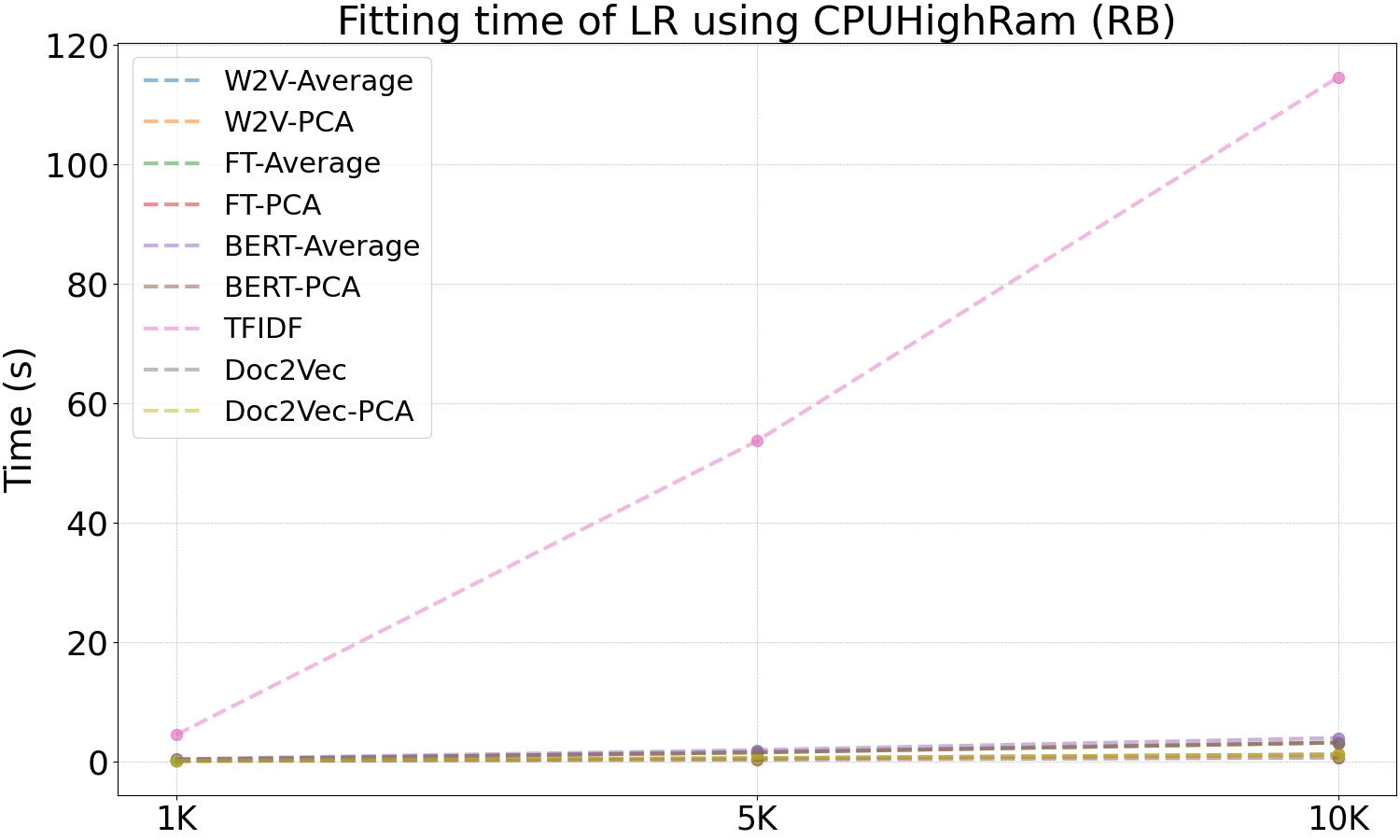}
\caption{Logistic regression}
\end{subfigure}
\begin{subfigure}[b]{0.49\textwidth}
\centering
\includegraphics[width = 0.94\textwidth]{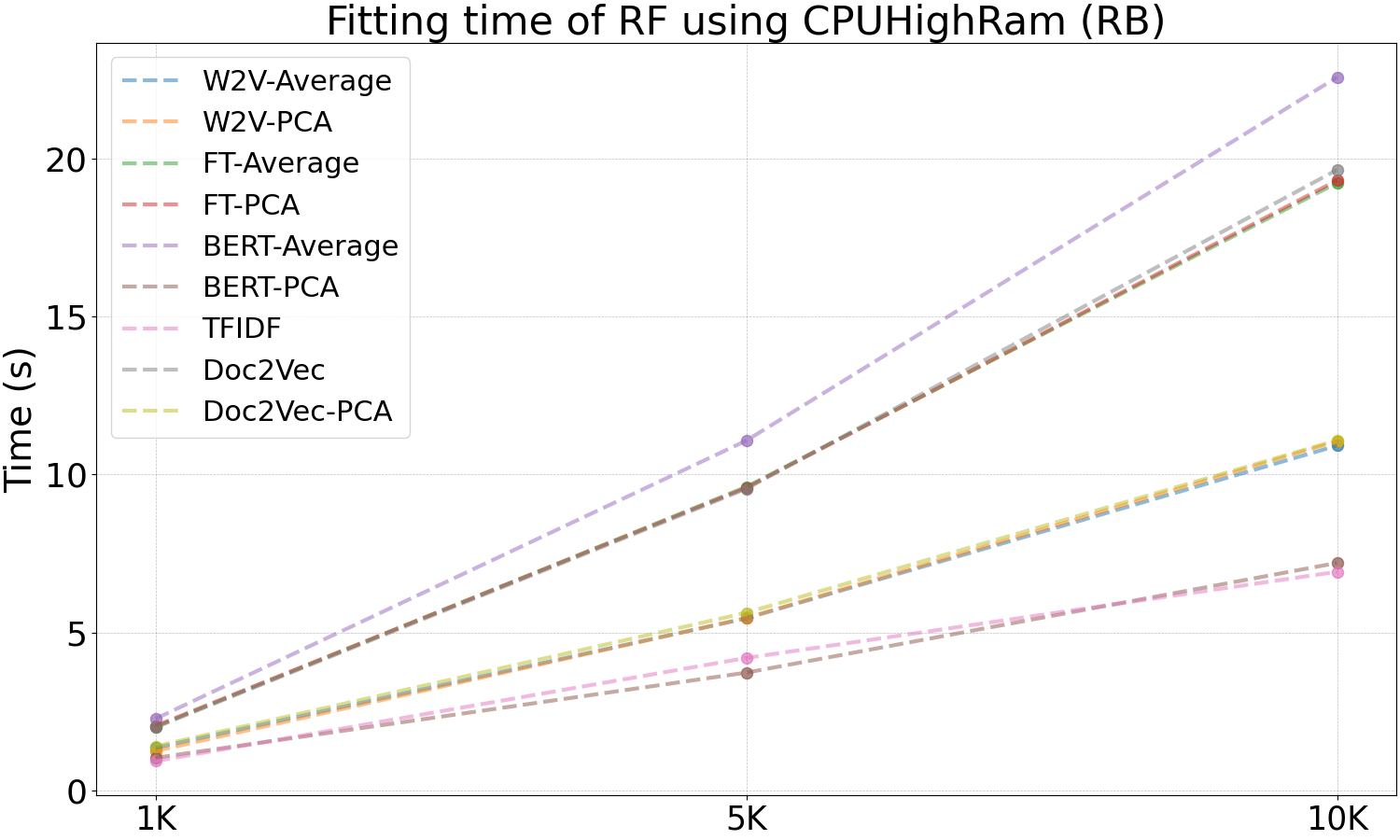}
\caption{Random Forest}
\end{subfigure}
\begin{subfigure}[b]{0.49\textwidth}
\centering
\includegraphics[width = 0.94\textwidth]{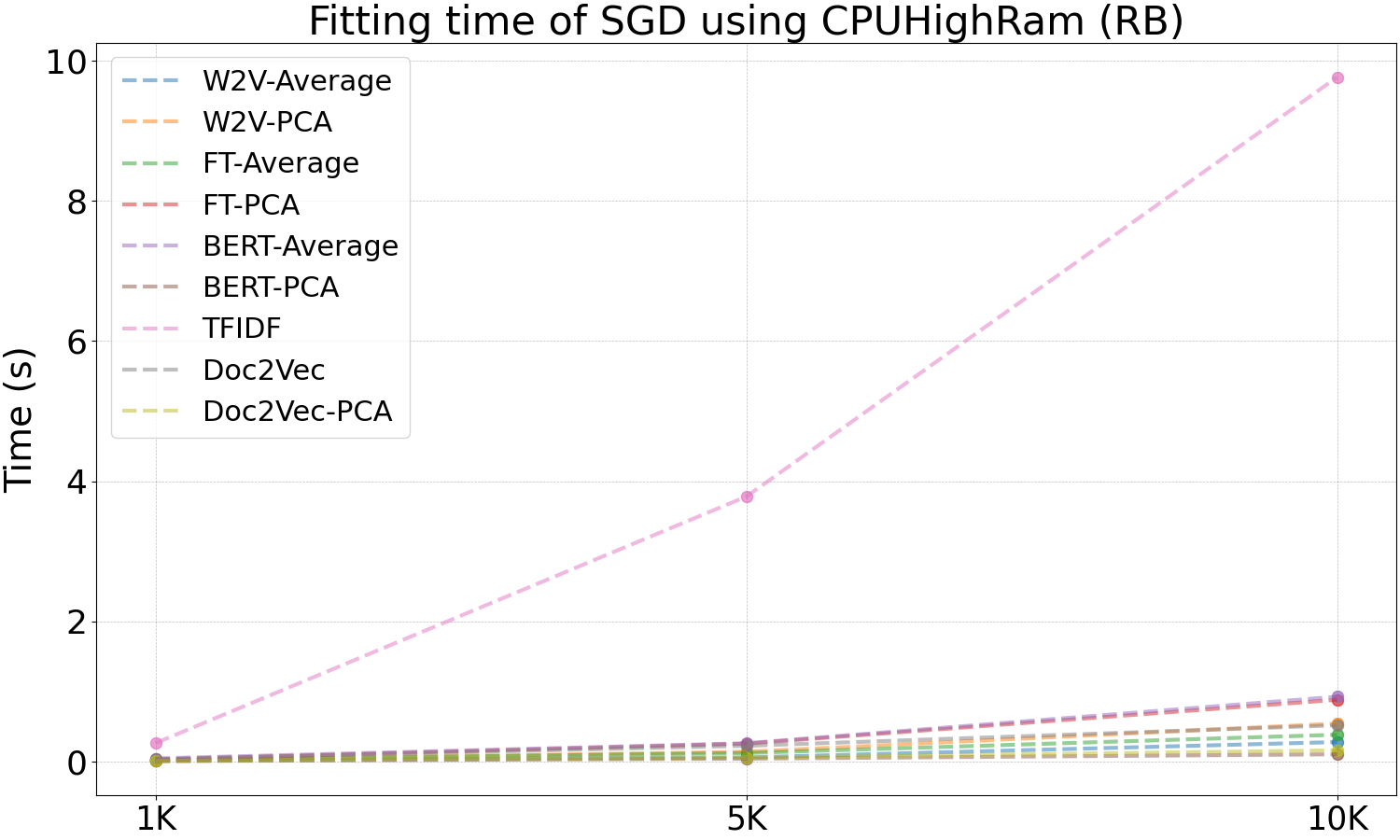}
\caption{SGD Classifier}
\end{subfigure}
\begin{subfigure}[b]{0.49\textwidth}
\centering
\includegraphics[width = 0.94\textwidth]{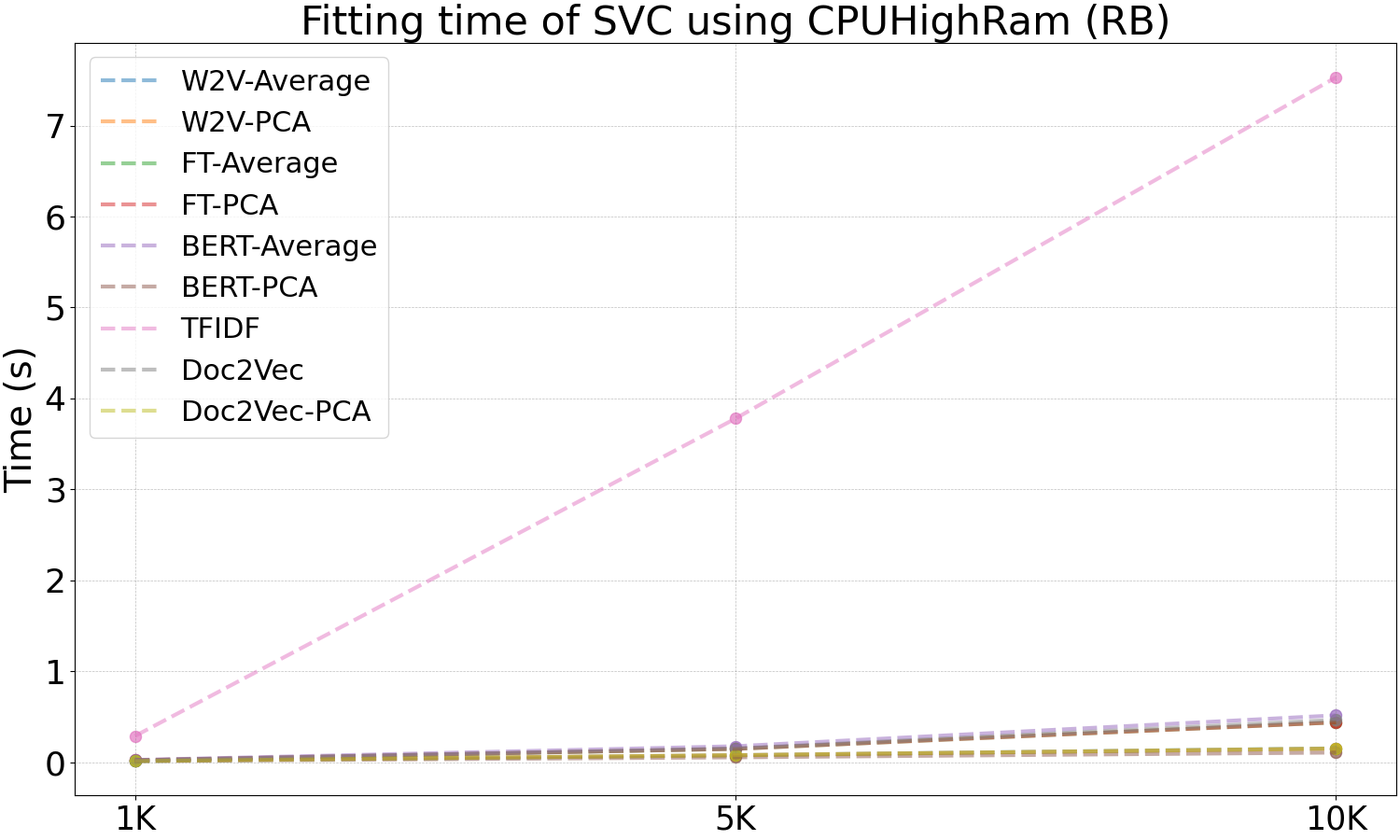}
\caption{Support Vector Machine Classifier}
\end{subfigure}
\caption{Classifiers' fitting times for 1K, 5K and 10K sample sizes using CPU-HighRAM for Radical Binary data. }
\label{Fitting3.1}
\end{figure}

The observed classifiers' times for the nine scenarios of text representations and two processing settings are  visualized in Figures \ref{Fitting1} - \ref{Fitting3} for Mixed Binary, Radical Binary and Multi-Class dataset, respectively. The graphs
show curves of the fitting times against WEs and each curve corresponds to one classifier, with a separate graph for each sample size, for an easy comparison between classifiers and WEs when the sample size is fixed. Moreover, Figure \ref{Fitting3.1} 
visualizes the effect of the sample size on fitting times for each WE with a separate graph for each classifier, for the Radical Binary dataset and CPU-HighRAM processing setting as an example since similar trends were observed for all three datasets.  

For all considered classifiers except the two ensemble methods, Gradient Boosting and Random Forest, using TF-IDF text representation resulted in the slowest training process. This is illustrated in Figure \ref{Fitting3.1}  
by the pink curves being above all other curves in graphs (a), (c), (d), (f) and (g). In particular, for the Radical Binary and Mixed Binary datasets, TF-IDF combined with logistic regression stood out as the slowest among all other WEs and classifiers' combinations, which can be observed as the distinctive peaks on the red curves in Figures \ref{Fitting1} and  \ref{Fitting2}. 

However, for Gradient Boosting and Random Forest, using BERT-Average required more time than any other WE, including TF-IDF. As expected, these two ensemble classifiers, marked by the blue and green curves in Figures \ref{Fitting1} - \ref{Fitting3}, generally required more time than the other methods for all WEs, which is in line with general properties of these algorithms. For Gradient Boosting, when using BERT-PCA, Doc2Vec-PCA and the two Word2Vec text representations the process of training was at least twice as fast as when using BERT-Average, TF-IDF, Doc2Vec and the two FastText WEs. The same conclusions can be made for Random Forest with one difference appearing for TF-IDF which seemed to work relatively fast for this classifier.

When increasing the sample size from 1000 to 5000 and 10,000, the training time grew in an approximately linear manner for all WEs and the steepest increase can be noticed for TF-IDF in most cases.  
For most classifiers, the increase was the least steep for BERT-PCA, D2V-PCA, W2V-Average and W2V-PCA. On the other hand, it was generally much steeper for BERT-Average, FT-Average, FT-PCA and Doc2Vec.

Overall, the same patterns held for both processing settings: CPU-HighRAM and T4-GPU-HighRAM, with the latter being marginally faster, which can be observed by comparing the graphs on the left- and right-hand side in Figures \ref{Fitting1} - \ref{Fitting3}.

In summary, BERT-PCA stood out as a text representation method that yielded relatively fast fitting time for all classifiers and was closely followed by Doc2Vec-PCA, Word2Vec-Average and Word2Vec-PCA. For BERT, applying PCA as opposed to taking the average of features (BERT-Average) yielded faster training of classifiers. Similarly, Doc2Vec-PCA worked faster than Doc2Vec for most classifiers. However, the same effect was not observed for FastText and Word2Vec. That is, FT-Average and FT-PCA required a similar amount of time for classifiers' training, as was the case for W2V-Average and W2V-PCA.
Moreover, using the two Word2Vec-based WEs generally led to faster training of classifiers than the two FastText-based features.

\subsection{Classifiers' Performance}
\label{classifiersperf}

In this section, we focus on the performance of each classifier in terms of macro-averages for F1-Score, precision and recall and assess the effects of the type of WEs used in the process of training. Moreover, two approaches to feature extraction from WEs are compared: taking averages of word vectors and applying a PCA-based approach. Finally, the classifiers are compared with respect to their overall performance.
 
The results are visualized in Figures \ref{F1-Scores-rb} - \ref{F1-Scores-mc} showing macro-average curves of F1-Scores against WEs and each curve corresponds to one classifier, with a separate graph for each sample size. For precision and recall, very similar patterns were observed as those visible in the graphs for F1-Scores.

\subsubsection{Comparison between word embeddings}
\label{Comparisonbetweenwordembeddings}
\begin{figure}[!htbp] 
\centering
\vspace{-5mm}
\begin{subfigure}[b]{1\textwidth}
\centering
\includegraphics[width = 0.65\textwidth]{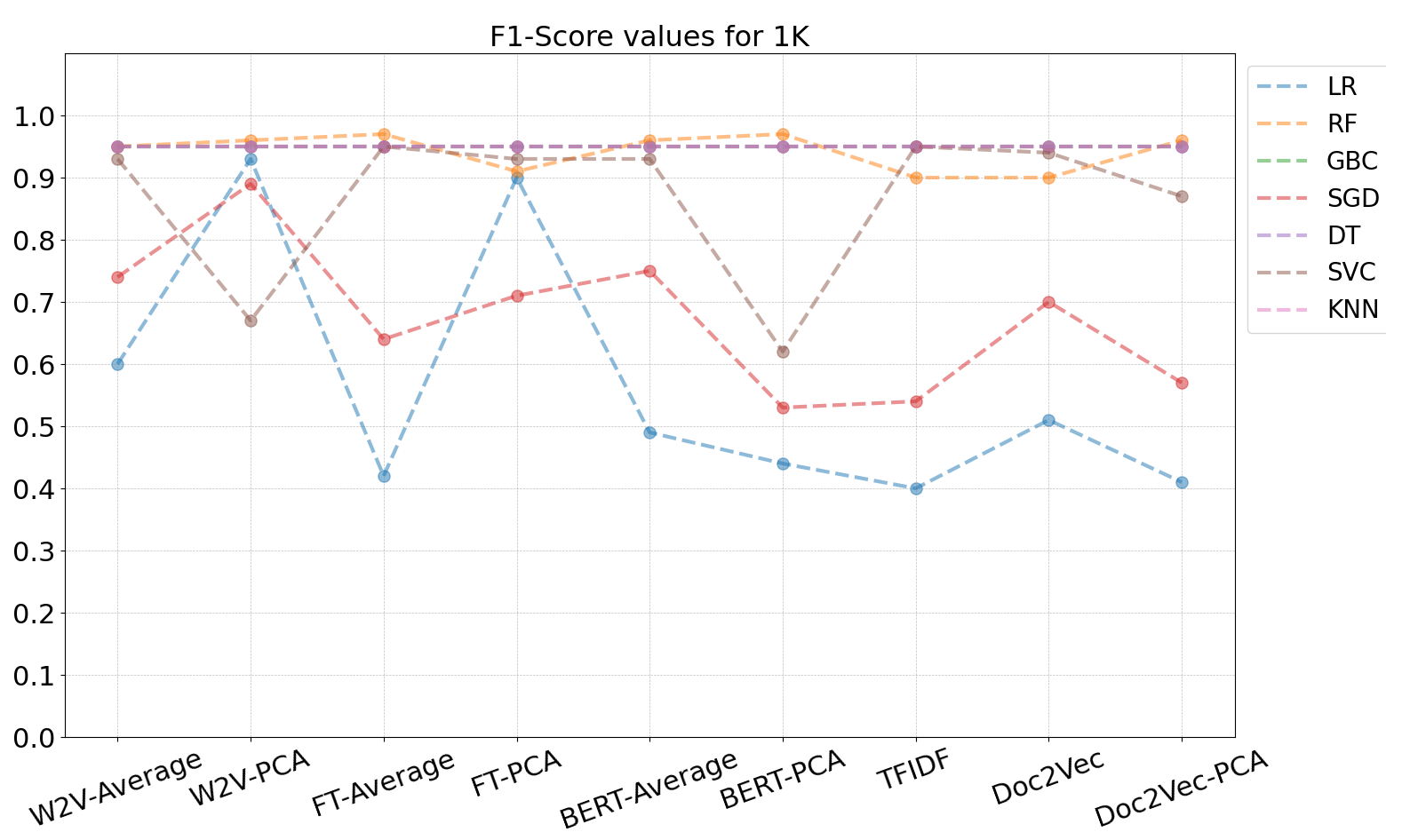}
\caption{$n=1000$}
\end{subfigure}

\begin{subfigure}[b]{1\textwidth}
\centering
\includegraphics[width = 0.65\textwidth]{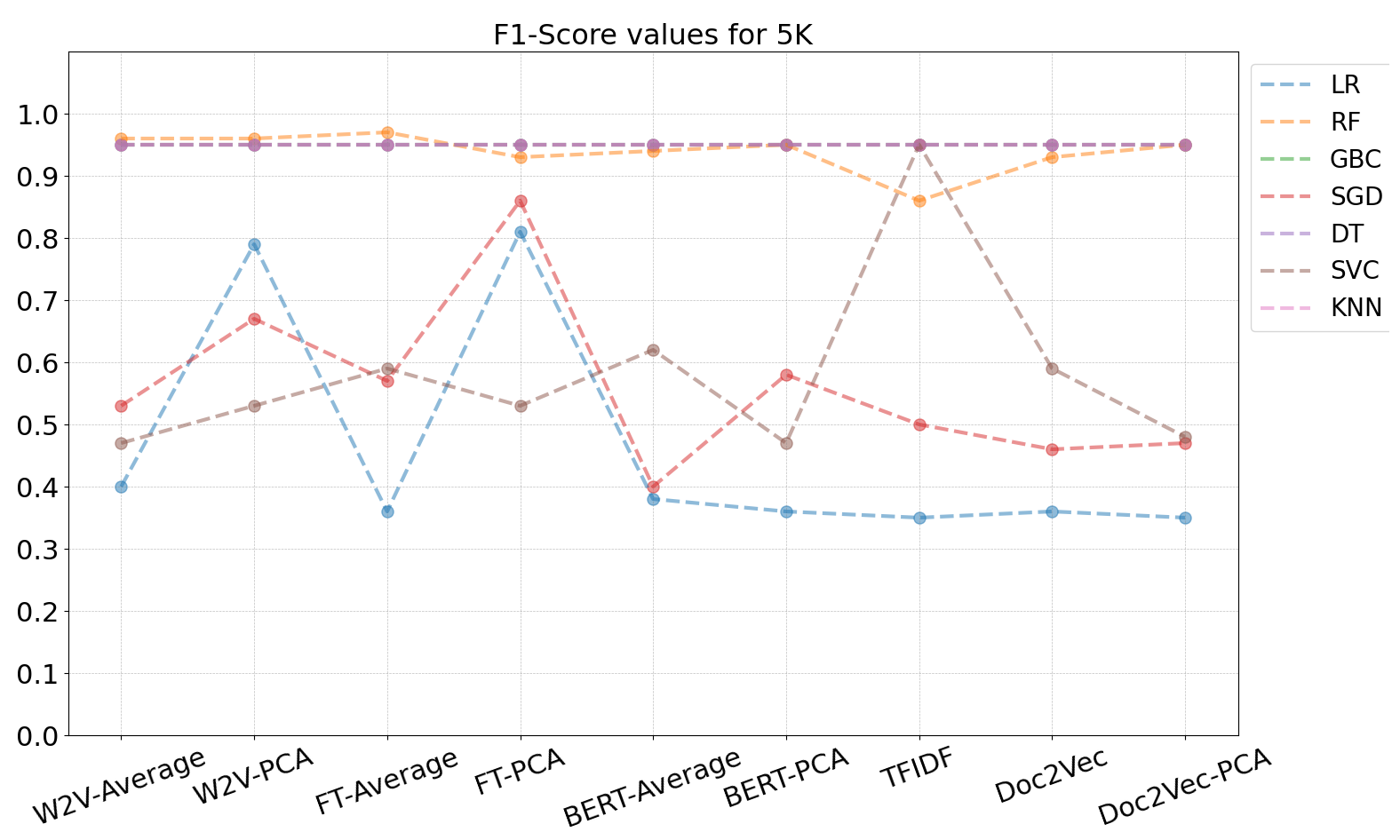}
\caption{$n=5000$}
\end{subfigure}

\begin{subfigure}[b]{1\textwidth}
\centering
\includegraphics[width = 0.65\textwidth]{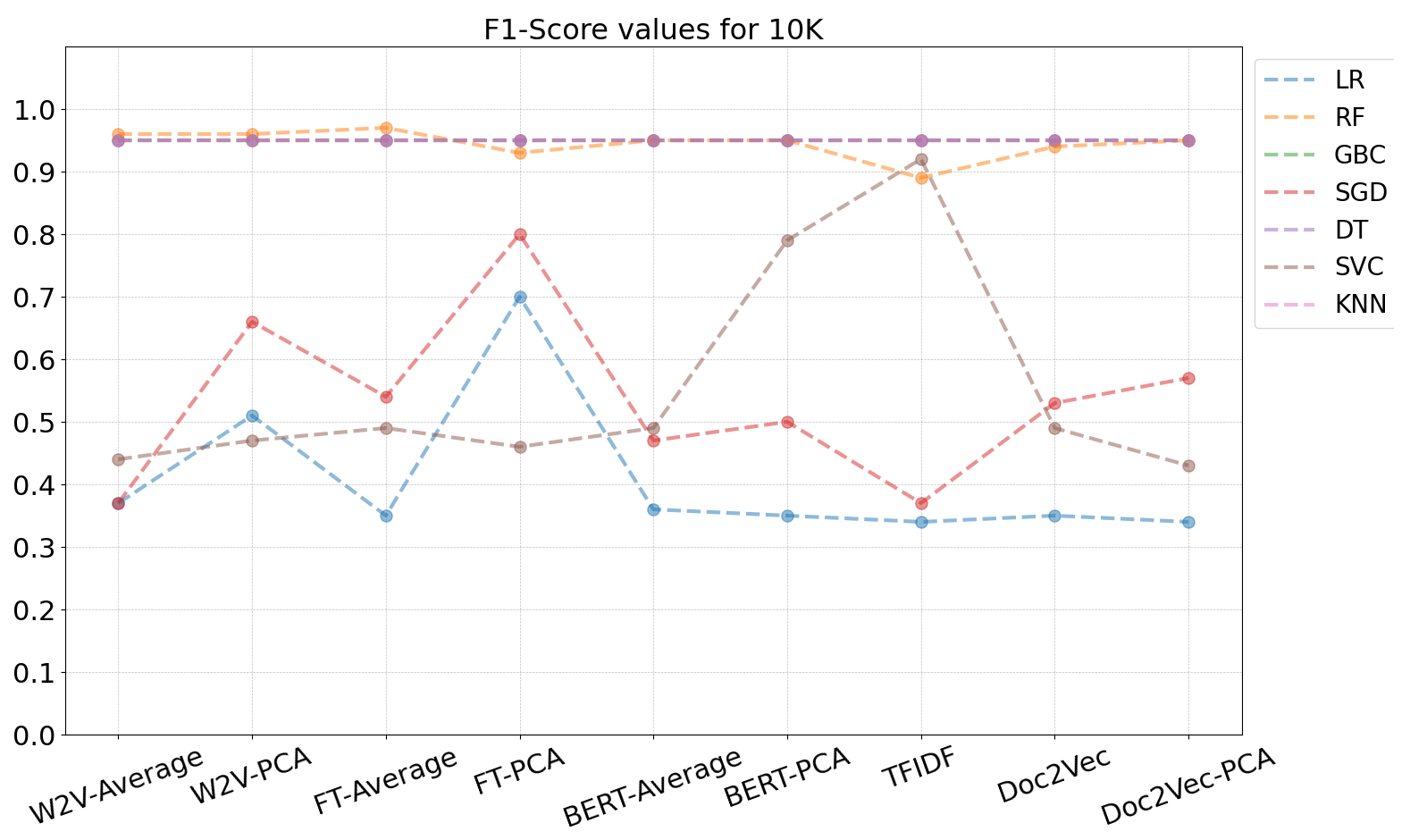}
\caption{$n=10,000$}
\end{subfigure}
\caption{F1-Scores for seven classifiers, each marked with a different colour, and nine WEs (marked on horizontal axis) for  Radical Binary  dataset with sample size of (a) $n=1000$, (b) $n=5000$ and (c) $n=10,000$.
}
\label{F1-Scores-rb}
\end{figure}

\begin{figure}[!htbp] 
\centering

\vspace{-5mm}
\begin{subfigure}[b]{1\textwidth}
\centering
\includegraphics[width = 0.65\textwidth]{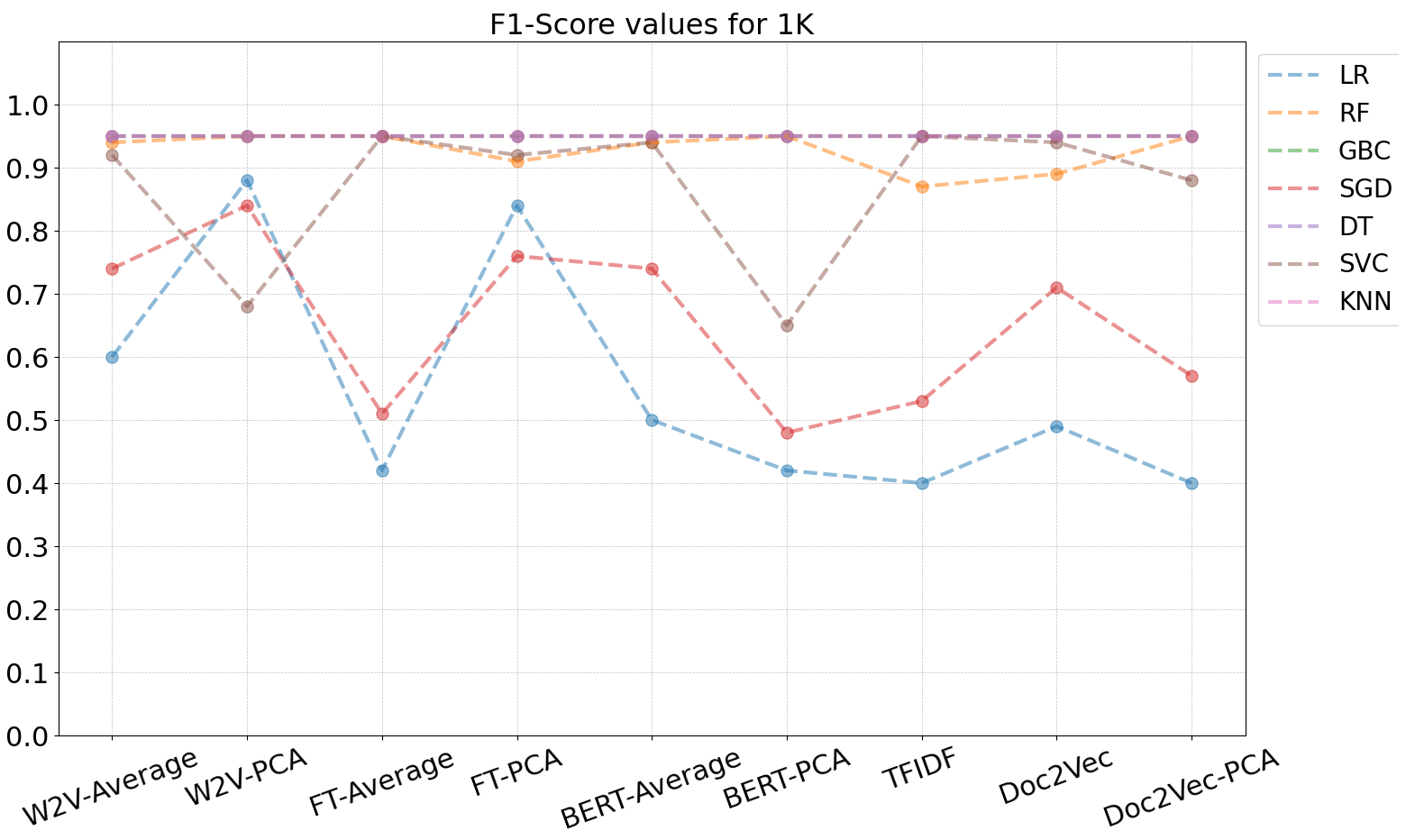 }
\caption{$n=1000$}
\end{subfigure}
\begin{subfigure}[b]{1\textwidth}
\centering
\includegraphics[width = 0.65\textwidth]{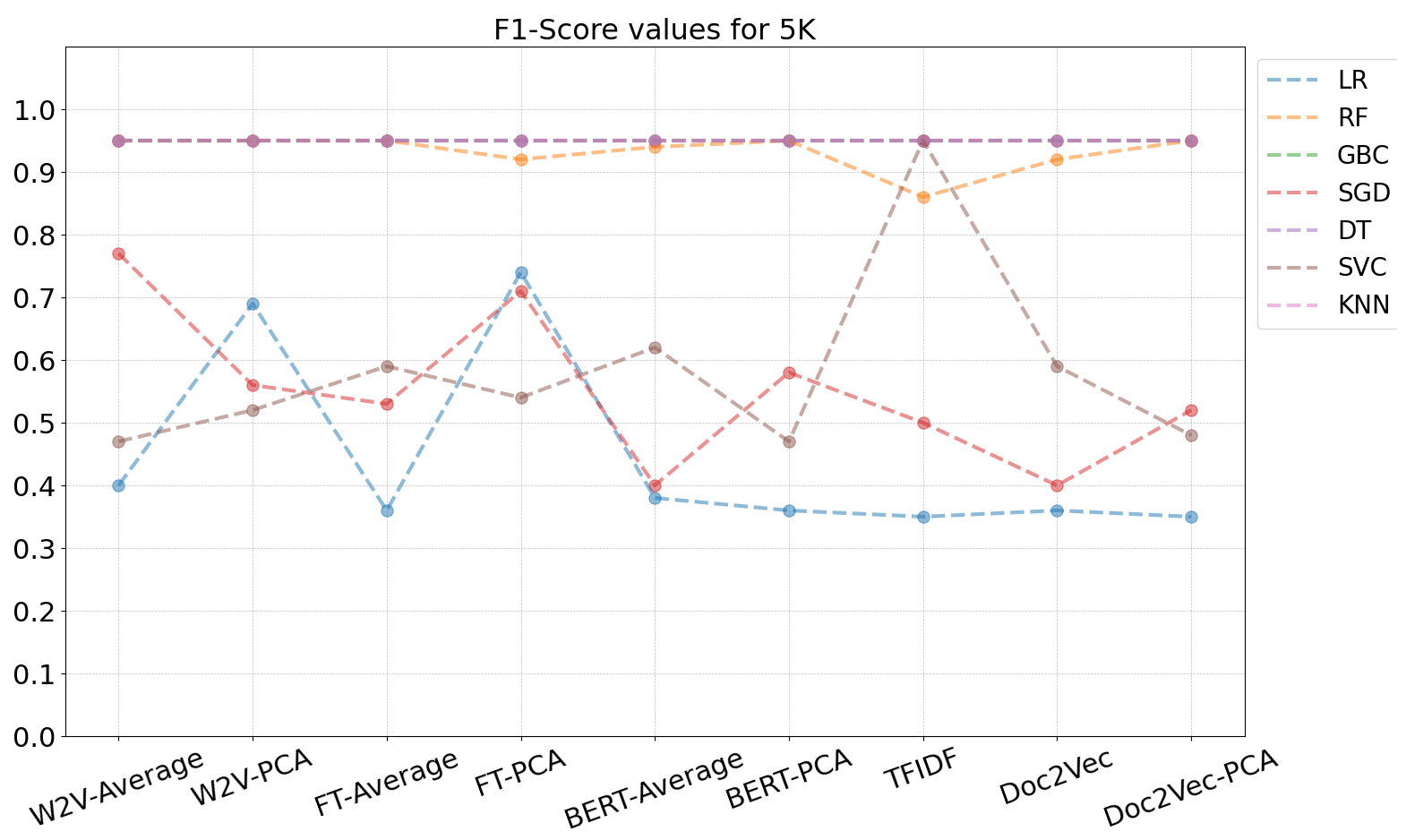}
\caption{$n=5000$}
\end{subfigure}
\begin{subfigure}[b]{1\textwidth}
\centering
\includegraphics[width = 0.65\textwidth]{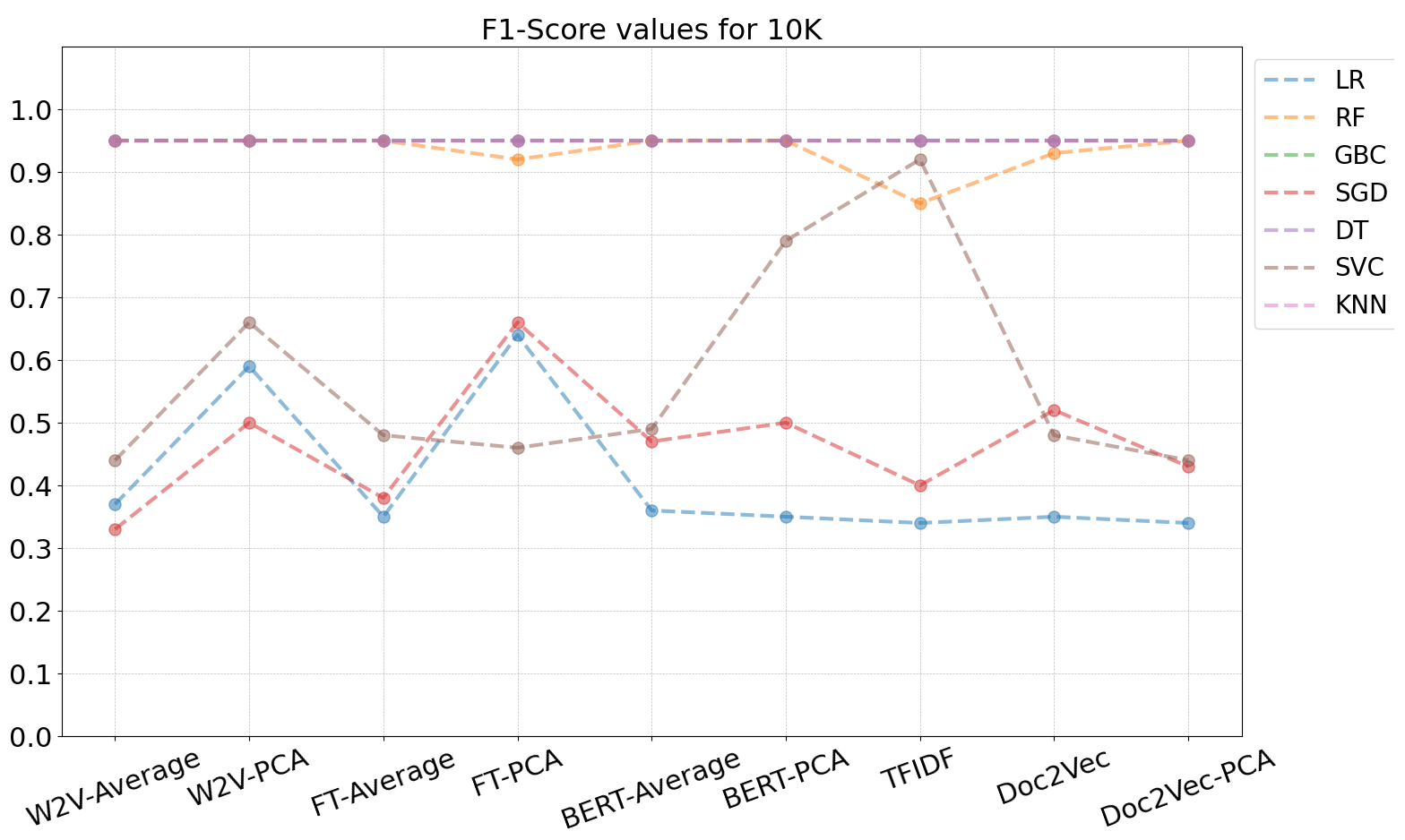}
\caption{$n=10,000$}
\end{subfigure}
\caption{F1-Scores for seven classifiers, each marked with a different colour, and nine WEs (marked on horizontal axis) for  Mixed Binary  dataset with sample size of (a) $n=1000$, (b) $n=5,000$ and (c) $n=10,000$.   }
\label{F1-Scores-mb}
\end{figure}

\begin{figure}[!htbp] 
\centering

\vspace{-5mm}
\begin{subfigure}[b]{1\textwidth}
\centering
\includegraphics[width = 0.65\textwidth]{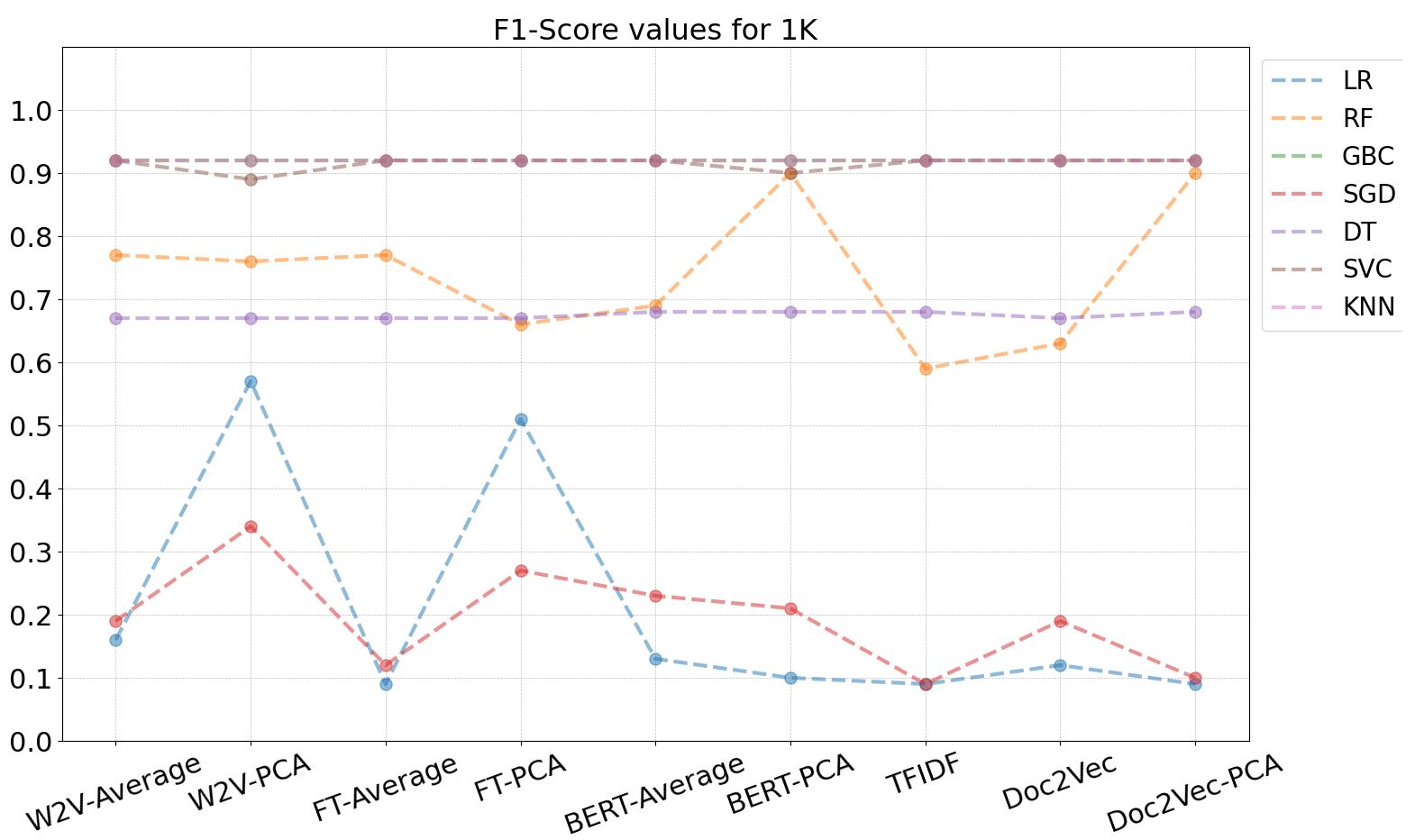}
\caption{$n=1000$}
\end{subfigure}
\begin{subfigure}[b]{1\textwidth}
\centering
\includegraphics[width = 0.65\textwidth]{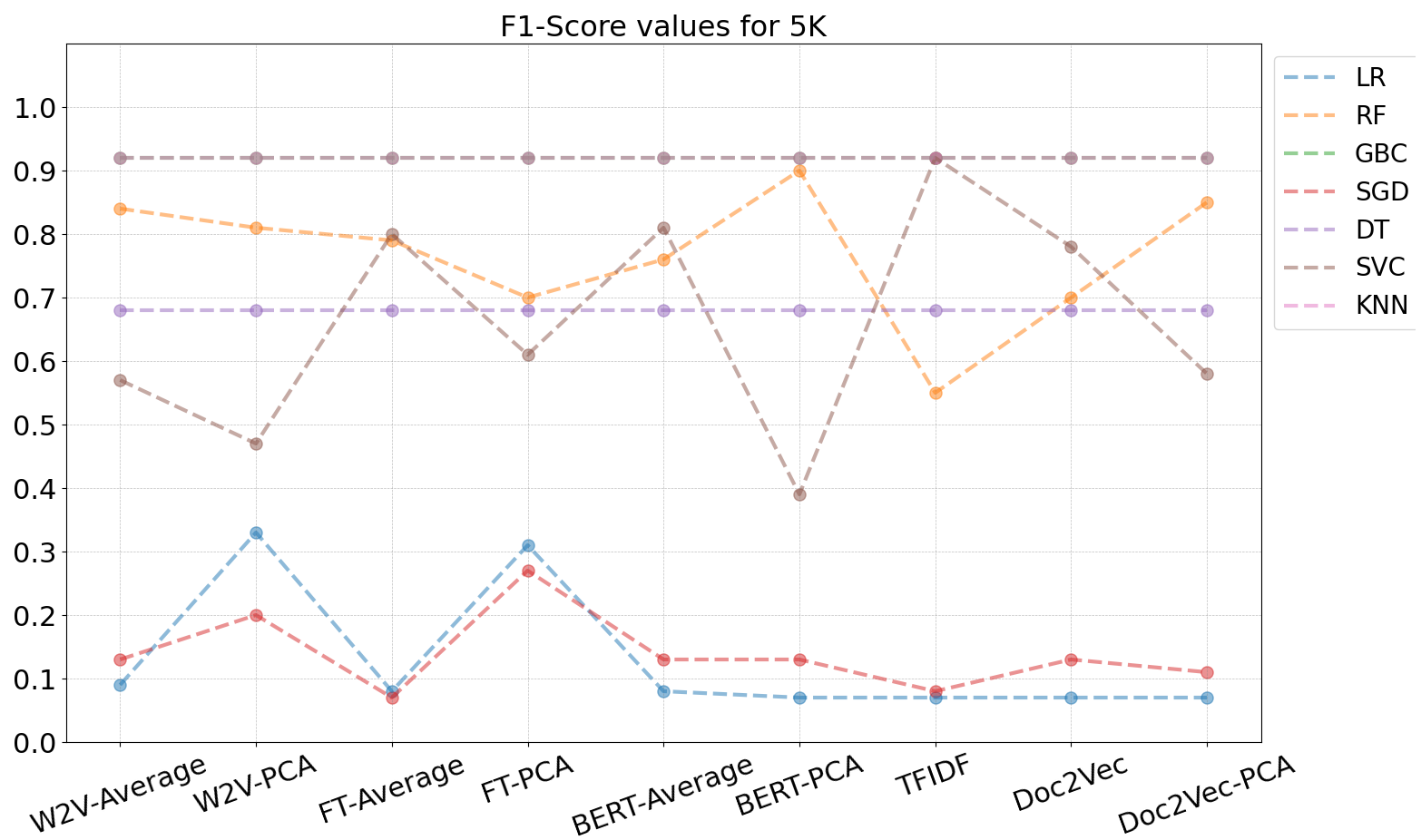}
\caption{$n=5000$}
\end{subfigure}
\begin{subfigure}[b]{1\textwidth}
\centering
\includegraphics[width = 0.65\textwidth]{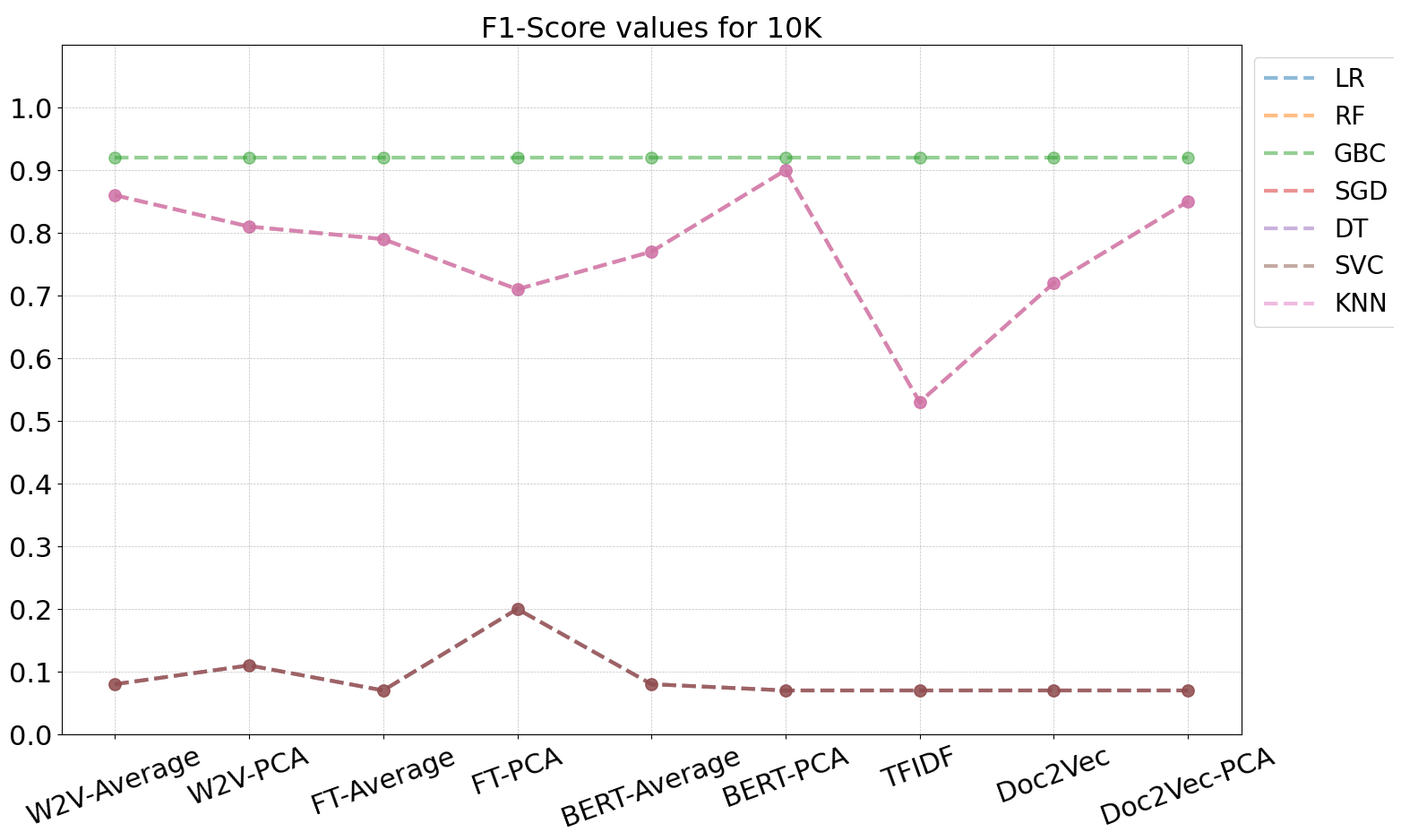}
\caption{$n=10,000$}
\end{subfigure}
\caption{F1-Scores for seven classifiers, each marked with a different colour, and nine WEs (marked on horizontal axis)  for  Multi-Class  dataset with sample size of (a) $n=1000$, (b) $n=5000$ and (c) $n=10,000$. }
\label{F1-Scores-mc}
\end{figure}

In this subsection, the performance of WEs is compared for each classifier separately.
\vspace{2mm}\\
{\bf Comparison of word embeddings for Random Forests} 
\vspace{1mm}\\
For Random Forests (RFs), using TF-IDF consistently led to the lowest F1-Scores, precisions and recalls for all datasets and all sample sizes. This can be observed as a distinct dip in the yellow curves in Figures \ref{F1-Scores-rb} - \ref{F1-Scores-mc}, where RFs applied to TF-IDF achieved F1-Scores ranging between 0.86 and 0.9 for Radical Binary  dataset, between 0.85 and 0.87 for  Mixed Binary  dataset and between 0.53 and 0.59 for the  Multi-Class  dataset, respectively. 

For  Radical Binary and  Mixed Binary datasets, almost all of the other WEs led to very similar results and achieved an excellent performance of RFs with F1-Scores in the narrow range between 0.95 and 0.97 for: Word2Vec-Aver, Word2Vec-PCA, FastText-Aver, BERT-Aver, BERT-PCA and Doc2Vec-PCA.  For FastText-PCA and Doc2Vec only slightly lower scores were observed: between 0.91 and 0.93 for FastText-PCA and between 0.89 and 0.94 for Doc2Vec. 

However, for the most challenging  Multi-Class datasets, the performance of WEs was more varied and BERT-PCA stood out with the highest F1-Scores for RFs equal to 0.9 for all three sample sizes. This is in contrast with the lowest F1-Scores ranging between 0.53 and 0.59 when using TF-IDF for the  Multi-Class  datasets. Doc2Vec-PCA performed well for this dataset with F1-Scores between 0.85 and 0.9 and Word2Vec-Aver performed very similarly to Doc2Vec-PCA for $n=5,000$ (F1-Score=0.84) and $n=10,000$ (F1-Score=0.86) but not as good for the smallest dataset (F1-Score=0.77 for $n=1,000$). On the other hand, Doc2Vec was the second worst after TF-IDF, with F1-Scores between 0.63 and 0.72 and FastText-PCA gave very similar results to Doc2Vec (F1-Scores ranging between 0.66 and 0.71). BERT-Average is marginally better than  Doc2Vec with F1-Scores ranging between 0.69 for $n=1,000$ and 0.77 for $n=10,000$. Finally, FastText-Aver and Word2Vec-PCA performed similarly to each other with F1-Scores between 0.77 and 0.81, which placed them above TF-IDF, Doc2Vec, FastText-PCA and BERT-Aver but below Word2Vec-Aver, Doc2Vec-PCA and BERT-PCA,  ordered from the worst to best-performing model.

In summary, the WEs that stood out for RFs are: TF-IDF, which consistently led to the lowest scores for all datasets and BERT-PCA, which achieved the highest scores for the most challenging classification tasks, closely followed by Doc2Vec-PCA with very similar scores. 
\vspace{2mm}\\
{\bf Comparison of word embeddings for Logistic Regression}
\vspace{1mm}\\
For Logistic Regression (LR), Word2Vec-PCA and FastText-PCA provided consistently better precisions, recalls and F1-Scores over all other WEs. This held true for all three datasets and sample sizes. For example, for the  Radical Binary  dataset of size $n=1000$, F1-Scores of 0.93 and 0.9  were observed for Word2Vec-PCA  and  FastText-PCA, respectively, while all other WEs ranged between 0.4 (TF-IDF) and 0.6 (Word2Vec-Aver).  
For another example, for the  Multi-Class  dataset of size $n=1000$, F1-Scores of 0.57 and 0.51 were observed for Word2Vec-PCA  and  FastText-PCA, respectively, while all other WEs ranged between 0.09 (TF-IDF) and 0.16 (Word2Vec-Aver) and the advantage of using Word2Vec-PCA or FastText-PCA is more pronounced for this most challenging dataset. The same pattern held for the datasets of sizes 5000 and 10000, however, the scores became lower as the sample size increased which suggests possible over-fitting of the model.

The TF-IDF method performed generally very poorly for LR, however its scores were very close to several other WEs with similar poor results. In particular, in most cases, Doc2Vec-PCA achieved the same F1-Scores as TF-IDF ranging between 0.35 and 0.4 for the  Radical Binary  and  Mixed Binary  datasets and between 0.07 and 0.09 for the  Multi-Class  datasets.
\vspace{2mm}\\
{\bf Comparison of word embeddings for Stochastic Gradient Decent}
\vspace{1mm}\\
For Stochastic Gradient Decent (SGD), the patterns were similar to those observed for LR in that Word2Vec-PCA and FasText-PCA led to the best scores in most cases. However, there are also a few differences here, in particular for the BERT WEs which performed relatively well in some cases and poorly in others.

For example, for the  Radical Binary  dataset of size $n=1000$, the best F1-Scores of $0.89$, $0.75$ and $0.74$ were observed for Word2Vec-PCA, BERT-Average and Word2Vec-Average, respectively, while all other WEs ranged between $0.53$ (BERT-PCA) and $0.71$ (FastText-PCA).  
For another example, for the  Mixed Binary  dataset of size $n=1000$, the highest F1-Scores of $0.84$ and $0.76$ were observed for Word2Vec-PCA  and  FastText-PCA, respectively, while all other WEs ranged between $0.48$ (BERT-PCA) and $0.74$ (W2V-Average and BERT-Average).
Finally, for the  Multi-Class  dataset of size $n=1000$, the highest F1-Scores of $0.34$ and $0.27$ were observed for Word2Vec-PCA  and  FastText-PCA, respectively, while all other WEs ranged between $0.09$ (TF-IDF) and $0.23$ (BERT-Average).

In summary, for the SGD classifier, Word2Vec-PCA and FasText-PCA stood out as leading to the best scores in most cases. However, there is no clear pattern regarding which WE leads to the lowest scores here, with TF-IDF, BERT-PCA and BERT-Average being the worst on some occasions, but not consistently.
\vspace{2mm}\\
{\bf Comparison of word embeddings for Support Vector Classifier}
\vspace{1mm}\\
For SVC, a different pattern occurred in comparison to other classifiers. Surprisingly, in most cases, TF-IDF text representation yielded the highest F1-Scores, precisions and recalls. This can be seen as a distinct peak in the brown curves in Figure \ref{F1-Scores-rb} (b)-(c), Figure \ref{F1-Scores-mb} (b)-(c) and Figure \ref{F1-Scores-mc} (b), where the F1-Scores for SVC with TF-IDF text embedding were between $0.92$ and $0.95$, roughly twice as high as for the lowest scoring WEs.

For the  Radical Binary  dataset of size $n=1000$, F1-Scores of $0.95$ for both FastText-Average and TF-IDF and $0.94$ for Doc2Vec-Average were observed, while all other WEs ranged between $0.62$ (BERT-PCA) and $0.93$ (W2V-Average). For the  Radical Binary  dataset of size $n=5000$, the F1-Score for TF-IDF was still high (0.95) but all the other WEs yielded lower F1-Scores ranging between $0.47$ for Word2Vec-Average and BERT-PCA and $0.62$ for BERT-Average. 
For the  Mixed Binary  dataset of size $n=1000$, F1-Scores of $0.95$ for both FastText-Average and TF-IDF and $0.94$ for Doc2Vec-Average and BERT-Average were observed, while all other WEs ranged between $0.65$ (BERT-PCA) and $0.92$ (W2V-Average). 
For the Multi-Class dataset of size $n=1000$, F1-Scores of $0.92$ for seven classifiers Word2Vec-Average, FastText-Average, FastText-PCA, BERT-Average, TF-IDF, Doc2Vec-Average and Doc2Vec-PCA were observed, while Word2Vec-PCA was $0.89$ and BERT-PCA was $0.90$.

Generally, BERT-PCA stood out for the SVC classifier as yielding the lowest F1-Scores in most cases for sample sizes $1000$ and $5000$, while TF-IDF stood out as the best-performing WE. Also, in many cases, the SVC classifier achieved better scores for smaller sample sizes, which can be indicative of overfitting.
\vspace{2mm}\\
{\bf Comparison of word embeddings for Decision Tree}
\vspace{1mm}\\
For Decision Tree (DT), it was found that in almost all cases all WE algorithms had the same performance.
For the  Radical Binary  dataset, F1-Scores and recalls were always $0.95$ and precisions were always $0.97$, for all sample sizes and all WEs. The same pattern occurred for the  Mixed Binary  dataset. 

 For the  Multi-Class  dataset of size $n=1000$, the F1-Scores varied only minimally, but this time they were lower and ranging between 0.67 and 0.68, while for $n=5000$, all WEs again yielded the same F1-Scores,  equal to 0.67. Finally, for $n=10,000$, different performances of WEs were observed with the highest F1-Score (0.9) achieved by BERT-PCA and the lowest (0.53) for TF-IDF.

In summary, for DTs, any differences between WEs occurred only for the  Multi-Class  dataset of size $n=10,000$. Where differences occurred, BERT-PCA yielded the highest F1-Score and TF-IDF the lowest. For the remaining eight datasets, the DT classifier was not sensitive to the type of WE employed. 
\vspace{2mm}\\
{\bf Comparison of word embeddings for KNN }
\vspace{1mm}\\
Following the same pattern as DTs, the KNN classifiers had identical values of F1-Scores ($0.95$), precisions ($0.97$) and recalls ($0.95$) for all WEs for the  Radical  and  Mixed Binary  dataset and for all sample sizes.
For the  Multi-Class  datasets of size $n=1000$ and $n=5000$, the KNN classifiers also had identical F1-Scores of $0.92$ for all WEs, while for size $n=10,000$ a variety of values for F1-Score were observed, with a minimum of $0.53$ for TF-IDF and a maximum of $0.9$ for BERT-PCA.
\vspace{2mm}\\
{\bf Comparison of word embeddings for Gradient Boosting}
\vspace{1mm}\\
For Gradient Boosting Classifiers (GBCs), a very similar pattern as for DTs and KNN method was observed. The GBCs had the same value of F1-Score ($0.95$), precision ($0.97$) and recall ($0.95$) for all WEs for both  Radical Binary  and  Mixed Binary  datasets of all three sample sizes. 
For the  Multi-Class  datasets of all sizes, 
the values were also constant for all WEs and equal to $0.93$ for precision and $0.92$ for F1-Scores and recalls. 
Therefore, the GBCs appeared to be insensitive to the type of WE employed for the studied datasets.
\vspace{2mm}\\
{\bf Conclusion}
\vspace{1mm}\\
The comparison of word embeddings across different classifiers revealed notable variations in their effectiveness. TF-IDF performed the worst for most classifiers, except for SVC.  On the other hand, BERT-PCA appeared as a strong performer, especially for multiclass classification tasks. 
Word2Vec-PCA and FastText-PCA demonstrated best results for LR and SGD. Moreover, DTs, KNN and Gradient Boosting classifiers were mostly insensitive to the type of WE, except in the largest multiclass dataset, where BERT-PCA excelled. These findings emphasize the importance of selecting appropriate WEs based on both the classification model and dataset complexity.
\FloatBarrier

\FloatBarrier

\subsubsection{Assessment of the PCA method for feature engineering}
\label{ComparisonbetweenaveragesandPCAforfeatureselection}

For Word2Vec, FastText, BERT and Doc2Vec, two ways of feature engineering were applied to determine a better approach in each case. For Word2Vec and FastText, we compare taking the arithmetic mean of vectors which is the more popular practice to using PCA which attempts to maximise  the amount of information extracted from the words' vectors. Moreover, for BERT, PCA is applied to reduce the dimension of pooled averages of word vectors from 384 to 50. In addition, we apply PCA to Doc2Vec document vectors, where the dimension is reduced from 300 to 100, to investigate its usefulness in comparison to taking the entire Doc2Vec document vector.

\begin{figure}[!htbp] 
\centering
\begin{subfigure}[b]{1\textwidth}
\centering
\includegraphics[width = 0.75\textwidth]{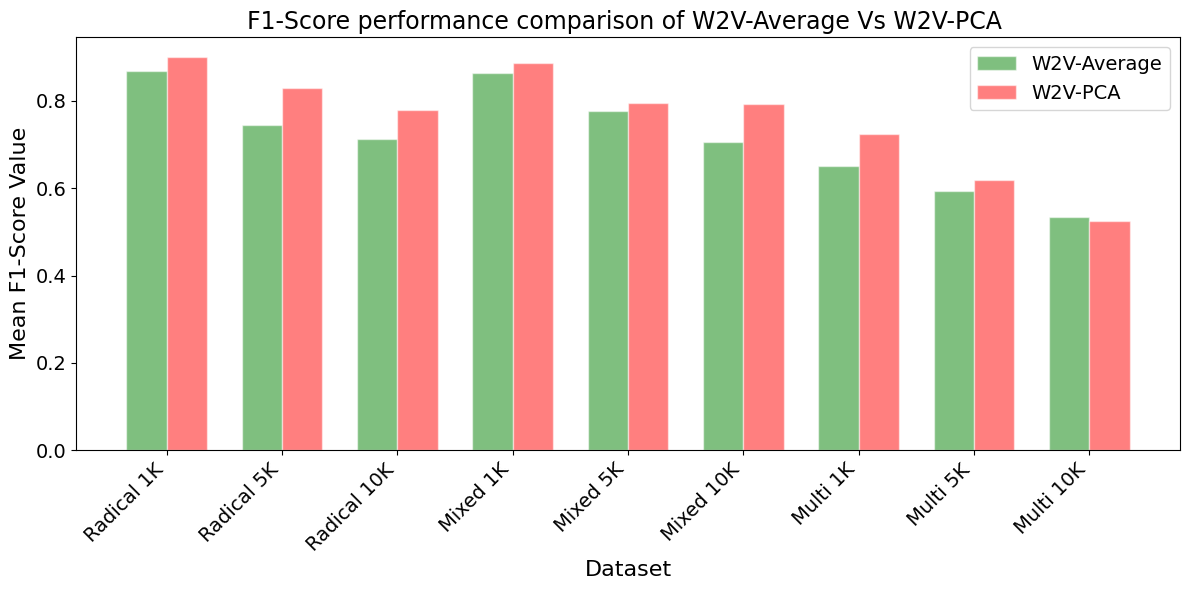}
\caption{Comparison of Word2Vec-Average vs Word2Vec-PCA} 
\vspace*{5mm}
\end{subfigure}

\begin{subfigure}[b]{1\textwidth}
\centering
\includegraphics[width = 0.75\textwidth]{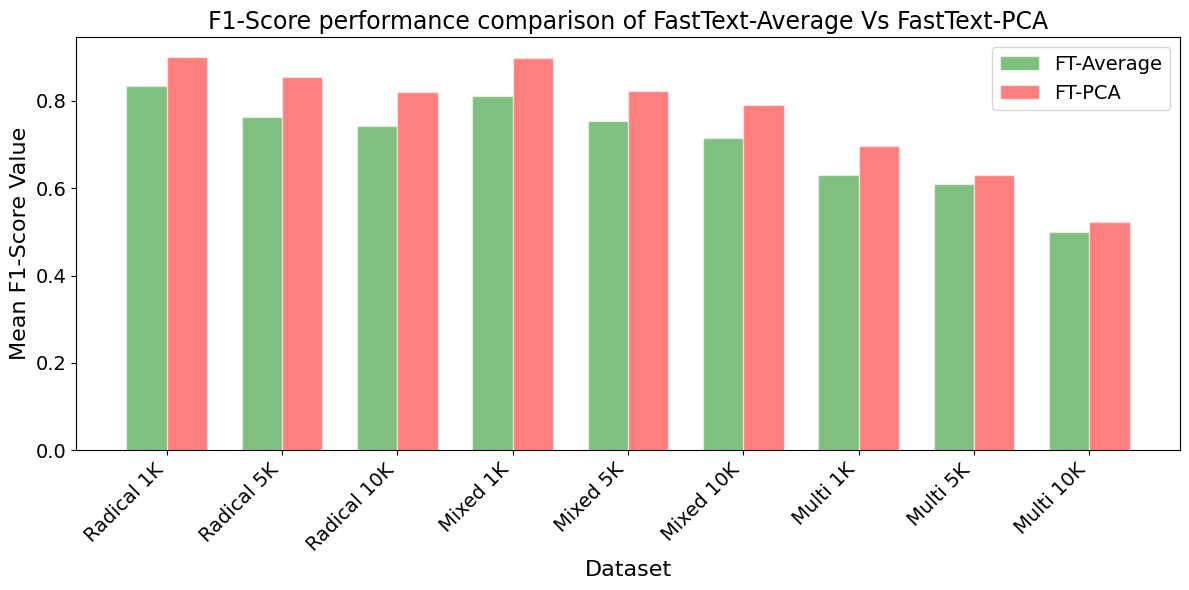}
\caption{Comparison of FastText-Average vs FastText-PCA}
\end{subfigure}
\centering
\caption{The averages of F1-Scores over seven classifiers for each dataset (marked on the horizontal axis), for Word2Vec-Average vs Word2Vec-PCA (top graph) and FastText-Average vs FastText-PCA (bottom graph).}
\label{Chart1}
\end{figure}

\begin{figure}[!htbp] 
\centering
\begin{subfigure}[b]{1\textwidth}
\centering
\includegraphics[width = 0.75\textwidth]{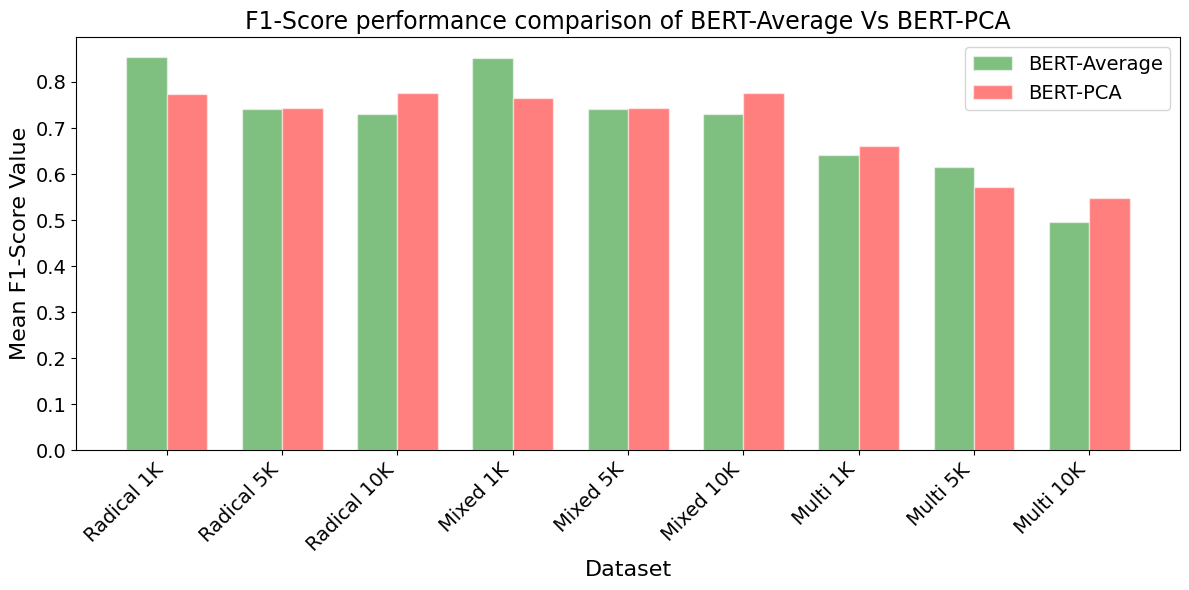}
\caption{Comparison of BERT-Average vs BERT-PCA} 
\vspace*{5mm}
\end{subfigure}
\begin{subfigure}[b]{1\textwidth}
\centering
\includegraphics[width = 0.75\textwidth]{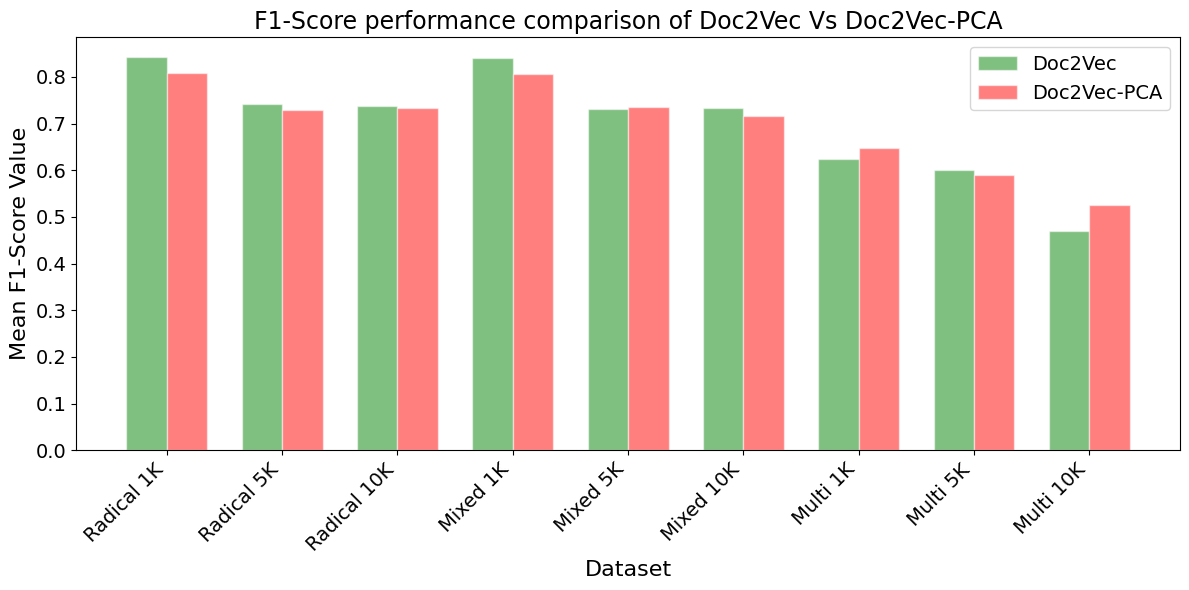}
\caption{Comparison of Doc2Vec vs Doc2Vec-PCA} 
\end{subfigure}
\centering
\caption{The averages of F1-Scores over seven classifiers for each  dataset (marked on the horizontal axis), for BERT-Average vs BERT-PCA (top graph) and Doc2Vec vs Doc2Vec-PCA.}
\label{Chart2}
\end{figure}

 Figures \ref{Chart1} and \ref{Chart2} compare the average F1-Score values over the seven classifiers, for each studied dataset, for Word2Vec-Average vs Word2Vec-PCA, FastText-Average vs FastText-PCA, BERT-Average vs BERT-PCA and Doc2Vec vs Doc2Vec-PCA, respectively. As shown in Figure \ref{Chart1}(a), for Word2Vec, the mean performance of classifiers was better for PCA than for averages for all datasets except one ("Multi 10K") where the mean F1-Score was only marginally worse for PCA. Similarly, for FastText, the mean performance of classifiers was consistently better for PCA than for taking averages, for every dataset, as shown in Figure  \ref{Chart1}(b). Here, the relative improvement in the mean F1-Score while using PCA instead of averages ranged between 3.5\% and 12\%. In most individual cases, PCA performed better than averages for Word2Vec and FastText.
For example, for the  Radical Binary  dataset with $n = 1000$, the F1-Score for LR with Word2Vec-Average was $0.6$ while with Word2Vec-PCA it was $0.93$ and the improvement was even bigger for FastText where the F1-Score for taking Average was $0.42$ while for PCA it was $0.9$. 

For BERT, when using features of a reduced dimension via PCA, the mean F1-Scores were better than or roughly equal to the F1-Scores when taking the entire 384-dimensional vectors for six out of nine datasets, as can be seen in Figure \ref{Chart2}(a). However, in three cases, using the non-reduced features performed better than PCA, on average. For example, for SGD with BERT-Average the F1-Score was $0.75$ and with BERT-PCA it was $0.53$.

Finally, the comparison in Figure \ref{Chart2}(b) reveals that for Doc2Vec, applying dimension reduction using PCA gave better mean F1-Scores only for three datasets and in the remaining six cases, taking the entire original Doc2Vec vector gave slightly better results, on average. This indicates that all the components of the Doc2Vec vectors carried useful information for the classification task.

\begin{figure}[!htbp] 
\centering

\begin{subfigure}[b]{0.46\textwidth}
\includegraphics[width = 1\textwidth]{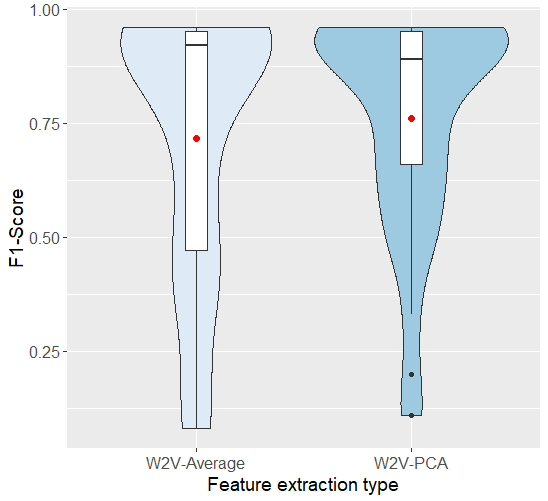}
\caption{Comparison between Word2Vec-Average vs Word2Vec-PCA} 
\end{subfigure}
\hspace*{5mm}
\begin{subfigure}[b]{0.46\textwidth}
\includegraphics[width = 1\textwidth]{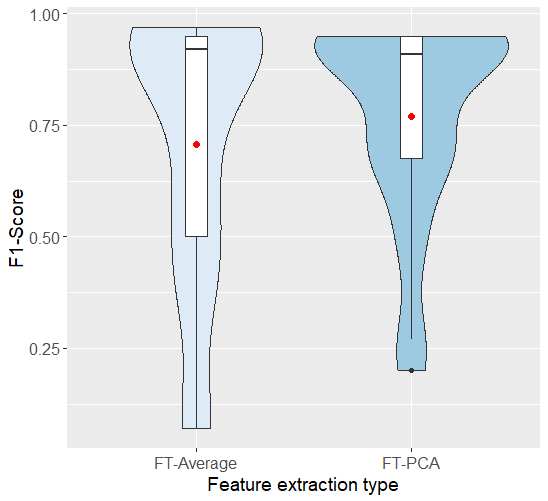}
\caption{Comparison between FastText-Average vs FastText-PCA} 
\end{subfigure}
\newline
\newline
\begin{subfigure}[b]{0.46\textwidth}
\includegraphics[width = 1\textwidth]{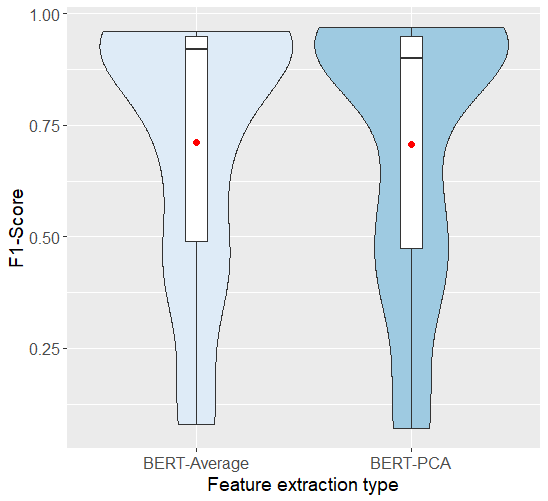}
\caption{Comparison between BERT-Average vs BERT-PCA} 
\end{subfigure}
\hspace*{5mm}
\begin{subfigure}[b]{0.46\textwidth}
\includegraphics[width = 1\textwidth]{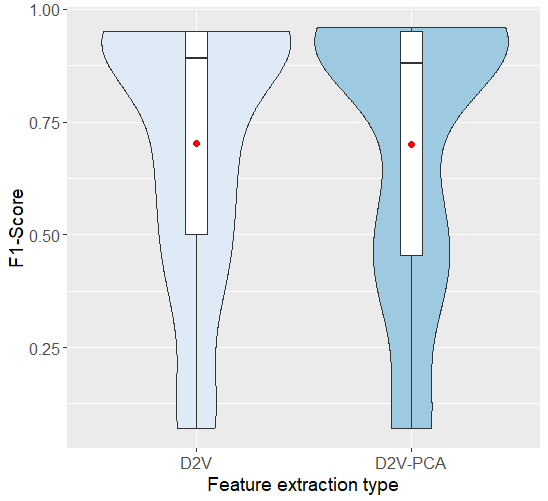}
\caption{Comparison between Doc2Vec vs Doc2Vec-PCA}
\end{subfigure}

\caption{Violin plots showing the distribution of 63 F1-Score values, over nine datasets and seven classifiers, for Word2Vec-Average vs Word2Vec-PCA (graph (a)), FastText-Average vs FastText-PCA (graph (b)), BERT-Average vs BERT-PCA (graph (c)) and Doc2Vec vs Doc2Vec-PCA (graph (d)). The red points indicate the mean value.}
\label{Chart3}
\end{figure}

Figure \ref{Chart3} shows the violin charts and box-and-whisker plots for the overall distributions of F1-Score values over 63 cases: 9 datasets and 7 classifiers, for Word2Vec-Average vs Word2Vec-PCA, FastText-Average vs FastText-PCA, BERT-Average vs BERT-PCA and Doc2Vec vs Doc2Vec-PCA, respectively. 
Based on these graphs, for Word2Vec and FastText, similar trends can be observed with the distributions of F1-Scores having longer lower tails when using averages in comparison to PCA and with considerably lower first quartiles when taking average (0.47 vs 0.66 for Word2Vec and 0.5 vs 0.66 for FastText), while the third quartiles were the same. The medians were slightly lower for PCA: 0.92 with averages vs 0.89 with PCA for Word2Vec and 0.92 with averages vs 0.91 with PCA for FastText. However, the opposite trend was observed for the mean values, which were markedly higher for PCA: 0.72 with averages vs 0.76 with PCA for Word2Vec and 0.71 with averages vs 0.77 with PCA for FastText. Generally, for Word2Vec and FastText, the overall distributions of the F1-Scores for PCA had a smaller range and were generally concentrated on higher values than for averages. For BERT and Doc2Vec, the distributions of F1-Scores for averages are very similar to the scores for PCA. 

In summary, for Word2Vec and FastText, our proposed PCA approach of combining word vectors using the first principal component shows clear advantages in performance over the traditional approach of taking the average. For BERT, the advantage of lowering the dimension of averaged vectors is less marked and generally the F1-Scores are similar for both approaches. Finally, for Doc2Vec, taking the entire WE vector tends to give slightly better F1-Scores than the PCA approach. Hence, when used for dimension reduction, the number of principal components could be tuned, for example via cross-validation, which, however, adds to the computational complexity of the task.

\FloatBarrier

\FloatBarrier

\subsubsection{Comparison between classifiers}
\label{Comparisonbetweentheclassifiers}

\begin{figure}[!htbp] 
\includegraphics[width = 1\textwidth]{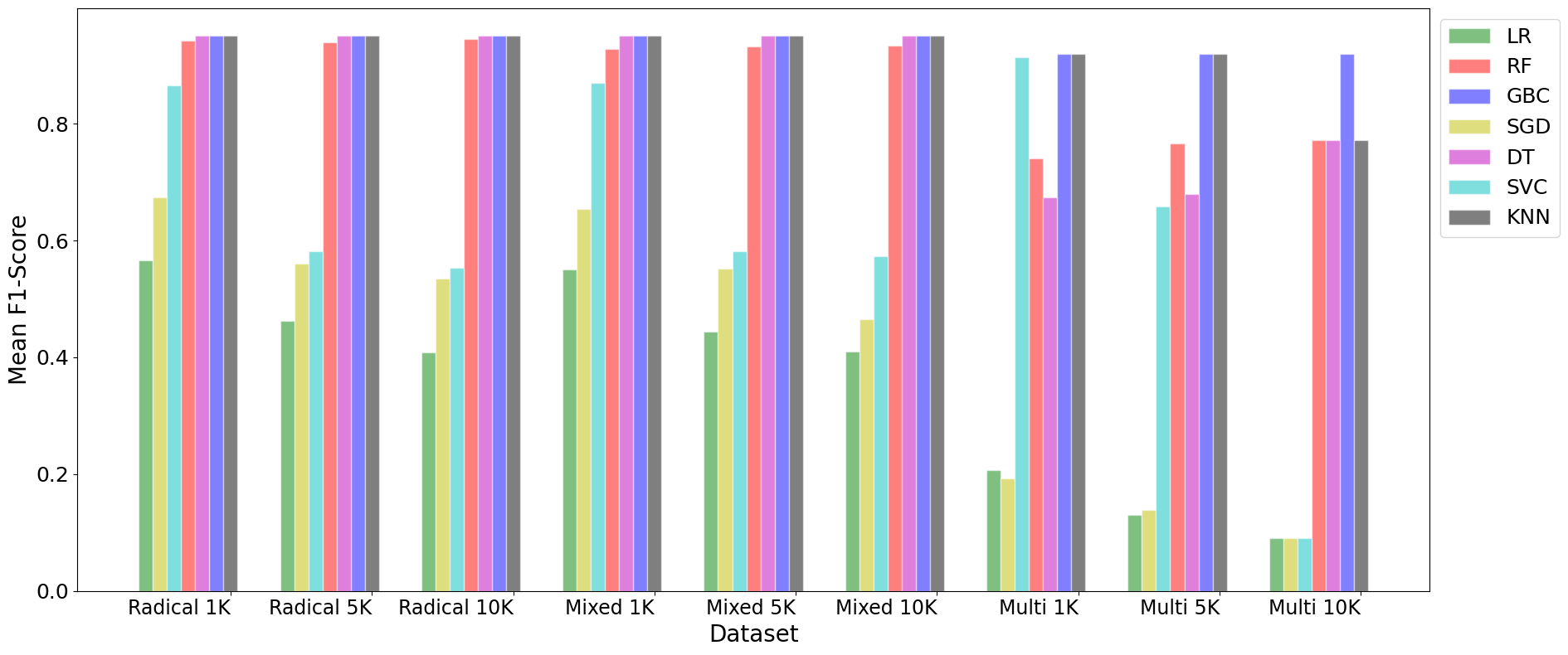}
\caption{Performance comparison between classifiers showing the average of F1-Score values over nine WEs, for each classifier and each dataset. }
\label{Chart7}
\end{figure}

\begin{figure}[!htbp] 
\centering
\includegraphics[width = 1\textwidth]{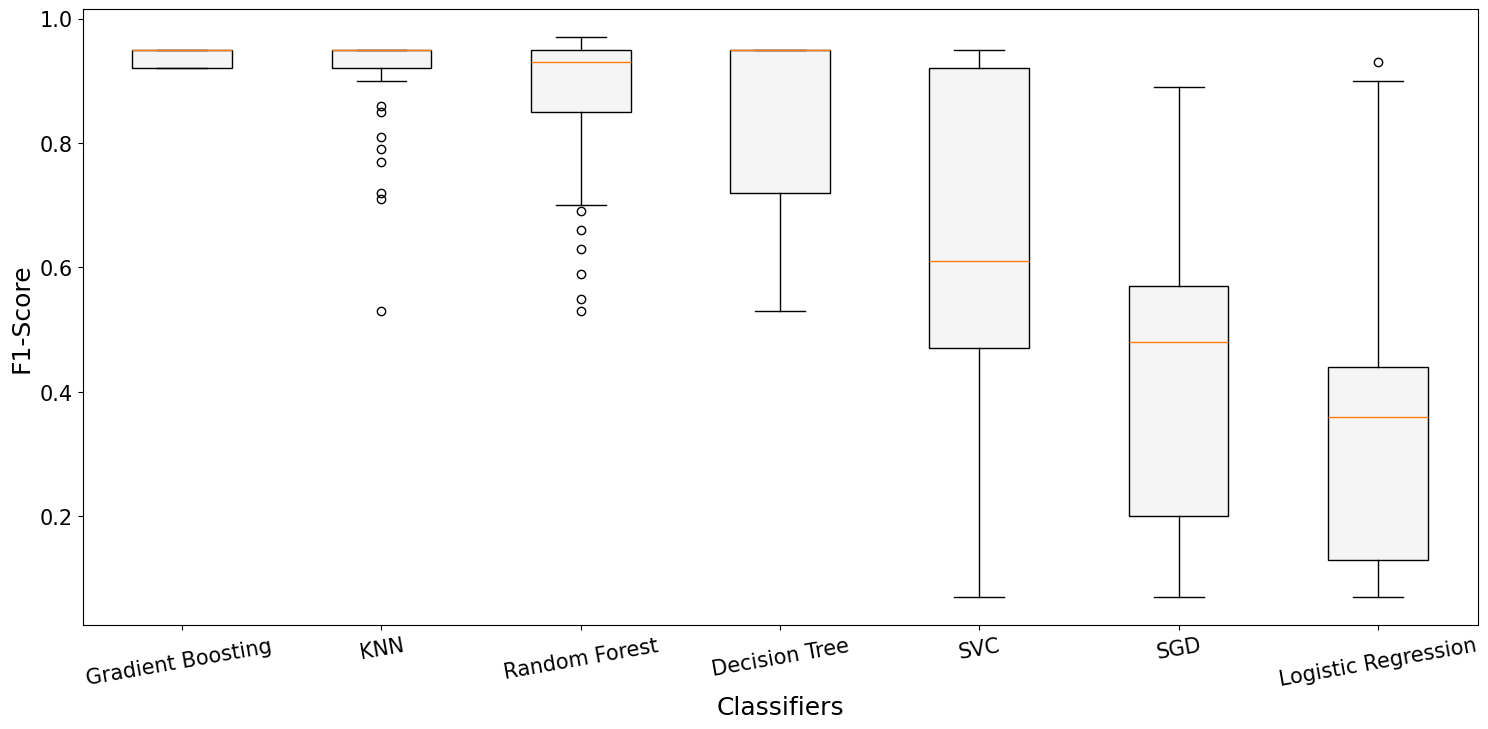}
\caption{Classifiers' performance comparison showing box plots of F1-Scores based on 63 values over the nine datasets and seven WEs, for each classifier, respectively.}
\label{Chart8}
\end{figure}

In this subsection, we comment on the overall performance of classifiers on all datasets and WEs considered.
Figure \ref{Chart7} visualises  the performance comparison between the seven classifiers for each datases with bars showing the mean F1-Score over all WEs for each classifier. Moreover, Figure \ref{Chart8} shows the box plots for the distributions of F1-Scores based on 63 values over the nine datasets and seven WEs, for each classifier, respectively. 

The charts reveal two classifiers with a consistently high performance: GBC and KNN. GBC had the same F1-Score value of $0.95$ for all WEs and all sample sizes for Radical Binary and Mixed Binary datasets and the value of $0.92$ for the Multi-Class datasets. 
For Radical Binary and Mixed Binary datasets, the same pattern occurred for KNN and DT, however for the Multi-Class datasets their scores were not as high as for GBC. Therefore, the GBC classifiers proved to be the most accurate even for the more challenging data with five classes. The KNN classifier was the second best after GBC, with lower scores occurring only for the Multi-Class dataset of size 10,000 for all WEs, ranging between 0.53 for TF-IDF and 0.9 for BERT-PCA.

The third best classifier was Random Forest (RF) with consistently high F1-Scores achieving a median of 0.93, mean of 0.87 and a narrow inter-quartile range of 0.1 indicating its stable performance over various WEs. Even though the RF was only slightly less accurate than the DT for the Radical Binary and Mixed Binary data, it outperformed the DT by a significant margin for the more challenging Multi-Class dataset achieving the mean F1-Score of 0.76 and median of 0.77 for this dataset, while the DT had the mean F1-Score of 0.71 and median of 0.68 in this case. 

The remaining three classifiers: SVC, SGD and LR performed much worse than GBC, KNN, RFs and DTs, with inconsistent scores affected by the sample size and WEs. 

The SVC classifier showed the highest variability and was the most sensitive to WEs among all other classifiers, with high accuracy in some cases and very low in others.  Its F1-Scores ranged between 0.07 and 0.95, with an interquartile range of 0.45, a median of 0.61 and a mean of 0.63. Moreover, the SVC classifier stood out as performing relatively well with the TF-IDF text representation.

The Stochastic Gradient Descent classifier performed poorly on the datasets studied. It achieved a median F1-Score of 0.48 and a mean of 0.42, which are lower by 0.13 and 0.19, respectively, than for the SVC. Particularly poor performance was observed for the Multi-Class datasets. However, the SGD classifier had less variation in performance than the SVC, with an overall interquartile range of 0.37. 

Lastly, the Logistic Regression (LR) model achieved an overall mean and median F1-Scores of 0.36, which is the lowest value among all employed classifiers. However, high scores were observed occasionally, with the maximum F1-Score of $0.93$ using Word2Vec-PCA and $0.9$ using FastText-PCA for Radical Binary dataset of size $n=1000$. Like for other classifiers, the poorest performance was observed for the Multi-Class dataset. In this case, the LR model failed to produce an effective discrimination rule and had F1-Scores close to zero in many cases, with a mean F1-Score of 0.14 and a median of 0.09. Like SGD and SVC, the LR classifier also had high variation in its performance showing relatively high sensitivity to the WEs employed. Overall, LR proved to be the weakest classifier for the considered datasets. 

In conclusion, the GBC and KNN classifiers demonstrated overall very high scores and stable performances and they were not particularly sensitive to WEs. RFs and DTs had slightly lower but also quite high scores and good overall performance. However, DTs performed better for simple binary datasets and were significantly less accurate for the multiclassification tasks. SVCs had poorer performance and stood out as particularly variable and sensitive to WEs. Lastly, SGD and LR performed very poorly and achieved the worst scores of all the considered classifiers, in almost all cases.

\FloatBarrier

\newpage \newpage

\subsection{Time - Classifiers' Performance Analysis}
\label{Time-Accuracy}

\begin{figure}[!htbp] 
\begin{center}
\includegraphics[height = 9cm]{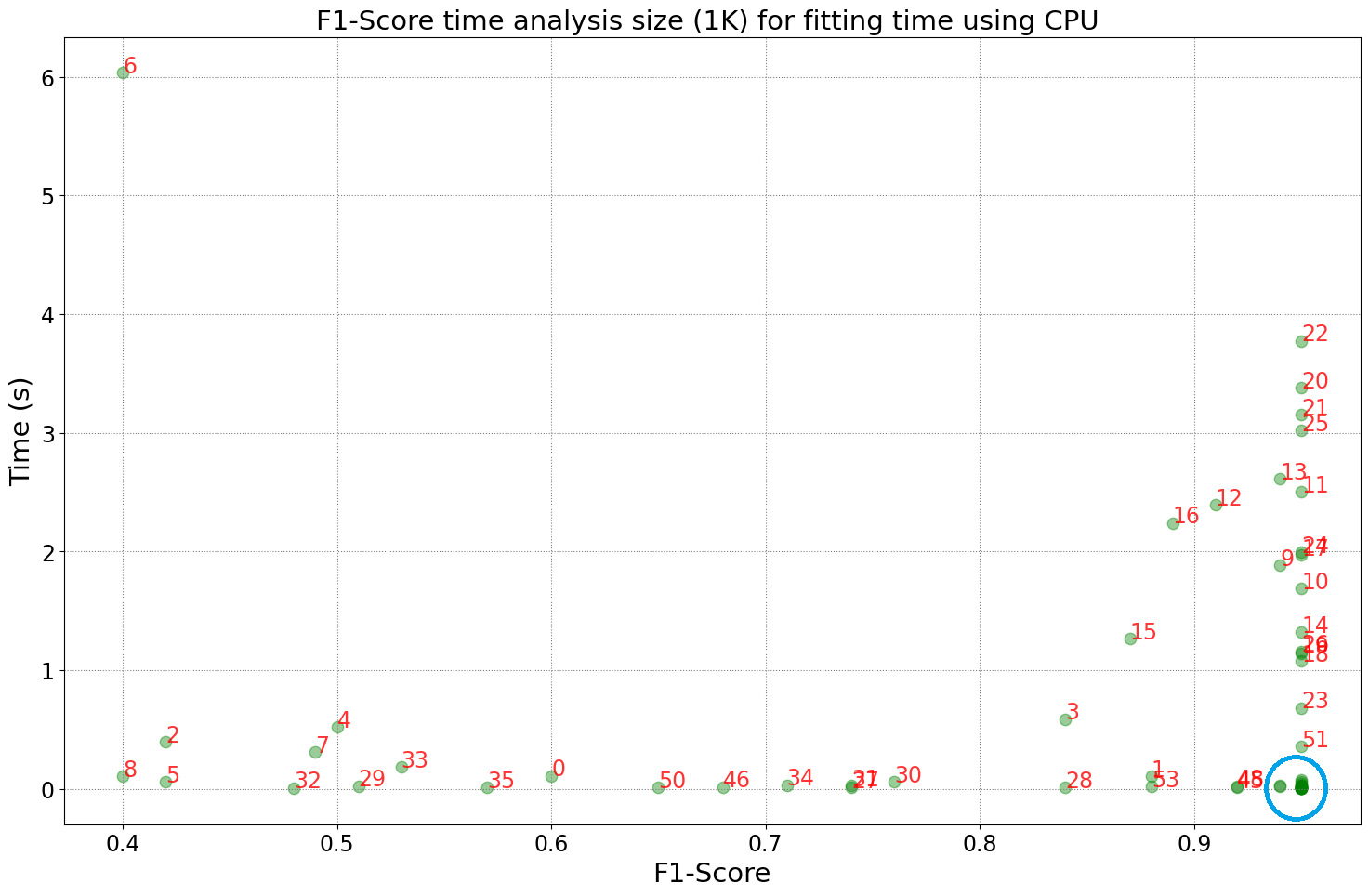 }
\end{center}
\caption{Fitting times against F1-Scores, using CPU, for the Mixed-Binary dataset of size 1,000, where the blue circle refers to overlapping points: 36-44, 47, 49, 52, 54-62. }
\label{Fig_F1ScoreTimeAnalysis3}
\end{figure}

In this subsection, we explore the relationship between time and F1-Score for the considered models and scenarios with the aim of assessing their efficiency and identifying the most and the least efficient approaches.
Figure \ref{Fig_F1ScoreTimeAnalysis3} shows a scatter plot of classifiers' fitting times versus F1-Scores for 63 cases, using the nine WEs and seven classifiers, for the Mixed Binary data of size 1,000, as an example, as very similar patterns occurred for all the other datasets. Each point is annotated by its index which can be interpreted using Table \ref{table:BMCTable35}, where the name of the WE and the classifier, as well as the fitting time, precision, recall and F1-Score values are listed. 

The scatter plot reveals that many points have a very high F1-Score values close to $1$ for a wide range of fitting times. 
On the other hand, many points have a very low fitting time for a wide range of F1-Scores. For example, for the Mixed Binary dataset, classifiers with fitting times below 0.11s have F1-Scores varying between 0.4 and 0.95. 
These patterns indicate that, overall, there is no straightforward or linear relationship between the fitting times and classifiers' performance. 

However, some positive correlation can be observed. Gradient Boosting classifiers (points marked with numbers 20, 21, 22, 25), as well as Random Forests (points marked with numbers 11, 13) had high performance coupled with relatively long fitting times and Logistic Regression (points 2, 5, 8) had low fitting times and low performance.  

We note an outlying point (number 6) associated with very low performance metrics and the longest fitting time that can be observed in Figure \ref{Fig_F1ScoreTimeAnalysis3}. 
This case refers to using TF-IDF with a LR classifier. This specific pair of TF-IDF coupled with LR had the same pattern of relatively high time consumption with low accuracy value also for other computing powers and data sizes indicating the worst overall efficiency.

The best efficiency scenario is characterized by a short fitting time and high F1-Score. For the Mixed-Binary data, this can be observed for the KNN and DTs classifiers (points clustered in the lower right corner of the graph in Figure \ref{Fig_F1ScoreTimeAnalysis3} marked with numbers 36 - 44, 47, 49, 52, 54 - 62), while for the Multi-Class dataset KNN is the most efficient, closely followed by most SVC classifiers. 

In conclusion, we find that the least efficient classifier was Logistic Regression applied to a TF-IDF text representation which had relatively long fitting times and poor performance, while the most efficient one was KNN with the fastest training times coupled with the best performance metrics irrespective of the WE used. 
Moreover, the relationship between the consumed time and performance measures can be described by a low to moderate positive correlation, indicating a lack of a simple or strong pattern for the relationship between the computational intensity and the quality of the resulting classification rule.

\FloatBarrier

\begin{table}[!htbp]
\scriptsize 
  \centering
  \begin{tabular}{c|c|l|c|c|c|c}
  \hline\hline
    Index&Classifier& Word embedding&Time (sec)&Precision&Recall&F1-Score\\ [0.5ex]

  \hline

0& & W2V-Average&0.11&0.6&0.7&0.6\\
1& &W2V-PCA&0.11&0.91&0.88&0.88\\
2& &FT-Average&0.39&0.38&0.56&0.42\\
3& Logistic &FT-PCA&0.59&0.89&0.85&0.84\\
4& Regression &BERT-Average&0.52&0.55&0.62&0.5\\
5& &BERT-PCA&0.06&0.38&0.56&0.42\\
6& &TF-IDF&6.03&0.32&0.55&0.4\\
7& &Doc2Vec&0.31&0.5&0.61&0.49\\
8& &Doc2Vec-PCA&0.11&0.37&0.55&0.4\\\hline
9& &W2V-Average&1.88&0.96&0.95&0.94\\
10& &W2V-PCA&1.69&0.96&0.95&0.95\\
11& &FT-Average&2.5&0.96&0.96&0.95\\
12& Random &FT-PCA&2.39&0.92&0.91&0.91\\
13& Forest &BERT-Average&2.61&0.95&0.94&0.94\\
14& &BERT-PCA&1.32&0.97&0.95&0.95\\
15& &TF-IDF&1.26&0.88&0.87&0.87\\
16& &Doc2Vec&2.24&0.9&0.89&0.89\\
17& &Doc2Vec-PCA&1.97&0.97&0.95&0.95\\\hline
18& &W2V-Average&1.08&0.97&0.95&0.95\\
19& &W2V-PCA&1.15&0.97&0.95&0.95\\
20& &FT-Average&3.38&0.97&0.95&0.95\\
21& Gradient &FT-PCA&3.15&0.97&0.95&0.95\\
22& Boosting &BERT-Average&3.77&0.97&0.95&0.95\\
23& &BERT-PCA&0.68&0.97&0.95&0.95\\
24& &TF-IDF&1.99&0.97&0.95&0.95\\
25& &Doc2Vec&3.02&0.97&0.95&0.95\\
26& &Doc2Vec-PCA&1.16&0.97&0.95&0.95\\\hline
27& & W2V-Average&0.01&0.74&0.8&0.74\\
28& &W2V-PCA&0.01&0.9&0.85&0.84\\
29& &FT-Average&0.02&0.51&0.62&0.51\\
30& Stochastic&FT-PCA&0.06&0.86&0.79&0.76\\
31& Gradient &BERT-Average&0.03&0.74&0.8&0.74\\
32& Descent &BERT-PCA&0.01&0.44&0.6&0.48\\
33& &TF-IDF&0.18&0.48&0.65&0.53\\
34& &Doc2Vec&0.03&0.68&0.78&0.71\\
35& &Doc2Vec-PCA&0.01&0.53&0.67&0.57\\\hline
36& & W2V-Average&0.01&0.97&0.95&0.95\\
37& &W2V-PCA&0.01&0.97&0.95&0.95\\
38& &FT-Average&0.03&0.97&0.95&0.95\\
39& Decision &FT-PCA&0.03&0.97&0.95&0.95\\
40& Tree &BERT-Average&0.04&0.97&0.95&0.95\\
41& &BERT-PCA&0.01&0.97&0.95&0.95\\
42& &TF-IDF&0.06&0.97&0.95&0.95\\
43& &Doc2Vec&0.03&0.97&0.95&0.95\\
44& &Doc2Vec-PCA&0.01&0.97&0.95&0.95\\\hline
45& &W2V-Average&0.01&0.93&0.92&0.92\\
46& &W2V-PCA&0.02&0.79&0.7&0.68\\
47& &FT-Average&0.02&0.97&0.95&0.95\\
48& Support &FT-PCA&0.02&0.93&0.92&0.92\\
49& Vector &BERT-Average&0.02&0.95&0.94&0.94\\
50& Classifier &BERT-PCA&0.01&0.78&0.68&0.65\\
51& &TF-IDF&0.36&0.97&0.95&0.95\\
52& &Doc2Vec&0.03&0.96&0.94&0.94\\
53& &Doc2Vec-PCA&0.02&0.9&0.88&0.88\\\hline
54& &W2V-Average&0.01&0.97&0.95&0.95\\
55& &W2V-PCA&0.01&0.97&0.95&0.95\\
56& &FT-Average&0.01&0.97&0.95&0.95\\
57& K-Nearest &FT-PCA&0.01&0.97&0.95&0.95\\
58& Neighbours &BERT-Average&0.01&0.97&0.95&0.95\\
59& &BERT-PCA&0.0&0.97&0.95&0.95\\
60& &TF-IDF&0.08&0.97&0.95&0.95\\
61& &Doc2Vec&0.01&0.97&0.95&0.95\\
62& &Doc2Vec-PCA &0.0&0.97&0.95&0.95\\\hline

\hline
  \end{tabular} 
\caption{Precision, Recall and F1-Score Time analysis using CPU for Mixed Binary data of size 1,000. }
\label{table:BMCTable35} 
\end{table}

\FloatBarrier

\subsection{Estimation of Energy Consumption for Word Embeddings}
\label{energy}

The use of machine learning models that run on modern computing systems such as cloud computing and parallel processing units including the use of GPUs, CPUs and TPUs, requires significant amounts of energy. 
Some estimates of the electricity consumption of data centers are of the order 2\% of the total energy generated in the world 
\citep{freitag2021real, terenius2023material}. 
The large use of energy required in machine learning and AI may have adverse environmental effects. This motivates studying the energy used in machine learning. The aim of this subsection is to illustrate energy consumption and CO2 emissions for the WEs employed in this study, for various computing settings and identify the most and the least environment friendly cases.

There have been some discussions in the scientific community on how to study machine learning algorithms that also considers energy consumption \citep{lottick2019, mehlin2023towards}. The paper by \cite{bannour2021evaluating} reviews energy consumption of NLP methods. Understanding their respective contributions to emissions is important in order to highlight the cases of unnecessary data storing and processing and their deep effects on the environment, in accordance with the recommendation of \cite{GarciaMartinA2019}, as more GPU usage is needed globally due to expanding applications of machine learning methods. 
\cite{GarciaMartinA2019} monitored many experiments to measure the estimated energy consumption by machine learning kernels. In 2006, \cite{LeeBrooks2006} presented a model to calculate the power consumed using regression analysis applied to a micro-architectural dataset. 
In 2014, \cite{Laurenzano2014} built an application for HPC benchmarks to calculate performance and energy by measuring wall power. 
In 2017,  \cite{Mazouz2017} presented a methodology to calculate the consumed energy of hardware processors.

In this short study, we use the Python library 'CodeCarbon' \citep{benoit_courty_2024_11171501} to measure the amount of energy and CO2 emissions,  
for various processing sets, 
including:
 T4-GPU-HighRAM,
 CPU-HighRAM,
 CPU,
 T4-GPU,
 A100-GPU-HighRAM,
 V100-GPU,
 V100-GPU-HighRAM.
Figure \ref{EmissionAndEnergyVsWE} shows produced CO2 emission and consumed energy vs Word Embedding, where each line reflects a different processing set for the Mixed-Binary dataset of size 1,000, as an example. These values are the average of ten trials for each of the seven processing sets.
Moreover, Figure \ref{FEEnergy} shows scatter plots of duration versus CO2 emission and duration versus energy consumed, respectively.

\subsubsection{Comparison of word embeddings for a fixed processing set}

\begin{figure}[!htbp] 
\begin{center}
\includegraphics[height = 9cm]{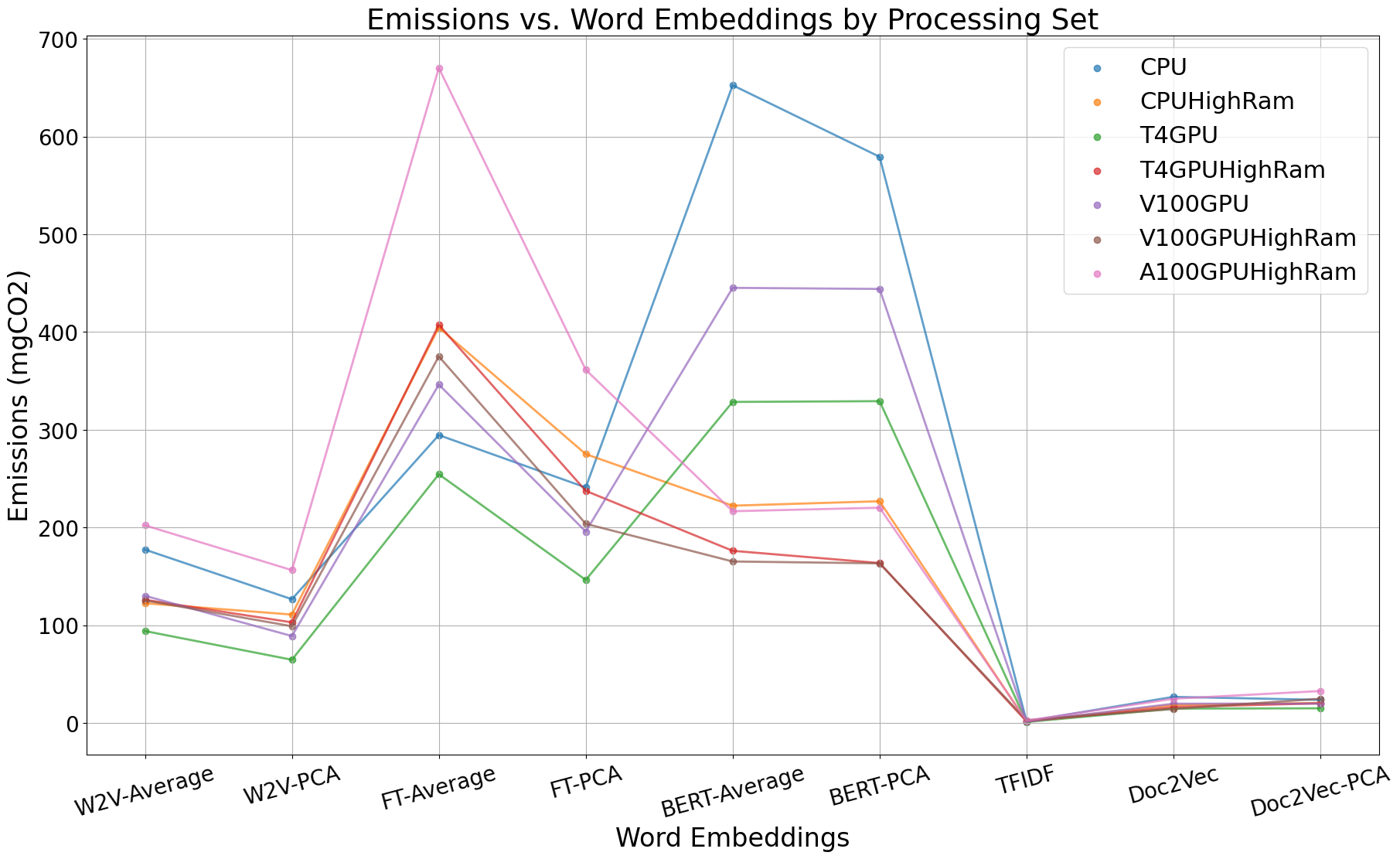}
\includegraphics[height = 9cm]{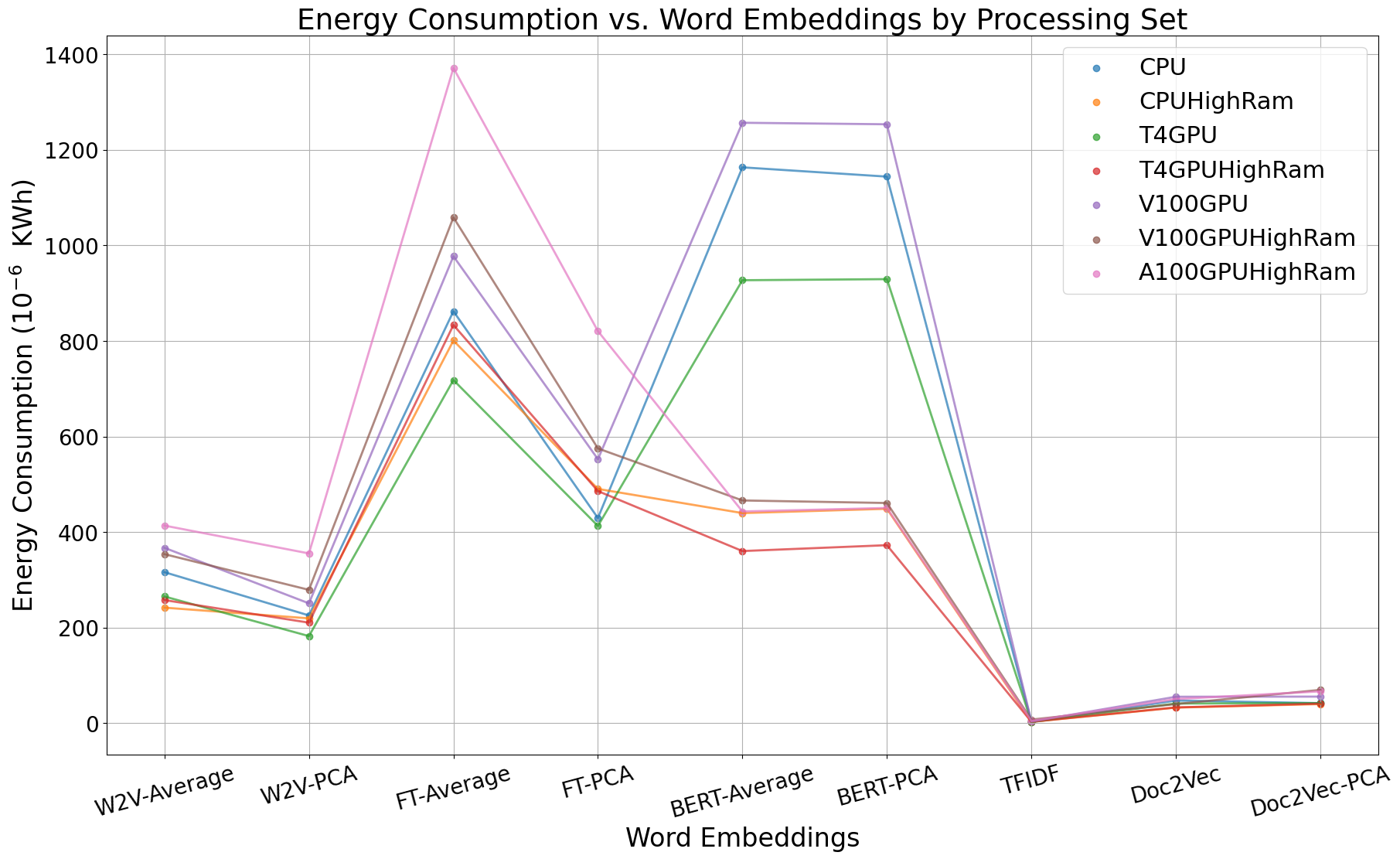}

\end{center}
\caption{Emissions and Energy Consumed vs Word Embedding graphs for Feature Extraction using Mixed-Binary dataset of size $1,000$, for several computing sets. }
\label{EmissionAndEnergyVsWE}
\end{figure}

Figure \ref{EmissionAndEnergyVsWE} reveals that for the four computing settings that use HighRAM (T4-GPU-HighRAM, CPU-HighRAM, A100-GPU-HighRAM, V100-GPU-HighRAM), FastText-Average WEs consumed the most energy and had the highest CO2 emissions. For example, for the T4-GPU-HighRAM processing set, FastText-Average had approximately 4 times more emissions and energy consumption than W2V-PCA and around 25 times more than Doc2Vec. FastText-PCA was the second most energy-consuming and CO2-emitting WE for HighRAM computing settings, with an energy consumption around 60\% of that needed by FastText-Average. On the other hand, for the three processing settings that do not use HighRAM (CPU, T4-GPU and V100-GPU), BERT-Average and BERT-PCA were the two most energy-consuming and CO2-emitting WEs, needing around 1.3 times as much energy as FastText-Average, which had the second highest energy consumption and CO2 emissions, around 3.7 times as much as W2V-Average and 25 times as much as Doc2Vec. 

On the other side of the spectrum, TF-IDF used the least energy and had the lowest CO2 emissions for all processing settings, by two orders of magnitude in most cases. For example, for T4-GPU-HighRAM, TF-IDF had around 260 times less emissions and energy consumption than FastText-Average.
This can be explained by the nature of TF-IDF, as even though it contains a large number of columns, the majority of them are null values and hence it is stored in the memory as sparse matrices and given that we used short texts in this study the TF-IDF process is quite energy efficient. The second most environmentally friendly WE was Doc2Vec and it was closely followed by Doc2Vec-PCA. These two WEs stood out from others by a significant margin, having at least an order of magnitude less usage of energy and CO2 emission. Lastly, the two Word2Vec WEs can be described as medium in their energy consumption and CO2 emissions compared to the others.

For a fixed processing set, the time consumption was strongly positively correlated with energy and emissions, respectively, as illustrated in Figure \ref{FEEnergy}. Hence, the WE with the shortest time always had the lowest energy consumption and CO2 emissions and vice versa. However, the magnitude of differences between WEs varied according to the slope coefficient, which was different for each processing set, so that, for example, having a double time would not necessarily be related with a double energy consumption or emissions.

\subsubsection{Comparison of processing sets}
\label{proc}

\begin{figure}[!htbp] 
\begin{center}
\includegraphics[height = 7cm]{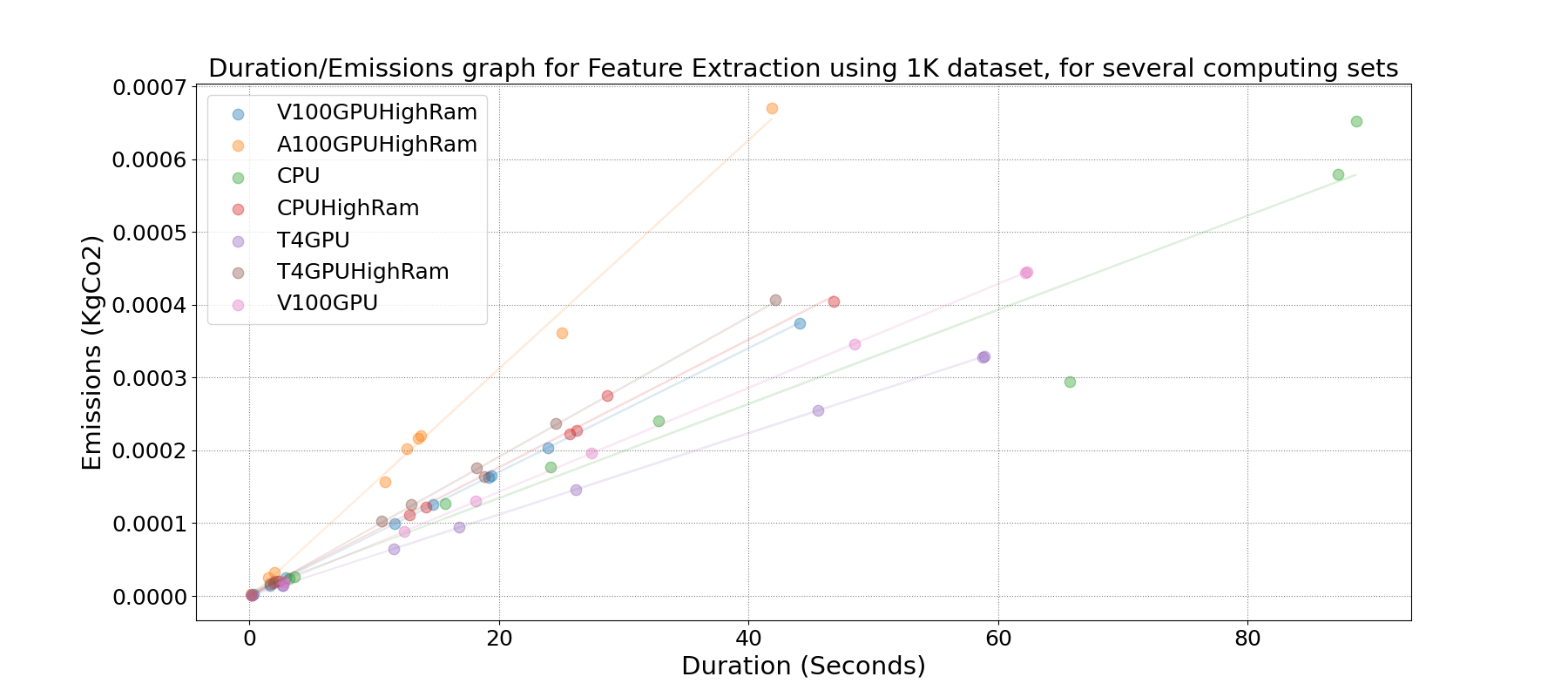}
\includegraphics[height = 7cm]{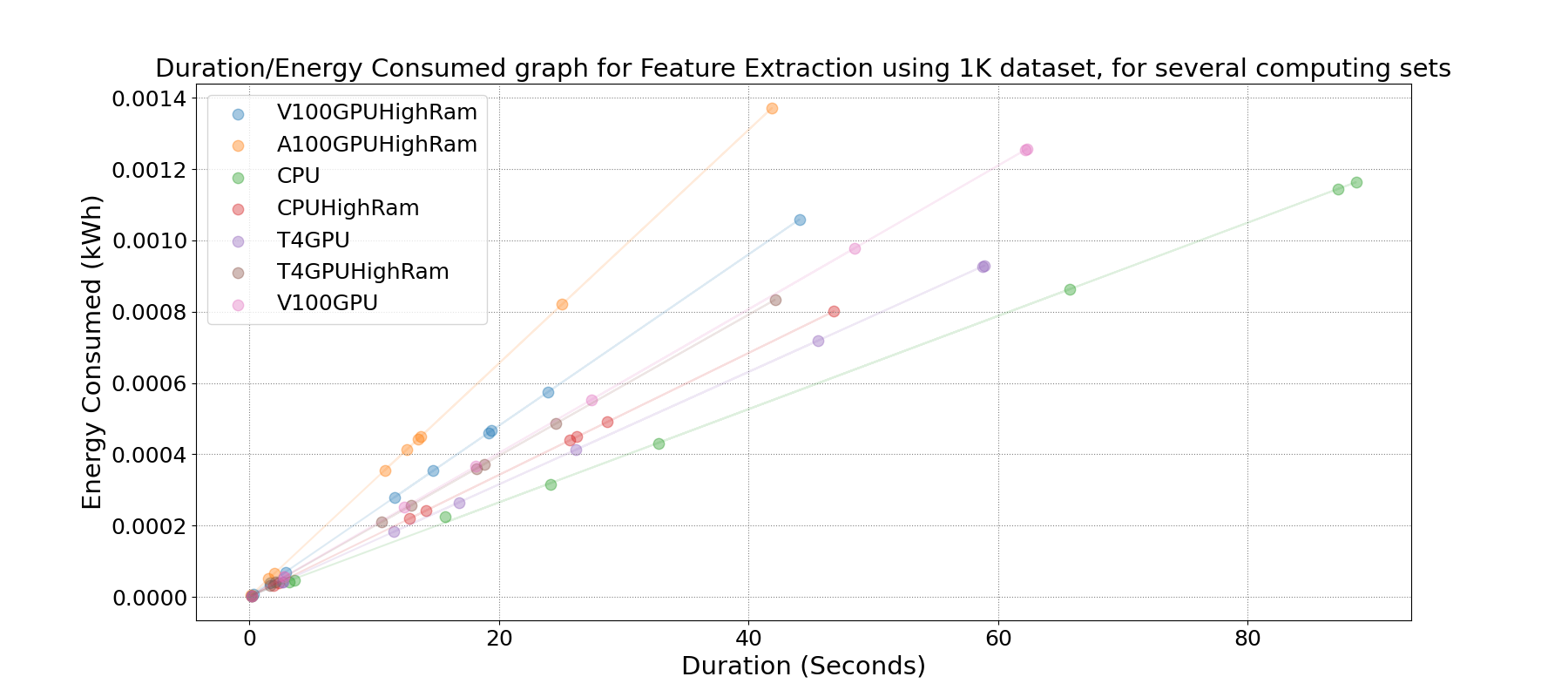}
\end{center}
\caption{Duration/Emissions and Energy Consumed graph for Feature Extraction using Mixed-Binary dataset of size $1,000$, for several computing sets, with added least squares lines.}
\label{FEEnergy}
\end{figure}

Changing the processing set also significantly affected CO2 emissions and energy consumed. 
 
For Word2Vec and FastText (both, Average and PCA approaches), T4-GPU led to the lowest CO2 emissions and energy consumption among all other processing sets, while A100-GPU-HighRAM - the highest.
A similar pattern occurred for TF-IDF, as T4-GPU was still the most environmentally friendly. However, in this case V100-GPU-HighRAM led to the highest CO2 emissions and energy consumption, with A100-GPU-HighRAM being the second highest. Also, for Doc2Vec and Doc2Vec-PCA, T4-GPU was related to the lowest CO2 emissions and CPU-HighRAM - the lowest energy consumption.

For BERT, the highest CO2 emissions and the most energy consumption were observed for CPU and V100-GPU. On the other hand, the lowest CO2 emissions and the least energy consumption occurred for V100-GPU-HighRAM and T4-GPU-HighRAM. Generally, BERT-based WEs showed a different pattern from the other WEs with respect to the processing sets' effects, in that HighRAM settings lead to a more environmentally friendly performance than non-highRAM settings. By comparing two related sets: CPU and CPU-HighRAM, we notice that the CPU-HighRAM values for Duration, Emission and Energy are always less than the corresponding CPU values and the same pattern holds for T4-GPU against T4-GPU-HighRAM and V100-GPU against V100-GPU-HighRAM. For example,  for BERT-Average, using the CPU processing set needed three times the amount of energy that is needed when using CPU-HighRAM and twice the amount of emission.

However, for other WEs this pattern does not always hold.  In particular, for FastText-Average, FastText-PCA and TF-IDF, HighRAM settings usually have higher CO2 emissions and energy consumption. It is also important to note that, in some cases, shorter time does not lead to smaller energy and CO2 emissions. For FastText-PCA and TF-IDF, although the task was completed for CPU-HighRAM in a shorter time than for CPU, the CO2 emissions and energy consumption were higher for CPU-HighRAM. 

The scatter plots in Figure \ref{FEEnergy} show that changing the processing set leads to a change in the slope of the best-fit line, which reflects the rate of change of emission or energy, with time. 
Processing sets that usually need more power like GPU and HighRAM, have a higher slope than standard processing sets like CPU.

For a fixed amount of time consumption, A100-GPU-HighRAM lead to the most energy consumed and the highest CO2 emissions, which is demonstrated in Figure \ref{FEEnergy} by the yellow line of best fit being above all other lines in both scatter plots. T4-GPU-HighRAM had the second highest emissions (brown line in the top scatter plot), while V100-GPU-HighRAM had the second highest energy consumption (blue line in the bottom scatter plot). On the other hand, T4-GPU had the lowest CO2 emissions (purple line in the top plot) and CPU had the lowest energy consumption (green line in the bottom plot) for a fixed duration.

 \subsubsection{Conclusion}

The study in this subsection emphasizes the environmental impact of computations and the need for more energy-saving models, as different processing sets and WEs lead to different values of produced emissions and consumed energy, which has a potential impact on the Earth's future.
We found that TF-IDF and Doc2Vec are the most energy efficient embeddings, while FastText-Average and BERT-Average and PCA models are the most resource-consuming. 

However, WE models require a trade-off between computational efficiency and classification accuracy, as faster models, such as TF-IDF, often lack the depth of contextual understanding compared to more sophisticated models that need more time to be trained, as seen in subsection \ref{Time-Accuracy}. Thus, when selecting an embedding approach, one must carefully balance speed, energy consumption and accuracy, depending on the specific requirements of the application.

\FloatBarrier

\section{Conclusion}
\label{ConclusionSection}

This paper examined various word embedding techniques for classifying customer star ratings based on telecom reviews, comparing their effectiveness across different classification models. The results showed that while traditional TF-IDF performed poorly in most cases, it excelled when used with the Support Vector Classifier. BERT-PCA and Doc2Vec-PCA emerged as the most effective WEs, particularly for complex multiclass classification tasks. Word2Vec-PCA and FastText-PCA performed consistently well with Logistic Regression and Stochastic Gradient Descent, highlighting their robustness in certain models. Decision Trees, KNN and Gradient Boosting classifiers demonstrated relative insensitivity to WEs, except in larger multiclass datasets where BERT-PCA led to superior results. Regarding feature engineering, for Word2Vec and FastText, our proposed PCA approach of combining word vectors using the first principal component shows clear advantages in performance over the traditional approach of taking the average. Additionally, energy consumption analysis revealed that computational efficiency varies significantly among embedding techniques, with TF-IDF being the most resource-efficient and FastText the most computationally demanding. The findings emphasize the importance of selecting the appropriate WE based on classification complexity, model compatibility and computational constraints.

\acks{We thank David Yearling, Sri Harish Kalidass and Michael Free for discussions, particularly on sources of datasets. 
}

\appendix

\section{Classifiers' settings for subsection \ref{classifiers1}}
\label{appendix4}

The classifiers were imported in Python language from the Sklearn library, using the following parameters, respectively:

\begin{itemize}
    \item Logistic Regression: {\it penalty= l2, solver='sag', C=1.0, fit\textunderscore intercept=True, solver='lbfgs', max\textunderscore iter=100}

    \item Decision Tree: {\it criterion='gini', max\textunderscore depth=3, splitter='best', min\textunderscore samples\textunderscore split=2, min\textunderscore samples\textunderscore leaf=1 }    
    \item Random Forest: {\it criterion='gini', n\textunderscore estimators=300, max\textunderscore depth=7, min\textunderscore samples\textunderscore split=2, min\textunderscore samples\textunderscore leaf=1, min\textunderscore weight\textunderscore fraction\textunderscore leaf=0.0}    
    \item Gradient Boosting : {\it n\textunderscore estimators=100, max\textunderscore depth=3, loss='log\textunderscore loss', learning\textunderscore rate=0.1, subsample=1.0, criterion='friedman\textunderscore mse' }    
    \item Stochastic Gradient Descent: {\it penalty='l2', loss='log', learning\textunderscore rate='optimal', alpha=0.0001, l1\textunderscore ratio=0.15, fit\textunderscore intercept=True}    
    \item Support Vector Machine: {\it kernel='rbf', max\textunderscore iter=100, C=1.0, gamma='auto', degree=3, coef0=0.0}    
    \item K-Nearest Neighbour: {\it n\textunderscore neighbors=5, weights='uniform', algorithm='auto', leaf\textunderscore size=30, p=2, metric='minkowski'}    
\end{itemize}

\vskip 0.2in
\bibliography{sample}

\end{document}